\documentclass[11pt]{article}

\usepackage[final]{acl}

\usepackage{times}
\usepackage{latexsym}

\usepackage[T1]{fontenc}
\usepackage[utf8]{inputenc}

\usepackage{microtype}

\usepackage{inconsolata}

\usepackage{graphicx}

\usepackage{booktabs}
\usepackage[table, dvipsnames]{xcolor}
\usepackage{amsmath}
\usepackage{amsfonts}
\usepackage{multirow}
\usepackage{mathtools}
\usepackage{tcolorbox}
\usepackage{enumitem}

\newcommand{\OurMethod}{BIRCH}
\newcommand{\OurMethodLong}{\textbf{B}alanced \textbf{I}nformativeness and Co\textbf{R}rectness with a Truthful An\textbf{CH}or}
\newcommand{\hr}{\noindent\rule{\textwidth}{1pt}}
\newcommand{\highlight}{\cellcolor{lightgray}}

\title{When Vision-Language Models Judge Without Seeing: Exposing Informativeness Bias}

\author{
    Xiaohan Zou\textsuperscript{1}\thanks{Work done during internship at Oracle AI.} \quad Roshan Sridhar\textsuperscript{2} \quad Mohammadtaher Safarzadeh\textsuperscript{2} \quad Dan Roth\textsuperscript{2}  \\[0.3em]
    \textsuperscript{1}The Pennsylvania State University \quad \textsuperscript{2}Oracle AI \\[0.3em] {\fontsize{10.5pt}{10.5pt}\selectfont\texttt{xfz5266@psu.edu}, \texttt{\{roshan.sridhar, mohammadtaher.safarzadeh, dan.roth\}@oracle.com}}
}

\begin{document}
\maketitle

\begin{abstract}
% The reliability of VLM-as-a-Judge is critical for evaluating vision-language models and providing trustworthy supervision rewards for training. 
The reliability of VLM-as-a-Judge is critical for the automatic evaluation of vision-language models (VLMs). Despite recent progress, our analysis reveals that VLM-as-a-Judge often pays limited attention to the image when making decisions. Instead, they often blindly favor the more informative answer, even when they can recognize it conflicts with the image content. We call this problem \emph{informativeness bias}, which significantly undermines judge reliability. To address it, we propose \OurMethod\ (\OurMethodLong), a judging paradigm that first corrects inconsistencies with the image content in candidate answers, and then compares the answers against this corrected version. This shifts the judge's focus from informativeness to image-grounded correctness. Experiments on multiple models and benchmarks show that \OurMethod\ reduces informativeness bias by up to 17\%, resulting in performance gains of up to 9.8\%. Our work reveals an overlooked but fundamental flaw in current VLM-as-a-Judge systems and highlights the need for more principled designs.
\end{abstract}
\begin{figure}[t]
    \centering

    \definecolor{mygreen}{HTML}{588157}
    \definecolor{myorange}{HTML}{ee6c4d}

    \newcommand{\greenbox}{%
      \smash{%
        \raisebox{-0.1ex}[0pt][0pt]{
          \tcbox[
            on line,
            colback=mygreen,
            colframe=mygreen,
            boxrule=0pt,
            arc=1pt,
            boxsep=0pt,
            left=0pt,right=0pt,top=0pt,bottom=0pt
          ]{\phantom{\rule{8pt}{6pt}}}%
        }%
      }%
    }
    
    \newcommand{\orangebox}{%
      \smash{%
        \raisebox{-0.1ex}[0pt][0pt]{
          \tcbox[
            on line,
            colback=myorange,
            colframe=myorange,
            boxrule=0pt,
            arc=1pt,
            boxsep=0pt,
            left=0pt,right=0pt,top=0pt,bottom=0pt
          ]{\phantom{\rule{8pt}{6pt}}}%
        }%
      }%
    }

    \includegraphics[width=\columnwidth]{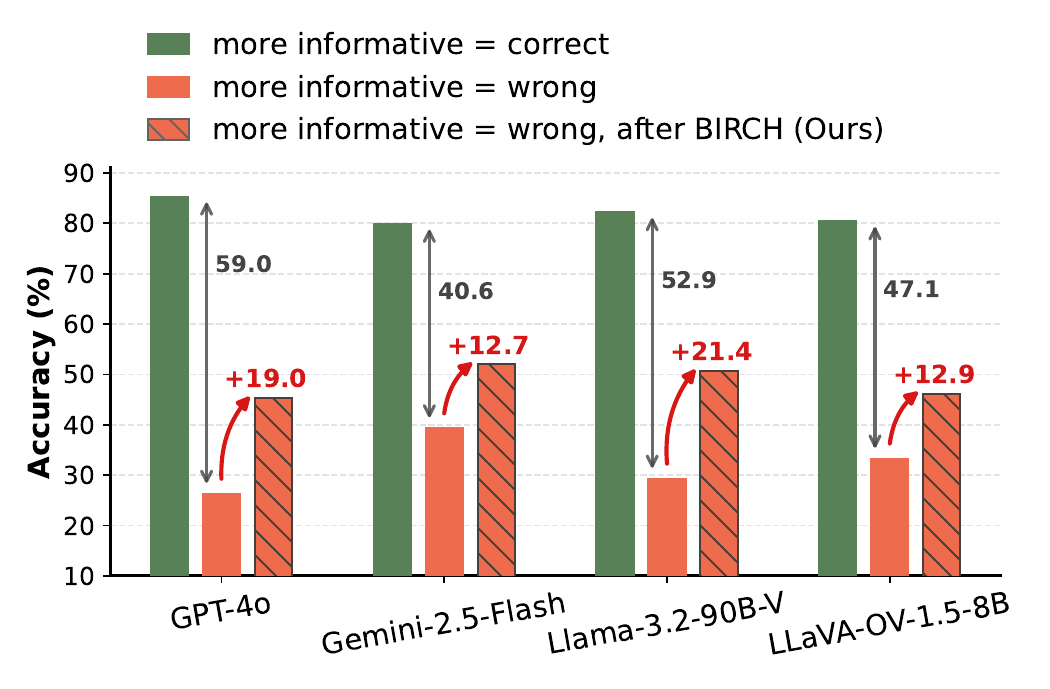}

    \caption{VLM-as-a-Judge performs well when the more informative answer happens to be correct (\!\greenbox\ more informative = correct). But when the more informative answer is actually wrong (\!\orangebox\ more informative = wrong), accuracy drops sharply by 40.6-59.0\% across models. This shows judges over-prioritize informativeness over truth. Our method \OurMethod\ greatly improves performance in these cases, thereby boosting overall results.}

    % \caption{\todo{legend (text: more info = correct/wrong, position -> circle + arrow + before/after, alignment -> accuracy)} We illustrate that VLM-as-a-Judge is heavily swayed by informativeness, performing much worse when the more informative answer is actually rejected by human (top). Our method improves performance on these cases and boosts overall results (bottom).}
    
    % \caption{\textbf{\todo{current issues we found with numbers} VLM-as-a-Judge with a carefully designed self-generated reference from \OurMethod\ outperforms the standard judge pipeline}. This performance gain comes from significantly reducing informativeness bias by improving accuracy on the correctness-dominated subset (CDS), which includes cases where humans prefer the more accurate but less informative answer.}
    \label{fig:results-showcase}
    
\end{figure}

\section{Introduction}

\begin{figure*}[t]
    \centering
    \includegraphics[width=\linewidth]{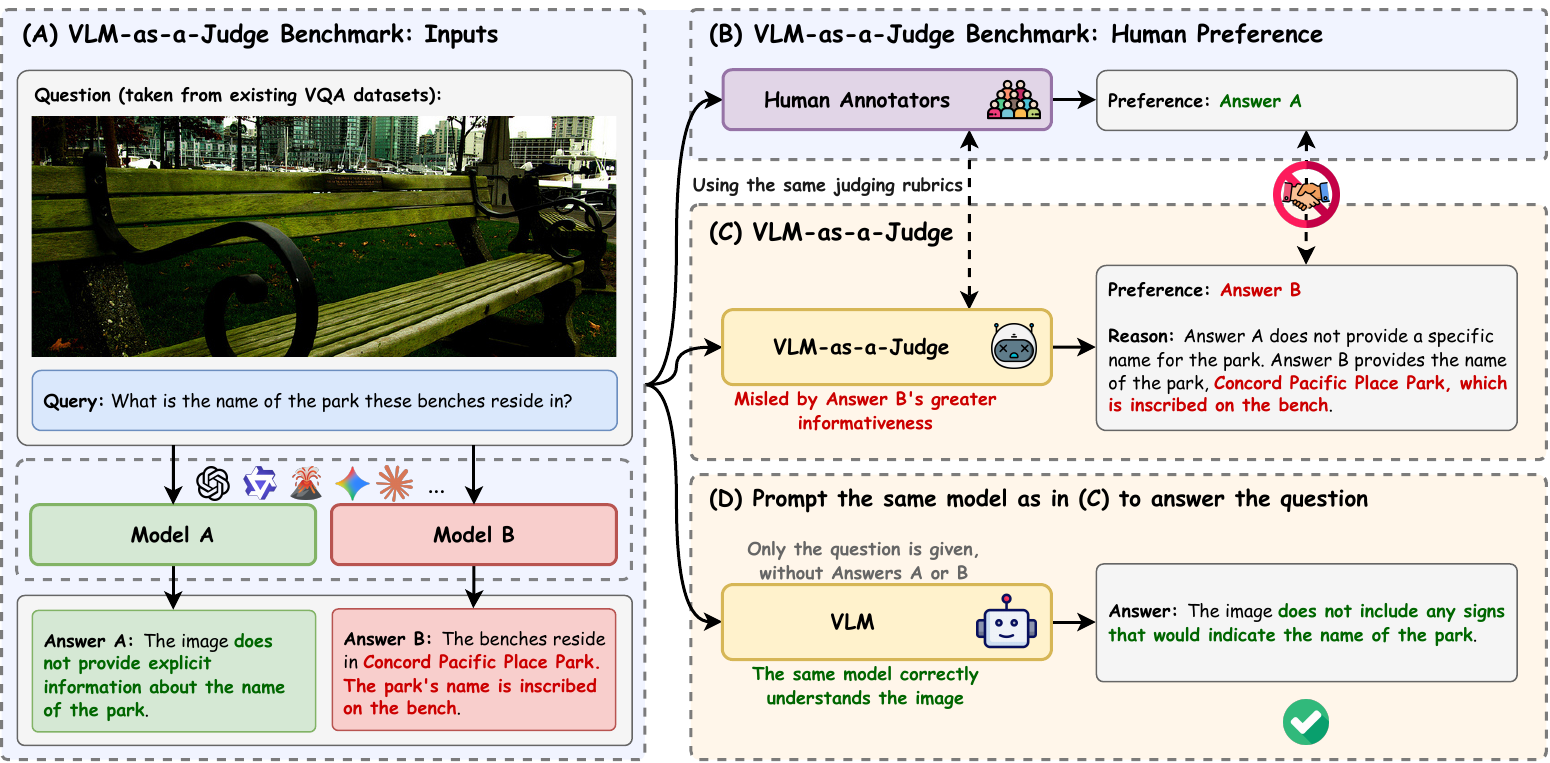}

    \caption{\textbf{(A) VLM-as-a-Judge input}: A VQA question (query + image) is given to two different response models, producing candidate answers A and B. The question asks for the park's name. Answer B looks more informative by giving a name and claiming it is inscribed on the bench, while Answer A is correct since the name is not visible in the image. \textbf{(B) Human preference}: Human annotators, given the query, image, answers, and judging rubrics, choose Answer A as the preference label. \textbf{(C) VLM-as-a-Judge decision}: The judge receives the same input and is prompted with the same judging rubrics as the humans. It is misled by Answer B's greater informativeness, mistakenly selecting it because it provides a specific (but wrong) name, while Answer A does not—misaligning with human judgment. \textbf{(D) Direct model response}: When asked to answer the question directly, the same VLM correctly sees that the name cannot be determined from the image. 
    % \textbf{(A) and (B)} are covered by existing benchmarks, which provide questions, candidate answers,  and human labels. \textbf{(C) and (D)} highlight a gap: VLM-as-a-Judge lags behind its vision-language understanding, as it can still be misled by informativeness even when inconsistencies with the image are detected. This paper aims to close that gap and fully leverage VLMs for reliable judgment.
    Here, \textbf{(A) and (B)} are covered by existing benchmarks. \textbf{(C) and (D)} highlight a gap between VLM-as-a-Judge's actual recognition ability and its judging behavior, caused by being misled by informativeness. This paper aims to close that gap for more reliable judgment.
    }
    \label{fig:intro}
    
\end{figure*}

Vision-Language Models (VLMs) have demonstrated impressive performance across a wide range of multimodal perception and reasoning tasks \cite{hurst2024gpt, zhang2021mme, yue2024mmmu}. Building on this, they are now widely used as automatic judges to evaluate responses and guide reward generation \cite{chen2024mllm, yasunaga2025multimodal}. VLM-as-a-Judge is typically applied in pairwise comparison tasks, where two candidate answers to a question about an image are compared to determine which is better. The judgment is then evaluated against human annotations \cite{jingfgaif, yu2024rlaif, zhou2024calibrated}, as shown in Figure \ref{fig:intro} (A), (B) and (C).

% Despite the growing adoption of VLM-as-a-Judge, its reliability remains underexplored. Various biases have been reported in LLM-as-a-Judge settings, raising concerns about fairness and trust in evaluation tasks \cite{gu2024survey, shi2024judging, ye2025justice}. For example, \citet{zheng2023judging} identify positional bias, where the judge favors certain answer positions; \citet{wang2024large} report length bias, where longer responses are preferred; and \citet{chen2024humans} show that fabricated citations can reduce judgment accuracy. While these studies highlight issues in LLM-as-a-Judge, the vision-language setting remains less explored. Existing work on VLM-as-a-Judge mainly focuses on increasing task diversity and difficulty \cite{chen2024mllm, li2025vl, yasunaga2025multimodal}, but often overlooks whether vision-language judgment is being made and evaluated as intended.

Despite the increasing use of VLM-as-a-Judge, its reliability is still underexplored. In LLM-as-a-Judge, several biases have been reported, such as positional bias, where judges favor certain answer positions \cite{zheng2023judging}, length bias, where longer responses are preferred \cite{wang2024large}, and authority bias, where fabricated citations reduce judgment accuracy \cite{chen2024humans}. These issues raise concerns about fairness and trust in evaluation tasks \cite{gu2024survey, shi2024judging, ye2025justice}. However, the vision-language setting remains less explored. Existing work on VLM-as-a-Judge mainly focuses on increasing task diversity and difficulty \cite{chen2024mllm, li2025vl, yasunaga2025multimodal}, but often overlooks whether vision-language judgment is being made and evaluated as intended.

To investigate this, we analyzed popular models and found that they pay little attention to images when making judgments. Removing the image from their inputs causes only a minimal drop in performance (Section \ref{sec:bias-blind}). A deeper look reveals a striking behavior: VLM-as-a-Judge is heavily deceived by the answer \emph{informativeness}, following the definition in \citet{lin-etal-2022-truthfulqa, wu2025balancing}:

\begin{figure}[h]
\centering

\vspace{-0.5em}

\begin{tcolorbox}[colback=gray!10, left=2pt, right=2pt, top=2pt, bottom=2pt, boxrule=1pt, arc=2pt, before upper={\fontsize{10}{12}\selectfont}]

\textbf{Informativeness:} How richly an answer goes beyond a minimal reply, by adding explanations, facts, context, or examples.

\end{tcolorbox}

\vspace{-1.5em}

\end{figure}

\noindent They often favor the more informative answer despite clear inconsistencies with the image, even when they can recognize those inconsistencies (Section \ref{sec:bias-ib}). For example, in Figure \ref{fig:intro} (A), Answer A is more truthful since it avoids a false claim about the park's name, but is less informative. Answer B is more informative by providing a name, though it is wrong. This shows that greater informativeness can increase the risk of speculation or error \cite{lin-etal-2022-truthfulqa, wu2025balancing}. However, VLM-as-a-Judge still incorrectly selects Answer B (Figure \ref{fig:intro} (C)), despite knowing the image gives no clue about the park's name (Figure \ref{fig:intro} (D)). As a result, VLM-as-a-Judge performs much worse when the more informative answer is wrong than when it happens to be truthful, as shown in Figure \ref{fig:results-showcase}.

% In Section \ref{sec:bias-ib-lb}, we quantify both informativeness bias and the commonly discussed length (or verbosity) bias \cite{zheng2023judging, gu2024survey}. 
In Section \ref{sec:bias-ib-lb}, we further show that informativeness bias is much stronger than the commonly discussed length bias \cite{zheng2023judging, gu2024survey} and remains significant even after controlling for length. This suggests that, although more informative answers tend to be longer, the two biases are not equivalent. Consequently, prior strategies for reducing length bias, such as instructing models to ignore length \cite{zheng2023judging} or equalizing response lengths \cite{hu2025explaining}, are insufficient, since informativeness bias cannot be resolved by addressing length effects alone.

In summary, informativeness bias is a key factor that pulls VLM-as-a-Judge away from image content, undermining reliability. As a result, their judging ability lags behind their vision-language understanding, leaving much of that capability underused. To bridge this gap and make VLMs more reliable for evaluation, we propose a novel judging paradigm \textbf{\OurMethod} (\OurMethodLong). The core idea is for the judge to first create a truthful "anchor" that encodes the necessary image content into text. Candidate answers are then compared against this anchor, bringing the judge's attention back to the image and reducing informativeness bias. To avoid over-prioritizing correctness and unfairly rejecting valid details, the anchor is made as informative as the candidates. We achieve this by asking the judge to revise the candidate answers using the query and image, removing unsupported claims with explanations and fixing inconsistencies. The resulting anchor is both accurate and informative, allowing every detail in the candidates to be verified against it, as described in Section \ref{sec:method}.

% However, prior work has rarely revealed this informativeness bias, mainly because models that focus only on textual informativeness can still achieve good performance, as many judgments can be made correctly without considering the image. Figures \ref{fig:intro} (B) show an example where the question asks about the properties of the object in the image. Both answers identify the object as a basketball. Answer B only mentions its shape, while Answer A adds details about color and size. Since “basketball” is already identified in the answers, these details can be validated as correct without using the image, allowing the judge to correctly prefer Answer A without relying on visual content. Despite this, informativeness bias remains a critical issue that undermines the reliability of VLM-as-a-Judge in perceiving image alignment.

% We conduct a comprehensive evaluation of 8 state-of-the-art judging models, including open-source systems such as LLaMA-3.2 Vision Instruct \cite{dubey2024llama} and commercial ones like GPT-4o \cite{hurst2024gpt}, 
We conduct a comprehensive evaluation of 8 state-of-the-art judging models, covering both open-source and commercial systems, on two benchmarks. \OurMethod\ consistently achieves better performance than all baselines. We also confirm that \OurMethod\ substantially reduces informativeness bias. As shown in Figure \ref{fig:results-showcase}, it greatly improves accuracy when the more informative answer is untruthful, demonstrating a stronger focus on image consistency, which drives the performance gains.

Our contributions can be summarized as follows: 

\begin{enumerate}[noitemsep, topsep=2pt]
    \item We identify \emph{informativeness bias}, where VLM-as-a-Judge is distracted by the informativeness of answers and pays little attention to image content. This leaves much of its vision-language capability underused for judgment tasks, thereby undermining reliability (Section \ref{sec:bias-blind}, \ref{sec:bias-ib}, and \ref{sec:bias-blind-ib}).
    \item We show that informativeness bias is distinct from and cannot be reduced to the commonly discussed length bias (Section \ref{sec:bias-ib-lb}), and thus calls for a solution that goes beyond simply removing length effects.
    \item We propose \OurMethod, a novel judging paradigm that mitigates informativeness bias and refocuses the judge's attention on the image, enabling better use of VLM capabilities and improving overall performance (Section \ref{sec:method}).

\end{enumerate}

\section{Related Work}

\subsection{LLM/VLM-as-a-Judge Benchmarks}

Developing effective LLM/VLM-as-a-Judge systems is essential for evaluating and aligning LLMs and VLMs \cite{ouyang2022training, dubey2024llama}. Existing benchmarks such as Anthropic Helpful and Harmless \cite{bai2022training}, OpenAI Summarization \cite{stiennon2020learning}, RewardBench \cite{lambert2025rewardbench}, and JudgeBench \cite{tan2024judgebench} have significantly advanced evaluation in the text modality (LLMs). 

To extend this progress to the multimodal domain, MLLM-as-a-Judge \cite{chen2024mllm} takes the first step in quantifying VLMs' performance as judges. VL-RewardBench \cite{li2025vl} targets scenarios where even state-of-the-art models struggle. Multimodal RewardBench \cite{yasunaga2025multimodal} incorporates expert knowledge domains and safety-related concerns. Judge Anything \cite{pu2025judge} expands this scope to multimodal generation tasks, incorporating both video and audio inputs. MM-RLHF \cite{zhangmm} improves size, diversity, annotation granularity, and includes safety-related tasks. However, despite this progress, existing works still lack a critical examination of the reliability of VLM-as-a-Judge behavior.

\subsection{Biases in LLM/VLM-as-a-Judge}

Recent work has found several cognitive biases that affect how LLMs evaluate \cite{gu2024survey}. Many studies \cite{zheng2023judging, shi2024judging} have discussed biases such as position, verbosity, and self-enhancement. Others \cite{koo2024benchmarking} highlight compassion fade, bandwagon effect, and attentional bias. Additional work \cite{chen2024humans, stureborg2024large} has noted fallacy-oversight, authority, and beauty bias. \citet{ye2025justice} propose an automated pipeline to measure diverse biases, while \citet{wei2025systematic} analyze biases across different prompt templates.

However, this topic remains underexplored in the multimodal domain. While modality bias, where multimodal LLMs  over-rely on language \cite{zheng2025mllmsdeeplyaffectedmodality, li2024naturalbench, liu2024paying}, is well known, its impact on judging behavior remains unclear. The most related work, \citet{lidevil}, shows that over-relying on text-only patterns during training hurts the generalization of multimodal reward models. Our work goes further by linking this issue to over-prioritizing informativeness.

\subsection{Debiasing in LLM/VLM-as-a-Judge}

To reduce bias in LLM-as-a-Judge, studies have asked judges to follow predefined rules \cite{zeng2024evaluating}, provided reference answers to avoid context confusion \cite{zheng2023judging, zhang2025reviseval}, swapped response positions to reduce positional bias \cite{wang2024large}, or equalized answer length to reduce length bias \cite{hu2025explaining}. However, these methods are not designed to handle informativeness bias and are therefore ineffective for this issue.

On the multimodal side, \citet{lidevil} reduces bias from textual spurious correlations by shifting the training distribution toward better multimodal understanding. Our work instead identifies informativeness bias as a cause and addresses it by enforcing attention to image consistency without additional training.

\section{Analysis Setup}

\subsection{Pairwise Comparison in VLM-as-a-Judge} 

We formalize VLM-as-a-Judge using a pairwise preference comparison paradigm \cite{chen2024mllm}. Let $\mathcal{D} \coloneq \{ d_i \}_{i=1}^N$ be a dataset with $N$ preference examples. Each example $d_i = (Q_i, I_i, y^c_i, y^r_i)$ contains a question about an image, the image itself, and two responses labeled by human preference. Here, $Q_i$ is the textual query, $I_i$ is the associated image, $y^c_i$ is the preferred response, and $y^r_i$ is the less preferred one. For simplicity, we drop the subscript $i$ when the context is clear.

The judge model $f(Q,I,y^c,y^r)$ takes a sample as input and decides which answer better addresses the question. Its performance is measured by accuracy against human preferences:

\vspace{-1.5em}

$$
\text{Acc} \;=\; \mathbb{E}_{(Q,I,y^c,y^r) \sim \mathcal{D}}
\Big [ \mathbb{I} \big ( f(Q,I,y^c,y^r) = y^c \big ) \Big ].
$$

\vspace{-1em}

\subsection{Quantifying Image Reliance}

VLMs are known to underuse visual content and over-rely on text \cite{li2024naturalbench, fu2024blink, liu2024paying, leng2024mitigating}. We ask whether they show a similar tendency when used as judges and measure how much they rely on visual input. Specifically, we define the \emph{Image Reliance Score (IRS)} as the accuracy gap of judge model $f$ when evaluated with and without the image $I$ as input:

\vspace{-1.5em}

\begin{equation*}
\begin{split}   
\text{IRS} = \mathbb{E}_{(Q, I, y^c, y^r) \sim \mathcal{D}}
\Big [ \mathbb{I} \big ( f(Q, I, y^c, y^r) = y^c \big ) \Big ] \\
    - \mathbb{E}_{(Q, y^c, y^r) \sim \mathcal{D}}
\Big [ \mathbb{I} \big ( f(Q, y^c, y^r) = y^c \big ) \Big ].
\end{split}
\end{equation*}

\vspace{-0.3em}

A VLM judge that depends more on image content will have a higher IRS, since its accuracy improves more when the image is included.

\subsection{Judging Informativeness} \label{sec:judge-info}

General-purpose alignment aims to produce responses that balance correctness and informativeness \cite{zheng2023judging}. This is reflected in common judging prompts, which often include rubrics such as: \emph{"Your evaluation should consider factors such as accuracy, helpfulness, relevance, depth, and level of detail"} \cite{zheng2023judging, chen2024mllm}, encouraging responses to go beyond a minimal correct answer and provide additional context \cite{lin-etal-2022-truthfulqa}. Here, \emph{accuracy} corresponds to correctness, which requires using the image to detect errors or inconsistencies, while \emph{helpfulness, relevance, depth, and level of detail} correspond to informativeness, which can be judged from the answers alone. We focus on how informativeness affects VLM-as-a-Judge behavior, so we separate it from the other judging criteria.

Based on this, we define the rubrics for informativeness as \emph{helpfulness, relevance, depth, and level of detail}, with \emph{accuracy} removed from the original set. We then ask GPT-4o to decide which answer, $y^c$ or $y^r$, is more informative using these rubrics. The full prompt is in Appendix \ref{sec:appendix-prompt}. Only the textual query $Q$ and the two candidate answers $(y^c, y^r)$ are given, while the image $I$ is withheld, since informativeness does not depend on it. GPT-4o is explicitly instructed to judge only on informativeness, without considering correctness or consistency with the image. The output is $y^i$, the more informative answer, where $y^i$ is either $y^c$ or $y^r$. See Appendix \ref{sec:appendix-exp-reliable-info} for an evaluation of the reliability of GPT-4o in measuring informativeness

\subsection{Measuring Informativeness Bias} \label{sec:ib}

Several studies have shown that prioritizing informativeness can come at the cost of correctness \cite{zhou2023lima, wu2025balancing}. We define \emph{Informativeness Bias (IB)} as the degree to which a judge favors informativeness over correctness.

We split the preference dataset $\mathcal{D}$ into two subsets. The \emph{informativeness-driven subset} $\mathcal{D}_\text{IDS}$ contains cases where the human-preferred answer $y^c$ is more informative than $y^r$ (i.e., $y^c = y^i$), meaning informativeness is a valid factor. The \emph{correctness-driven subset} $\mathcal{D}_\text{CDS}$ includes cases where the preferred answer $y^c$ is less informative (i.e., $y^r = y^i$), meaning correctness is the main factor and informativeness is misleading. We compute IB as:

% calculate the accuracies $\text{Acc}_\text{IDS}$ and $\text{Acc}_\text{CDS}$ on the two subsets, and compute IB, the degree to which a judge favors informativeness over correctness, as:

\vspace{-1.5em}

\begin{equation*}
\begin{split} 
\text{IB} &= \mathbb{E}_{d \sim \mathcal{D}_\text{IDS}}
\Big [ \mathbb{I} \big ( f(d) = y^c \big ) \Big ] \quad (\text{Acc}_\text{IDS}) \\
&- \mathbb{E}_{d \sim \mathcal{D}_\text{CDS}}
\Big [ \mathbb{I} \big ( f(d) = y^c \big ) \Big ] \quad (\text{Acc}_\text{CDS}).
\end{split}
\end{equation*}

\vspace{-0.5em}

In other words, we treat human annotators as unbiased, and obviously $\text{IB} = 0$ for humans since their $\text{Acc}_\text{IDS}$ and $\text{Acc}_\text{CDS}$ are both 100\%. For an unbiased judge, IB should also be close to zero, meaning it performs equally well whether or not informativeness is misleading. A positive IB indicates the judge performs better on $\mathcal{D}_\text{IDS}$, showing a preference for informativeness. An ideal judge should achieve high $\text{Acc}_\text{IDS}$ and $\text{Acc}_\text{CDS}$ for strong overall performance, while keeping IB low to balance informativeness and correctness.

\subsection{Measuring Length Bias} \label{sec:lb}

Prior studies show that LLM-as-a-Judge often favors longer responses, known as \emph{length bias} \cite{zheng2023judging, ye2025justice}. Since informativeness is often tied to length, with more informative answers usually being longer, we also measure \emph{Length Bias (LB)} for comparison to investigate how the two biases relate.

Following the definition of informativeness bias, we split the dataset $\mathcal{D}$ into two subsets, one where the longer answer is preferred and one where the shorter answer is preferred. We then calculate LB as the difference in a judge's accuracy between them: $\text{LB} = \text{Acc}_{y^c \text{ is longer}} - \text{Acc}_{y^c \text{ is shorter}}$. The magnitude $|\text{LB}|$ shows how strongly the judge is influenced by length, while the sign shows whether it favors longer or shorter responses.

\begin{table}[t]
    \centering
    
    \resizebox{\columnwidth}{!}{
    \begin{tabular}{lcccccc}
        \toprule
        
        \multirow{2}{*}{\textbf{Models}} & 
        \multicolumn{3}{c}{\textbf{MLLM-as-a-Judge}} & 
        \multicolumn{3}{c}{\textbf{VL-RewardBench}} \\
        \cmidrule(lr){2-4} \cmidrule(lr){5-7}
        & \textbf{With Image} & \textbf{No Image} & \highlight \textbf{IRS} & \textbf{With Image} & \textbf{No Image} & \highlight \textbf{IRS} \\
        
        \midrule
        
        \textbf{GPT-4o}  & 66.45 & 65.46 & \highlight 0.99 & 61.12 & 57.92 & \highlight 3.20 \\

        \textbf{Gemini-2.5-Flash}  & 66.92 & 64.94 & \highlight 1.98 & 56.25 & 53.71 & \highlight 2.54 \\

        \midrule
        
        \textbf{Llama-3.2-V-90B} & 65.28 & 62.46 & \highlight 2.82 & 59.46 &  55.44 & \highlight 4.02 \\

        \textbf{LLaVA-1.5-13B} & 56.21 & 53.32 & \highlight 2.89 & 44.76 & 42.63 & \highlight 2.13 \\

        \textbf{LLaVA-1.6-34B} & 58.95 & 59.96 & \highlight -1.01 & 48.83 & 46.31 & \highlight 2.52 \\

        \textbf{LLaVA-OV-1.5-8B} & 65.35 & 65.48 & \highlight -0.13 & 56.01 & 54.13 & \highlight 1.88 \\
        
        \bottomrule
    \end{tabular}
    }

    % \caption{Accuracy of judges with and without image input. The image reliance score (IRS, defined as accuracy gain from adding the image compared to not using it) ranges from 0\% to 100\%, with higher values indicating stronger dependence on image input. IRS remains minimal across all models and benchmarks.}

    \caption{The image reliance score (IRS, the accuracy gain from including the image in the judge's input, higher values mean stronger image dependence) remains minimal across all models and benchmarks.}

    \label{table:irs}
\end{table}

% \begin{figure}[t]
%     \centering
%     \begin{minipage}[t]{0.48\linewidth}
%         \centering
%         \includegraphics[width=\linewidth]{images/mllmjudge-lb-ib-overall.pdf}
%         % \caption{Informativeness bias (IB) and length bias (LB) comparison on MLLM-as-a-judge benchmark.}
%         % \label{fig:mllmjudge-lb-ib-overall}
%     \end{minipage}
%     \hfill
%     \begin{minipage}[t]{0.48\linewidth}
%         \centering
%         \includegraphics[width=\linewidth]{images/vlrb-lb-ib-overall.pdf}
%         % \caption{Informativeness bias (IB) and length bias (LB) comparison on VL-RewardBench benchmark.}
%         % \label{fig:vlrb-lb-ib-overall}
%     \end{minipage}
    
%     \caption{Informativeness bias (IB) and length bias (LB) comparison on MLLM-as-a-judge (left) and VL-RewardBench (right) benchmark.}
%     \label{fig:lb-ib-overall}
% \end{figure}

\begin{figure}[t]
    \centering
    \includegraphics[width=\linewidth]{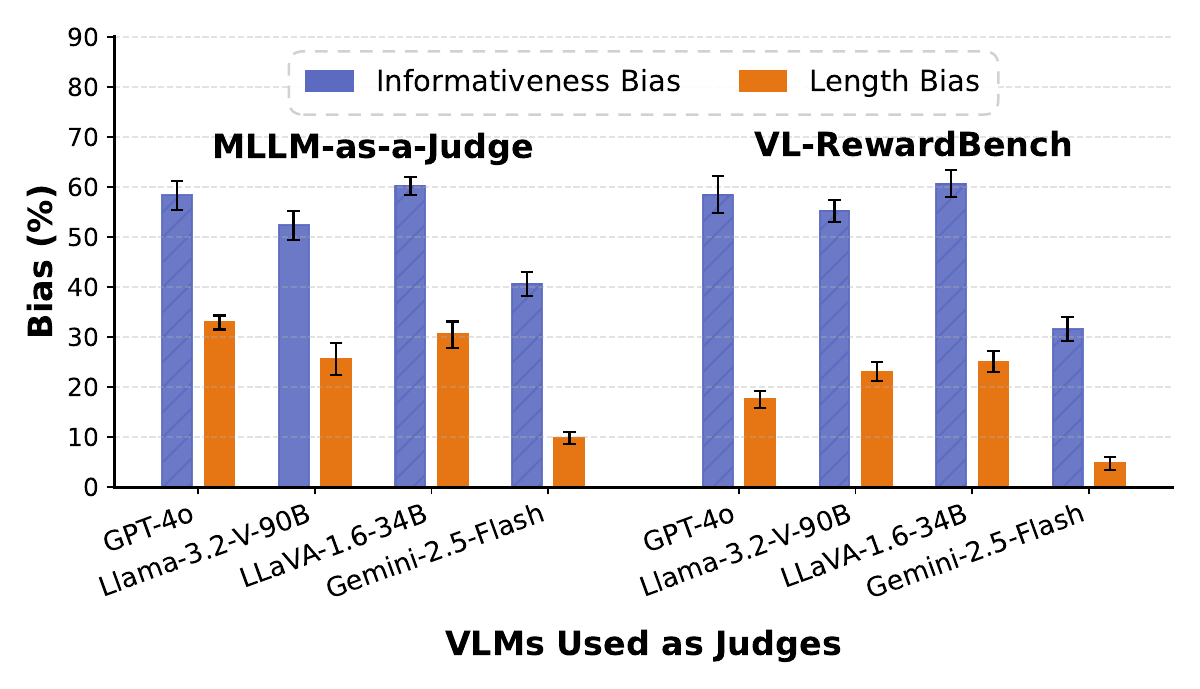}

    \caption{
    % Informativeness bias (IB) and length bias (LB) across different models and benchmarks. 
    % The mean and variance are computed by downsampling the subsets at different ratios, each sampled multiple times. 
    Informativeness bias is consistently stronger than length bias across all models and benchmarks, with both showing low variance, indicating informativeness has a greater impact on the judge's behavior than length.}
    
    \label{fig:lb-ib-overall}
\end{figure}

\begin{figure}[t]
    \centering
    \includegraphics[width=\columnwidth]{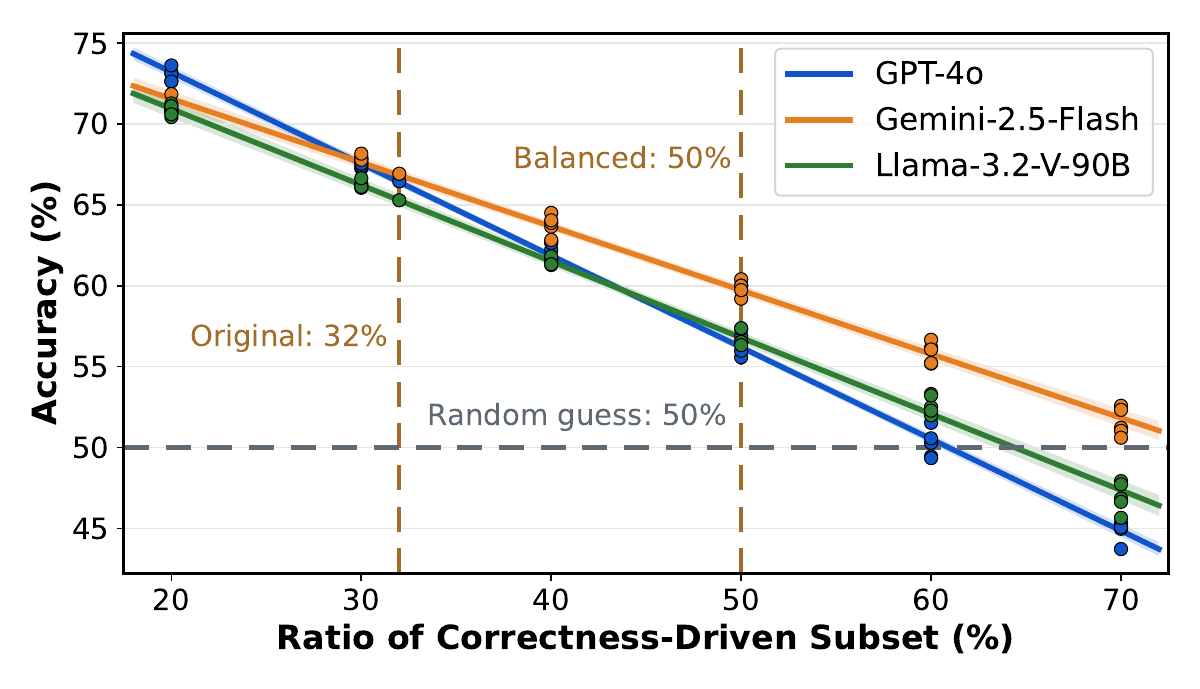}

    \caption{Judge performance is strongly affected by how often informativeness is misleading, i.e., the ratio of correctness-driven subset ($\mathcal{D}_\text{CDS}$). We control this ratio in MLLM-as-a-Judge by downsampling. Judges achieve good accuracy (>65\%) at the original low ratio (32\%), but when the data is balanced (50\%), their accuracy drops to only slightly above random guessing.}

    \label{fig:acc-vs-ratio}
\end{figure}

\begin{figure}[t]
  \includegraphics[width=\columnwidth]{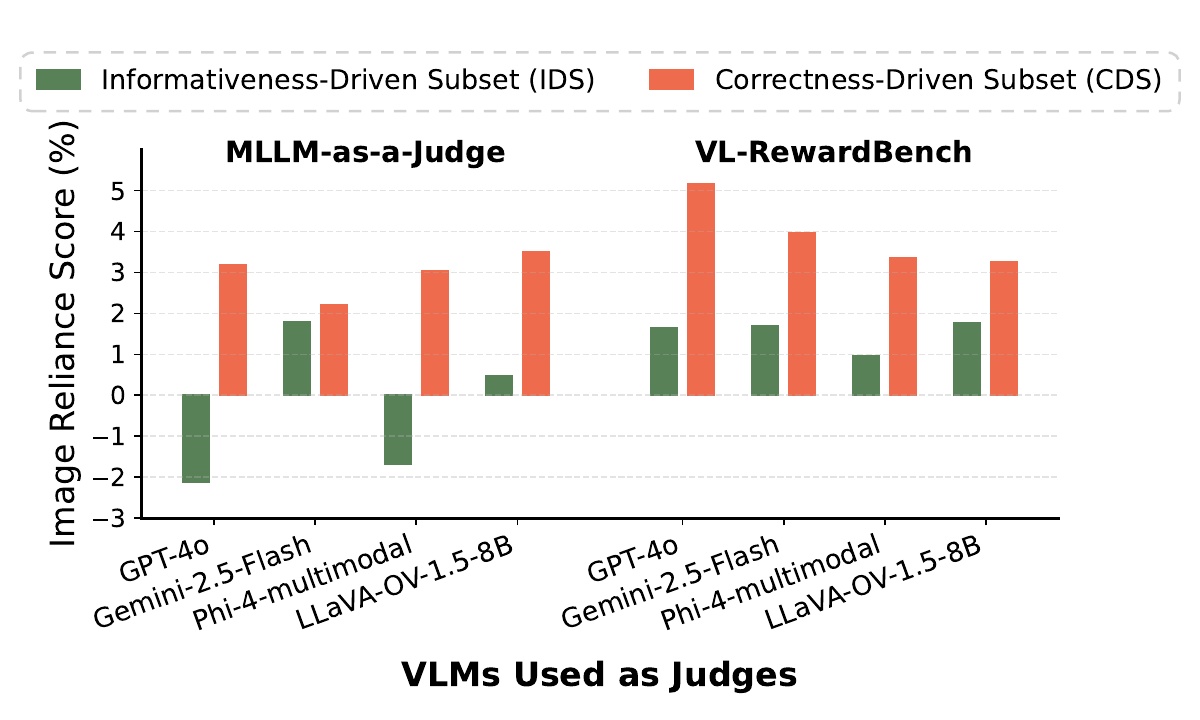}

  \caption{Image reliance score is consistently higher on correctness-driven subsets than on informativeness-driven subsets across all models and benchmarks, showing that images matter more when informativeness is misleading and correctness drives the decision.}

  \label{fig:irs-ids-cds-overall}
\end{figure}

\begin{figure}[t]
    \includegraphics[width=\columnwidth]{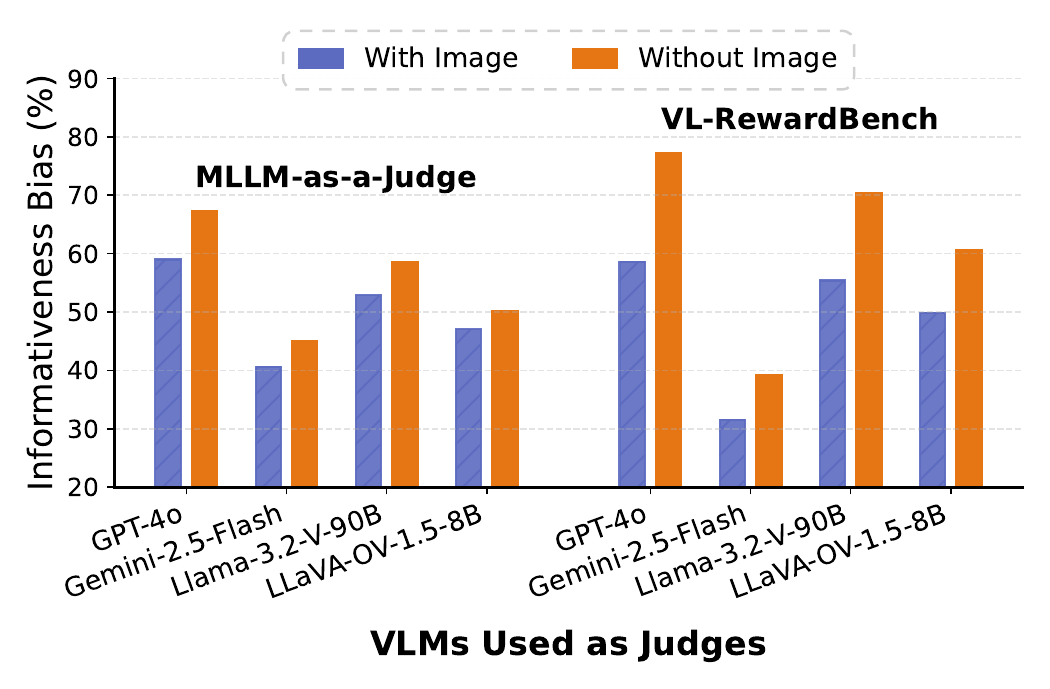}
    
    \caption{Informativeness bias is higher when the image is removed than when it is included in the input.}
    \label{fig:ib-w-wo-image}
    
\end{figure}

\section{Exposing Informativeness Bias} \label{sec:bias}

\subsection{Judges Are Nearly Blind to Images} \label{sec:bias-blind}

Table \ref{table:irs} shows that the Image Reliance Score (IRS) is minimal across all models and benchmarks, meaning image content provides only negligible gains (less than 3\% on MLLM-as-a-Judge \cite{chen2024mllm} and under 5\% on VL-RewardBench \cite{li2025vl}). We further divide MLLM-as-a-Judge into different source domains and show in Figure \ref{fig:with-no-img-compare} that adding the image yields only slight accuracy improvements across all domains, and in some cases even reduces performance. This suggests that VLM-as-a-Judge often ignores the image when evaluating answers to VQA questions, raising concerns about their reliability and highlighting the need for deeper investigation of this issue.

\subsection{Judges Are Deceived by Informativeness} \label{sec:bias-ib}

\noindent \textbf{Judges over-prioritize informativeness.} As shown in Figure \ref{fig:lb-ib-overall}, we find significant informativeness bias (IB): over 30\% for Gemini-2.5-Flash and over 50\% for other models. We also report IB variance by downsampling $\mathcal{D}_\text{IDS}$ and $\mathcal{D}_\text{CDS}$ at different ratios (see Appendix \ref{sec:appendix-exp-downsampling}), showing reasonably low variance. The high IB is also clear in Figure \ref{fig:ida-cda}, where $\text{Acc}_\text{IDS}$ is always much higher than $\text{Acc}_\text{CDS}$. These indicate that VLM judges perform far worse when informativeness is misleading, revealing their strong tendency to overvalue informativeness.

\vspace{5pt}

\noindent \textbf{Judge performance heavily depends on how often informativeness is misleading.} We downsample to control the ratio of the correctness-driven subset $\mathcal{D}_\text{CDS}$ and examine how performance changes with it. A higher ratio means informativeness is misleading more often. As shown in Figure \ref{fig:acc-vs-ratio}, judge accuracies decrease as the proportion of $\mathcal{D}_\text{CDS}$ grows. Since the proportion of $\mathcal{D}_\text{CDS}$ is low in the original benchmark (32\%), judges can appear strong overall (>65\%) even with poor $\text{Acc}_\text{CDS}$ (Figure \ref{fig:ida-cda}). But with a balanced ratio (50\%), which better reflects their real performance, overall accuracy drops to around 55\%, only slightly above random guessing. This reveals a serious and previously overlooked unreliability in VLM-as-a-Judge.

\vspace{5pt}

\noindent \textbf{Judges are misled by informativeness, even when they know the correct answer.} We show that poor $\text{Acc}_\text{CDS}$ is not due to a lack of vision-language ability in the judge. As illustrated in Figure \ref{fig:intro} (D) and additional examples in Appendix \ref{sec:appendix-case}, when directly prompted to answer the question $Q$, the judge often produces a correct response that aligns with the human-chosen answer and points out inconsistencies or errors in the rejected one. Yet, when acting as a judge, it still selects the human-rejected answer as the better choice, contradicting its own response. A common pattern is that the chosen answer appears more informative, offering detailed explanations or direct answers, but without considering correctness.

These findings show that judges can understand questions and spot conflicts between answers and the image. However, they are deceived by informativeness, giving it far more weight than correctness.

\subsection{Informativeness Bias Pulls Judge's Attention Away from the Image} \label{sec:bias-blind-ib}

From Section \ref{sec:bias-blind} and \ref{sec:bias-ib}, we hypothesize that informativeness bias (IB) distracts the judge's attention from the image, leading to a low image reliance score (IRS). We now verify this connection.

\vspace{5pt}

\noindent \textbf{Image reliance is higher on correctness-driven cases $\mathcal{D}_\text{CDS}$.} As shown in Figure \ref{fig:irs-ids-cds-overall}, IRS is consistently higher on correctness-driven cases $\mathcal{D}_\text{CDS}$ than on informativeness-driven cases $\mathcal{D}_\text{IDS}$ across all models and benchmarks, indicating that judges rely more on images when correctness drives the decision. This supports our hypothesis: when informativeness is misleading, correctness becomes crucial, and as correctness cannot be verified without the image, visual input is far more valuable.

\vspace{5pt}

\noindent \textbf{Informativeness bias worsens when the judge lacks image input.} We test how removing the image from the judge's input affects IB. As shown in Figure \ref{fig:ib-w-wo-image}, judges without images consistently show stronger IB than those with images, with gaps of up to 18.6\%. This also supports our claim: without the image, the judge cannot verify correctness and is forced to rely only on informativeness.

\subsection{Informativeness Bias Is Distinct from Length Bias} \label{sec:bias-ib-lb}

\begin{table}[t]
    \centering
    
    \resizebox{\linewidth}{!}{
    \begin{tabular}{lcc}
        \toprule
        
        \textbf{Models} & \textbf{MLLM-as-a-Judge} & \textbf{VL-RewardBench} \\
        
        \midrule

        \textbf{GPT-4o}  & \textcolor{gray}{59.0 $\to$} 44.6 & \textcolor{gray}{58.6 $\to$} 45.5 \\

        \textbf{Gemini-2.5-Flash}  & \textcolor{gray}{40.6 $\to$} 33.66 & \textcolor{gray}{31.6 $\to$} 26.0 \\

        \textbf{Llama-3.2-Vision-90B} & \textcolor{gray}{52.9 $\to$} 40.2 & \textcolor{gray}{55.4 $\to$} 39.1 \\

        \textbf{LLaVA-1.6-34B} & \textcolor{gray}{60.2 $\to$} 36.7 & \textcolor{gray}{60.6 $\to$} 42.3 \\
        
        \bottomrule
    \end{tabular}
    }

    \caption{Informativeness bias \textcolor{gray}{before $\to$} after equalizing answer length to remove length bias. Although they decrease after eliminating length bias, the values remain high, indicating that it is the root cause of length bias.}

    \label{table:ib-after-eliminating-lb}
\end{table}

Figure \ref{fig:lb-ib-overall} compares informativeness bias (IB) with length bias (LB). LB is consistently much weaker, trailing IB by about 30\% or more across all models and benchmarks. Figure \ref{fig:mllmjudge-lb-ib-per-dataset} further breaks this down and shows IB is higher than LB in nearly all domains. This suggests that IB is not merely a proxy for LB; rather, it might be a more substantial issue.

% Since more informative answers are usually longer, judges that favor informativeness may also appear to favor length, yet the two phenomena are not equivalent.

To investigate further, we remove the effect of LB by creating a version of the MLLM-as-a-Judge benchmark where the length difference between candidate answers is below a small threshold (see Appendix \ref{sec:appendix-exp-ablate-lb}). Table \ref{table:ib-after-eliminating-lb} shows IB after this length equalization. While IB decreases, it remains substantial (26\%-45.5\%). This suggests that, although IB and LB are correlated, IB cannot be resolved by addressing LB alone, as a substantial portion of IB remains after controlling for length. The correlation is expected, since more informative answers are often longer, making a preference for informativeness appear as a preference for length. However, mitigating IB requires directly addressing informativeness bias rather than focusing only on length effects \cite{zheng2023judging, hu2025explaining}.

% \subsection{Summary}

% Based on the above observations, we conclude that: (1) VLM judges show minimal reliance on image input, largely due to informativeness bias, where they blindly choose the seemingly more informative option; (2) they can still be misled by informativeness even when they are capable of detecting inconsistencies with the image; (3) current benchmarks are poorly designed, as many samples can be judged simply by selecting the more informative answer, allowing biased, image-blind judges to perform well and inflate evaluation metrics; and (4) informativeness bias is much stronger than length bias, with length bias being only a superficial effect of informativeness bias, since more informative answers are usually longer.
\begin{figure*}[t]
  \includegraphics[width=\linewidth]{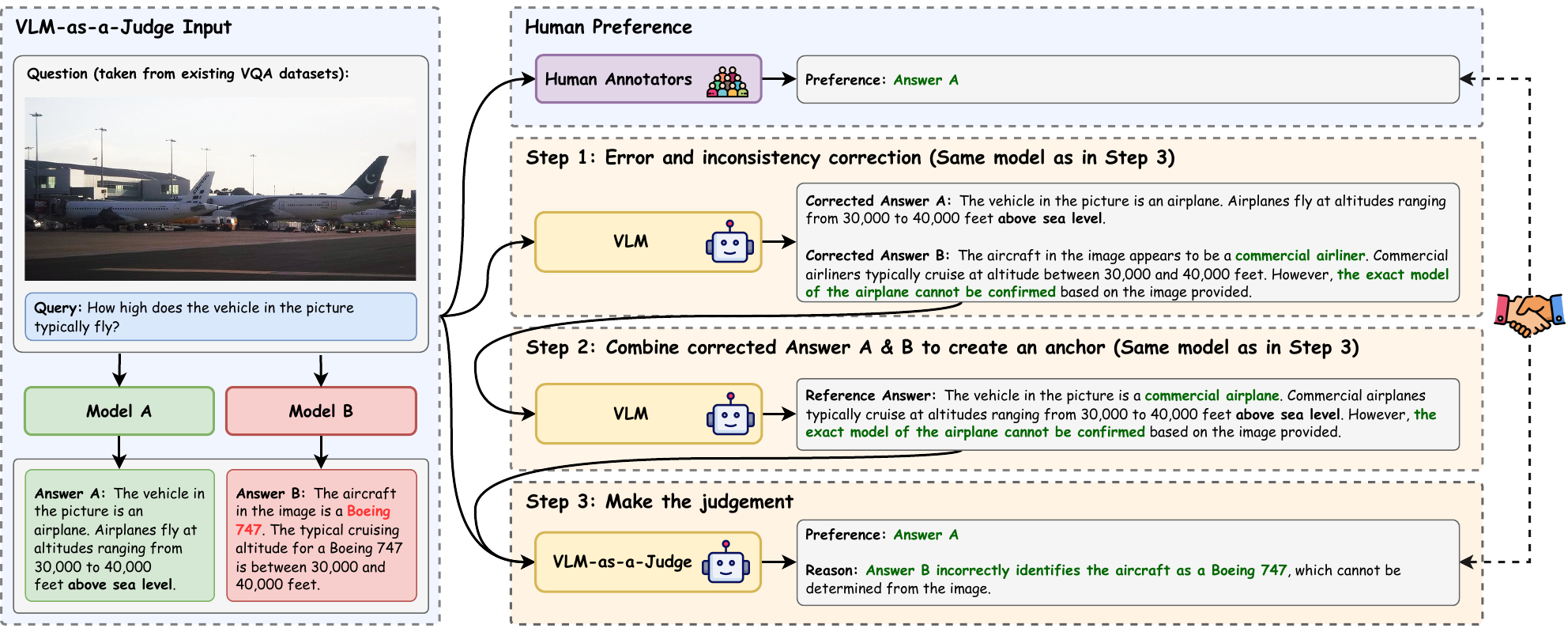}

  \caption{Illustration of the proposed \OurMethod. \textbf{Step 1}: Each candidate answer is checked and corrected based on the image. In Answer B, the unsupported detail "Boeing 747" is replaced with "a commercial airliner" and clarified as "the exact model cannot be confirmed". \textbf{Step 2}: The corrected answers are merged into an anchor that is both truthful and informative (all details from the candidates are either corrected or preserved). \textbf{Step 3}: Candidates are compared with the anchor. This forces the judge to focus on the image content encoded in the anchor, balance informativeness with correctness, and avoid both over-prioritizing and over-penalizing informativeness.}

  \label{fig:method}
\end{figure*}

\section{Method} \label{sec:method}

Based on Section \ref{sec:bias}, VLM-as-a-Judge can answer questions correctly but gets distracted by informativeness and fails to focus on the image. Therefore, we explore whether using its own answer as a reference can refocus attention on image consistency. Unlike prior work that improves LLM-as-a-Judge by providing ground-truth references \cite{zheng2023judging, hagos2024recent, zhou2025graders}, we instead use a self-generated reference.

\subsection{Reducing Informativeness Bias} \label{sec:method-base}

Figure \ref{fig:standard-ref-bad} illustrates the \texttt{Standard Ref} pipeline. The model first answers the question directly, encoding key image details relevant to the question into a textual reference. It then judges by comparing the two candidate answers to this reference. This naturally shifts the focus from informativeness alone to consistency with the image, since the reference already includes image details.

Figure \ref{fig:standard-ref-bad} (A) shows a successful case: with its own answer as reference, the judge correctly detects that Answer B gives a park name not visible in the image and selects Answer A instead. Table \ref{table:acc-ida-cda-ib-before-after} confirms that this greatly reduces informativeness bias (IB) and improves $\text{Acc}_\text{CDS}$, boosting performance when correctness should drive the decision.

However, this approach can also cause judges to over-prioritize correctness and unfairly reject valid details in candidates simply because they are missing from the reference. In Figure \ref{fig:standard-ref-bad} (B), Answer A is rejected just for mentioning the correct position of a sign omitted in the reference. Table \ref{table:acc-ida-cda-ib-before-after} shows a clear drop in $\text{Acc}_\text{IDS}$, meaning the judge performs worse when informativeness should guide the decision. This matches prior findings \cite{wu2025balancing, zhou2023lima} that greater emphasis on correctness often leads to over-caution about extra information. As a result, while this method improves $\text{Acc}_\text{CDS}$ and lowers IB, it still reduces overall accuracy compared to the base pipeline.

% \subsection{\OurMethod: Balancing Informativeness and Correctness}  \label{sec:method-ours}

\subsection{Our Method: \OurMethod}  \label{sec:method-ours}

We propose \OurMethod, which balances informativeness and correctness using an "anchor" that is both truthful and as informative as the candidates. As shown in Figure \ref{fig:method}, the judge first modifies each candidate answer by correcting inconsistencies with the image or removing unsupported claims with explanations. The corrected answers are then merged into a single anchor. This ensures the generated anchor encodes the truthful version of all relevant image details included in the candidates. As a result, when candidates are compared to the anchor, every detail can be verified against it, preserving informativeness while enforcing correctness.

\begin{table*}[t]
    \centering
    
    \resizebox{\linewidth}{!}{
    \begin{tabular}{lcccccccc}
        \toprule
        
        \multirow{2}{*}{\textbf{Models}} & 
        \multicolumn{4}{c}{\textbf{MLLM-as-a-Judge} \cite{chen2024mllm}} & 
        \multicolumn{4}{c}{\textbf{VL-RewardBench} \cite{li2025vl}} \\
        \cmidrule(lr){2-5} \cmidrule(lr){6-9}
        & \textbf{Base} & \textbf{Image Caption} & \textbf{Standard Ref} & \highlight \textbf{\OurMethod} & \textbf{Base} &  \textbf{Image Caption} & \textbf{Standard Ref} & \highlight \textbf{\OurMethod} \\
        
        \midrule
        \multicolumn{9}{c}{Proprietary Models} \\
        \midrule
        
        \textbf{GPT-4o}  & 66.45 & 65.79 & 65.14 & \highlight \textbf{75.78} & 61.12 & 59.36 & 60.42 & \highlight \textbf{66.95} \\

        \textbf{Gemini-2.5-Flash} & 66.92 & 66.90 & 68.34 & \highlight \textbf{73.37} & 56.25 & 56.65 & 57.84 & \highlight \textbf{61.08} \\

        \textbf{Gemini-2.5-Pro} & 67.63 & 68.04 & 69.21 & \highlight \textbf{73.37} & 57.55 & 57.33 & 59.21 & \highlight \textbf{61.49} \\

        \midrule
        \multicolumn{9}{c}{Open-Source Models} \\
        \midrule

        \textbf{Llama-3.2-Vision-90B} & 65.28 & 66.73 & 63.18 &  \highlight \textbf{75.12} & 59.46 & 59.94 & 59.21 & \highlight \textbf{64.52} \\

        \textbf{LLaVA-OneVision-1.5-8B} &  65.35 &  65.67 & 66.21 & \highlight \textbf{71.17} & 56.01 & 57.07 & 57.95 & \highlight \textbf{60.01} \\

        \textbf{LLaVA-1.5-13B} & 56.21 & 54.69 & 53.28 & \highlight \textbf{59.48} & 44.76 & 46.24 & 46.59 & \highlight \textbf{47.32} \\

        \textbf{LLaVA-1.6-34B} & 58.95 & 58.88 & 58.02 & \highlight \textbf{61.54} & 48.83 & 49.29 & 50.19 & \highlight \textbf{52.03} \\

        \textbf{Phi-4-multimodal} &  59.19 & 59.07 & 59.11 & \highlight \textbf{63.11} & 54.72 & 54.49 & 55.45 & \highlight \textbf{57.45} \\
        
        \bottomrule
    \end{tabular}
    }

    \caption{Accuracy of VLM-as-a-Judge. Our proposed \OurMethod\ significantly improves the performance of both open-source and proprietary models across different benchmarks.}

    \label{table:main}
\end{table*}

\begin{table}[t]
    \centering
    
    \resizebox{\linewidth}{!}{
    \begin{tabular}{lccccc}
        \toprule
        
        \textbf{Methods} & \textbf{Acc} & \textbf{$\text{Acc}_\text{IDS}$} & \textbf{$\text{Acc}_\text{CDS}$} & \textbf{IB} & \textbf{LB} \\
        
        \midrule

        \textbf{GPT-4o (Base)}  & 66.5 & 85.3 & 26.3 & 59.0 & 32.9 \\

        \textbf{\; + Standard Ref} & 65.1 \textcolor{BrickRed}{(-1.4)} & 80.7 \textcolor{BrickRed}{(-4.6)} & 31.8 \textcolor{PineGreen}{(+5.5)}  & 48.9 \textcolor{PineGreen}{(-10.1)} & 16.9 \textcolor{PineGreen}{(-16.0)} \\

        \textbf{\; + \OurMethod\ (Ours)} & 75.8 \textcolor{PineGreen}{(+9.3)} & 90.1 \textcolor{PineGreen}{(+4.8)} & 45.3 \textcolor{PineGreen}{(+19)} & 44.8 \textcolor{PineGreen}{(-14.2)} & 20.3 \textcolor{PineGreen}{(-12.6)} \\
        
        \bottomrule
    \end{tabular}
    }

    % \caption{Overall accuracy, accuracy on informativeness-driven cases $\text{Acc}_\text{IDS}$ and correctness-driven cases $\text{Acc}_\text{CDS}$, and informativeness bias (IB) before and after applying \texttt{Standard Ref} and \texttt{\OurMethod} on MLLM-as-a-Judge \cite{chen2024mllm} benchmark.}

    \caption{Changes across different metrics on MLLM-as-a-Judge \cite{chen2024mllm}. Our \OurMethod\ reduces informativeness bias (IB) by significantly improving $\text{Acc}_\text{CDS}$ while keeping strong $\text{Acc}_\text{IDS}$, explaining its strong overall performance.}

    \label{table:acc-ida-cda-ib-before-after}
\end{table}

\begin{table}[t]
    \centering
        
    \resizebox{.9\linewidth}{!}{
    
    \begin{tabular}{lc}
        \toprule
        
        \textbf{Text} & \textbf{Average Score} \\
        
        \midrule

        Answer 1 & 3.1034 \\
        Answer 2 & 3.2126 \\
        Anchor (Gemini-2.5-Flash) & 4.3793 \\
        Anchor (LLaVA-1.5-13B) & 3.7364 \\
        
        \bottomrule
    \end{tabular}
    }

    \caption{Comparison of GPT-5 quality scores for generated anchors and candidate answers, using inputs from MLLM-as-a-Judge \cite{chen2024mllm}.}

    \label{table:anchor-quality}
\end{table}

\begin{table}[t]
    \centering
        
    \resizebox{.9\linewidth}{!}{
    
    \begin{tabular}{lcc}
        \toprule
        
        \textbf{Method} & Base & \OurMethod \\
        
        \midrule

        \textbf{Inference Time (s)} & 3.87 & 7.47 \textcolor{BrickRed}{(+3.6)} \\
        
        \bottomrule
    \end{tabular}
    }

    \caption{Inference time comparison per data point.}

    \label{table:inference-time}
\end{table}

\section{Experiments} \label{sec:exp}

\subsection{Evaluation Settings}

We evaluate our approach on MLLM-as-a-Judge \cite{chen2024mllm} and VL-RewardBench \cite{li2025vl} benchmarks. We use a pairwise comparison setup \cite{chen2024mllm, li2025vl} to measure agreement between model and human judgments.
% During evaluation, we found labeling errors that reduced the reliability of the benchmarks. We manually corrected these errors and conducted evaluations on the revised data, as described in Appendix \ref{sec:appendix-bench}.
We manually corrected human labeling errors that reduced benchmark reliability and evaluated on the revised data (see Appendix \ref{sec:appendix-bench}).

% Our experiments cover open-source models from 5.6B to 90B parameters, 
Our experiments include open-source models such as Phi-4-multimodal \cite{abdin2024phi4technicalreport}, LLaVA-OneVision-1.5-8B \cite{li2025llavaonevision}, LLaVA-1.5-13B, LLaVA-1.6-34B \cite{liu2024improved}, and Llama-3.2-Vision-90B \cite{dubey2024llama}, as well as commercial models like GPT-4o \cite{hurst2024gpt}, Gemini-2.5-Flash, and Gemini-2.5-Pro \cite{comanici2025gemini25pushingfrontier}.

\subsection{Evaluation of \OurMethod}

\noindent \textbf{Main Results.} As shown in Table \ref{table:main}, \texttt{\OurMethod} achieves the best performance across all tested models and benchmarks, with improvements over \texttt{Base} (a standard judge pipeline following \cite{chen2024mllm} with no extra input) ranging from 2.73\% to 9.84\%. Improvements are larger on stronger models like GPT-4o, Gemini, and Llama-3.2. Although these models are affected by informativeness bias, their stronger vision ability helps them detect and correct errors in candidate answers and generate more reliable anchors. Gains are smaller for weaker models such as LLaVA-1.5/1.6, which lack strong vision understanding, and for Phi-4-Multimodal, which shows less informativeness bias to begin with (see Appendix \ref{sec:appendix-exp-phi4}), leaving less room for improvement. 

\vspace{3pt}

\noindent \textbf{Ablation Studies.} \texttt{\OurMethod} introduces an anchor generation module that corrects and merges candidate answers to assist the judge. To evaluate this design, Table \ref{table:main} compares it with two alternatives: (1) \texttt{Image Caption}, which generates an image caption and uses it with the query, image, and candidate answers for judging, as illustrated in Figure \ref{fig:method-image-caption}, and (2) \texttt{Standard Ref}, which produces a reference answer without using the candidates and then compares the candidates against it, as described in Section \ref{sec:method-base}. \texttt{\OurMethod} outperforms all baselines. \texttt{Image Caption} and \texttt{Standard Ref} sometimes slightly outperform \texttt{Base}, but in some cases they perform worse than \texttt{Base}.

% \subsection{\OurMethod\ Balances Informativeness and Correctness}

\subsection{Why \OurMethod\ Works Well?}

\noindent \textbf{Quantitative Results.} We analyze changes across multiple metrics with and without \texttt{\OurMethod} in Table \ref{table:ib-after-eliminating-lb} and \ref{table:acc-ida-cda-ib-lb-full}. \texttt{\OurMethod} shows a much lower IB than both \texttt{Base} and \texttt{Standard Ref} across all models. 
% Table \ref{table:ib-after-eliminating-lb} also reports $\text{Acc}_\text{IDS}$ and $\text{Acc}_\text{CDS}$ to show performance on informativeness-driven and correctness-driven cases. 
For GPT-4o, \texttt{\OurMethod} boosts $\text{Acc}_\text{CDS}$ by 19\% over \texttt{Base}, much higher than the gain from \texttt{Standard Ref}, and also raises $\text{Acc}_\text{IDS}$ by 4.8\% at the same time, resulting in a 9.3\% overall accuracy gain. In contrast, \texttt{Standard Ref} causes a 4.6\% drop in $\text{Acc}_\text{IDS}$, which finally lowers overall accuracy. These results show that \texttt{\OurMethod} maintains a good balance without over-prioritizing or over-penalizing informativeness. Moreover, \texttt{\OurMethod} also clearly reduces length bias (LB), consistent with our finding that IB and LB are not independent and that mitigating IB also alleviates LB.

\vspace{3pt}

\noindent \textbf{Case study.} We provide qualitative examples in Figure \ref{fig:method}, \ref{fig:standard-ref-bad}, \ref{fig:method-image-caption} and Appendix \ref{sec:appendix-case} to illustrate how \texttt{\OurMethod} balances informativeness and correctness, and why \texttt{Standard Ref} and \texttt{Image Caption} fail. \texttt{Base} is easily misled by a more informative but incorrect answer, even when it clearly conflicts with the judge’s own response. \texttt{\OurMethod} creates an anchor by correcting or removing inconsistencies while preserving valid details. This allows the judge to verify every detail in the candidates and make correct judgments. In contrast, \texttt{Standard Ref} and \texttt{Image Caption} generate references without using the candidate answers and often miss details mentioned in them. As a result, \texttt{Standard Ref} makes the judge overly cautious and reject correct details that are not in the reference, while \texttt{Image Caption} accepts unsupported details when they cannot be verified against the caption. \texttt{\OurMethod} reduces both incorrect rejection and acceptance and achieves a better balance.

% In Figure \ref{fig:method}, the question asks, "How high does the vehicle in the picture typically fly?" Both answers correctly state "30,000 to 40,000 feet", but Answer B incorrectly identifies the aircraft as a Boeing 747, even though the image does not provide enough evidence to determine the model. GPT-4o nevertheless chooses Answer B because it includes extra information about the airplane type, revealing its Informativeness Bias. With \OurMethod, the reference answer explicitly states that "the specific model of the airplane cannot be determined from the visual details provided", which helps GPT-4o reject Answer A and select the correct one.

% In Figure \ref{fig:standard-ref-bad}, the question asks, "What does the red sign mean?" Both answers say "stop sign", but Answer A adds location information: "The signs are placed at the beginning of the street", which is correct. The human-preferred answer is therefore Answer A, as it is both correct and more informative. However, the standard reference answer omits the location detail, leading GPT-4o to wrongly penalize Answer A with the explanation: "Answer A introduces speculative reasoning about the positions of the signs without support from the image". With \OurMethod, the reference answer also includes the location information, allowing GPT-4o to verify it and select the correct answer.

\subsection{Anchor Quality}

The effectiveness of \texttt{\OurMethod} depends on the quality of the generated anchors. These anchors should contain fewer factual errors than the original candidate answers to help the judge verify image details. To evaluate this, we use GPT-5 to score the correctness of the anchors and the two original candidate answers for Gemini-2.5-Flash and LLaVA-1.5-13B. As shown in Table \ref{table:anchor-quality}, the anchors achieve higher average scores than both original answers, indicating that VLMs can correct many image inconsistencies and produce more accurate anchors.

\subsection{Inference Effencify}

Although anchor generation in \texttt{\OurMethod} involves two steps, answer correction and merging, they are implemented in a single pass, as shown in Table \ref{table:birch-ref-prompt}. Therefore, \texttt{\OurMethod} requires only one additional pass compared to the plain judge pipeline \texttt{Base}.

We further illustrate this by reporting the average inference time per data point in seconds for LLaVA-One-Vision-1.5-8B on a single A100 GPU for \texttt{\OurMethod} and \texttt{Base} in Table \ref{table:inference-time}. The inference time of \texttt{\OurMethod} is about twice that of \texttt{Base}, which is expected since it performs one extra pass. This overhead is comparable to or lower than many LLM-as-a-judge methods \cite{tang-etal-2024-metrics, pmlr-v235-khan24a} that require multiple extra calls.

\section{Conclusion}

% In this paper, we reveal and address a critical challenge in VLM-as-a-Judge: Informativeness Bias, which undermines their ability to judge vision-language responses. Our analysis shows that even when a VLM detects misalignment between a candidate answer and the image, it often still prefers the more informative answer without properly considering the image. 

We reveal and address a fundamental flaw in current VLM-as-a-Judge systems, namely \emph{informativeness bias}, the tendency to blindly favor more informative answers while giving little attention to the image. To improve reliability, we propose \OurMethod, which compares candidate answers against a self-generated anchor that corrects or verifies all image details mentioned. This shifts the judge's focus back to image content, ensuring every detail is properly verified and thus balancing informativeness with correctness. \OurMethod\ substantially reduces informativeness bias and delivers consistent gains across multiple VLMs and benchmarks.

\section{Limitations}

Our work analyzes the issues that undermine the reliability of VLM-as-a-Judge and proposes \OurMethod\ to reduce informativeness bias and refocus on image content for more truthful judgments. \OurMethod\ is simple to implement and requires no extra training.

However, informativeness bias still remains after applying \OurMethod\ (Table \ref{table:main}), which shows that it does not fully resolve the problem. Future work could address the following issues. \OurMethod\ builds an anchor by correcting candidate answers to encode key image details and guide the judge toward correctness. However, the model can still be influenced by informativeness during this process and fail to remove or fix incorrect claims, which reduces the quality of the anchor. Besides, even when \OurMethod\ generates a reliable and accurate anchor, the judge may still favor a more informative but incorrect answer (Appendix \ref{sec:appendix-case-failure}).

In addition, \OurMethod\ is designed to boost performance by reducing informativeness bias. However, for models like Phi-4-Multimodal, where the bias is already small (Section \ref{sec:appendix-exp-phi4}), the gains are limited (Table \ref{table:main}). Future work should explore ways to further strengthen judging ability in such models.

\section{Ethical Considerations}

In this paper, we introduce \OurMethod, a new judging framework designed to make VLM-as-a-Judge more reliable and better aligned with human preferences. However, human annotators themselves are subject to bias, which in turn makes aligning VLM-as-a-Judge with their preferences also biased. We leave mitigating such biases in both benchmarks and VLM-as-a-Judge to future work.

\bibliography{custom}

@inproceedings{ouyang2022training,
  author       = {Long Ouyang and
                  Jeffrey Wu and
                  Xu Jiang and
                  Diogo Almeida and
                  Carroll L. Wainwright and
                  Pamela Mishkin and
                  Chong Zhang and
                  Sandhini Agarwal and
                  Katarina Slama and
                  Alex Ray and
                  John Schulman and
                  Jacob Hilton and
                  Fraser Kelton and
                  Luke Miller and
                  Maddie Simens and
                  Amanda Askell and
                  Peter Welinder and
                  Paul F. Christiano and
                  Jan Leike and
                  Ryan Lowe},
  editor       = {Sanmi Koyejo and
                  S. Mohamed and
                  A. Agarwal and
                  Danielle Belgrave and
                  K. Cho and
                  A. Oh},
  title        = {Training language models to follow instructions with human feedback},
  booktitle    = {Advances in Neural Information Processing Systems 35: Annual Conference
                  on Neural Information Processing Systems 2022, NeurIPS 2022, New Orleans,
                  LA, USA, November 28 - December 9, 2022},
  year         = {2022},
  url          = {http://papers.nips.cc/paper\_files/paper/2022/hash/b1efde53be364a73914f58805a001731-Abstract-Conference.html},
  bibsource    = {dblp computer science bibliography, https://dblp.org}
}

@misc{dubey2024llama,
      title={The Llama 3 Herd of Models}, 
      author={Aaron Grattafiori and Abhimanyu Dubey and Abhinav Jauhri and Abhinav Pandey and Abhishek Kadian and Ahmad Al-Dahle and Aiesha Letman and Akhil Mathur and Alan Schelten and Alex Vaughan and Amy Yang and Angela Fan and Anirudh Goyal and Anthony Hartshorn and Aobo Yang and Archi Mitra and Archie Sravankumar and Artem Korenev and Arthur Hinsvark and Arun Rao and Aston Zhang and Aurelien Rodriguez and Austen Gregerson and Ava Spataru and Baptiste Roziere and Bethany Biron and Binh Tang and Bobbie Chern and Charlotte Caucheteux and Chaya Nayak and Chloe Bi and Chris Marra and Chris McConnell and Christian Keller and Christophe Touret and Chunyang Wu and Corinne Wong and Cristian Canton Ferrer and Cyrus Nikolaidis and Damien Allonsius and Daniel Song and Danielle Pintz and Danny Livshits and Danny Wyatt and David Esiobu and Dhruv Choudhary and Dhruv Mahajan and Diego Garcia-Olano and Diego Perino and Dieuwke Hupkes and Egor Lakomkin and Ehab AlBadawy and Elina Lobanova and Emily Dinan and Eric Michael Smith and Filip Radenovic and Francisco Guzmán and Frank Zhang and Gabriel Synnaeve and Gabrielle Lee and Georgia Lewis Anderson and Govind Thattai and Graeme Nail and Gregoire Mialon and Guan Pang and Guillem Cucurell and Hailey Nguyen and Hannah Korevaar and Hu Xu and Hugo Touvron and Iliyan Zarov and Imanol Arrieta Ibarra and Isabel Kloumann and Ishan Misra and Ivan Evtimov and Jack Zhang and Jade Copet and Jaewon Lee and Jan Geffert and Jana Vranes and Jason Park and Jay Mahadeokar and Jeet Shah and Jelmer van der Linde and Jennifer Billock and Jenny Hong and Jenya Lee and Jeremy Fu and Jianfeng Chi and Jianyu Huang and Jiawen Liu and Jie Wang and Jiecao Yu and Joanna Bitton and Joe Spisak and Jongsoo Park and Joseph Rocca and Joshua Johnstun and Joshua Saxe and Junteng Jia and Kalyan Vasuden Alwala and Karthik Prasad and Kartikeya Upasani and Kate Plawiak and Ke Li and Kenneth Heafield and Kevin Stone and Khalid El-Arini and Krithika Iyer and Kshitiz Malik and Kuenley Chiu and Kunal Bhalla and Kushal Lakhotia and Lauren Rantala-Yeary and Laurens van der Maaten and Lawrence Chen and Liang Tan and Liz Jenkins and Louis Martin and Lovish Madaan and Lubo Malo and Lukas Blecher and Lukas Landzaat and Luke de Oliveira and Madeline Muzzi and Mahesh Pasupuleti and Mannat Singh and Manohar Paluri and Marcin Kardas and Maria Tsimpoukelli and Mathew Oldham and Mathieu Rita and Maya Pavlova and Melanie Kambadur and Mike Lewis and Min Si and Mitesh Kumar Singh and Mona Hassan and Naman Goyal and Narjes Torabi and Nikolay Bashlykov and Nikolay Bogoychev and Niladri Chatterji and Ning Zhang and Olivier Duchenne and Onur Çelebi and Patrick Alrassy and Pengchuan Zhang and Pengwei Li and Petar Vasic and Peter Weng and Prajjwal Bhargava and Pratik Dubal and Praveen Krishnan and Punit Singh Koura and Puxin Xu and Qing He and Qingxiao Dong and Ragavan Srinivasan and Raj Ganapathy and Ramon Calderer and Ricardo Silveira Cabral and Robert Stojnic and Roberta Raileanu and Rohan Maheswari and Rohit Girdhar and Rohit Patel and Romain Sauvestre and Ronnie Polidoro and Roshan Sumbaly and Ross Taylor and Ruan Silva and Rui Hou and Rui Wang and Saghar Hosseini and Sahana Chennabasappa and Sanjay Singh and Sean Bell and Seohyun Sonia Kim and Sergey Edunov and Shaoliang Nie and Sharan Narang and Sharath Raparthy and Sheng Shen and Shengye Wan and Shruti Bhosale and Shun Zhang and Simon Vandenhende and Soumya Batra and Spencer Whitman and Sten Sootla and Stephane Collot and Suchin Gururangan and Sydney Borodinsky and Tamar Herman and Tara Fowler and Tarek Sheasha and Thomas Georgiou and Thomas Scialom and Tobias Speckbacher and Todor Mihaylov and Tong Xiao and Ujjwal Karn and Vedanuj Goswami and Vibhor Gupta and Vignesh Ramanathan and Viktor Kerkez and Vincent Gonguet and Virginie Do and Vish Vogeti and Vítor Albiero and Vladan Petrovic and Weiwei Chu and Wenhan Xiong and Wenyin Fu and Whitney Meers and Xavier Martinet and Xiaodong Wang and Xiaofang Wang and Xiaoqing Ellen Tan and Xide Xia and Xinfeng Xie and Xuchao Jia and Xuewei Wang and Yaelle Goldschlag and Yashesh Gaur and Yasmine Babaei and Yi Wen and Yiwen Song and Yuchen Zhang and Yue Li and Yuning Mao and Zacharie Delpierre Coudert and Zheng Yan and Zhengxing Chen and Zoe Papakipos and Aaditya Singh and Aayushi Srivastava and Abha Jain and Adam Kelsey and Adam Shajnfeld and Adithya Gangidi and Adolfo Victoria and Ahuva Goldstand and Ajay Menon and Ajay Sharma and Alex Boesenberg and Alexei Baevski and Allie Feinstein and Amanda Kallet and Amit Sangani and Amos Teo and Anam Yunus and Andrei Lupu and Andres Alvarado and Andrew Caples and Andrew Gu and Andrew Ho and Andrew Poulton and Andrew Ryan and Ankit Ramchandani and Annie Dong and Annie Franco and Anuj Goyal and Aparajita Saraf and Arkabandhu Chowdhury and Ashley Gabriel and Ashwin Bharambe and Assaf Eisenman and Azadeh Yazdan and Beau James and Ben Maurer and Benjamin Leonhardi and Bernie Huang and Beth Loyd and Beto De Paola and Bhargavi Paranjape and Bing Liu and Bo Wu and Boyu Ni and Braden Hancock and Bram Wasti and Brandon Spence and Brani Stojkovic and Brian Gamido and Britt Montalvo and Carl Parker and Carly Burton and Catalina Mejia and Ce Liu and Changhan Wang and Changkyu Kim and Chao Zhou and Chester Hu and Ching-Hsiang Chu and Chris Cai and Chris Tindal and Christoph Feichtenhofer and Cynthia Gao and Damon Civin and Dana Beaty and Daniel Kreymer and Daniel Li and David Adkins and David Xu and Davide Testuggine and Delia David and Devi Parikh and Diana Liskovich and Didem Foss and Dingkang Wang and Duc Le and Dustin Holland and Edward Dowling and Eissa Jamil and Elaine Montgomery and Eleonora Presani and Emily Hahn and Emily Wood and Eric-Tuan Le and Erik Brinkman and Esteban Arcaute and Evan Dunbar and Evan Smothers and Fei Sun and Felix Kreuk and Feng Tian and Filippos Kokkinos and Firat Ozgenel and Francesco Caggioni and Frank Kanayet and Frank Seide and Gabriela Medina Florez and Gabriella Schwarz and Gada Badeer and Georgia Swee and Gil Halpern and Grant Herman and Grigory Sizov and Guangyi and Zhang and Guna Lakshminarayanan and Hakan Inan and Hamid Shojanazeri and Han Zou and Hannah Wang and Hanwen Zha and Haroun Habeeb and Harrison Rudolph and Helen Suk and Henry Aspegren and Hunter Goldman and Hongyuan Zhan and Ibrahim Damlaj and Igor Molybog and Igor Tufanov and Ilias Leontiadis and Irina-Elena Veliche and Itai Gat and Jake Weissman and James Geboski and James Kohli and Janice Lam and Japhet Asher and Jean-Baptiste Gaya and Jeff Marcus and Jeff Tang and Jennifer Chan and Jenny Zhen and Jeremy Reizenstein and Jeremy Teboul and Jessica Zhong and Jian Jin and Jingyi Yang and Joe Cummings and Jon Carvill and Jon Shepard and Jonathan McPhie and Jonathan Torres and Josh Ginsburg and Junjie Wang and Kai Wu and Kam Hou U and Karan Saxena and Kartikay Khandelwal and Katayoun Zand and Kathy Matosich and Kaushik Veeraraghavan and Kelly Michelena and Keqian Li and Kiran Jagadeesh and Kun Huang and Kunal Chawla and Kyle Huang and Lailin Chen and Lakshya Garg and Lavender A and Leandro Silva and Lee Bell and Lei Zhang and Liangpeng Guo and Licheng Yu and Liron Moshkovich and Luca Wehrstedt and Madian Khabsa and Manav Avalani and Manish Bhatt and Martynas Mankus and Matan Hasson and Matthew Lennie and Matthias Reso and Maxim Groshev and Maxim Naumov and Maya Lathi and Meghan Keneally and Miao Liu and Michael L. Seltzer and Michal Valko and Michelle Restrepo and Mihir Patel and Mik Vyatskov and Mikayel Samvelyan and Mike Clark and Mike Macey and Mike Wang and Miquel Jubert Hermoso and Mo Metanat and Mohammad Rastegari and Munish Bansal and Nandhini Santhanam and Natascha Parks and Natasha White and Navyata Bawa and Nayan Singhal and Nick Egebo and Nicolas Usunier and Nikhil Mehta and Nikolay Pavlovich Laptev and Ning Dong and Norman Cheng and Oleg Chernoguz and Olivia Hart and Omkar Salpekar and Ozlem Kalinli and Parkin Kent and Parth Parekh and Paul Saab and Pavan Balaji and Pedro Rittner and Philip Bontrager and Pierre Roux and Piotr Dollar and Polina Zvyagina and Prashant Ratanchandani and Pritish Yuvraj and Qian Liang and Rachad Alao and Rachel Rodriguez and Rafi Ayub and Raghotham Murthy and Raghu Nayani and Rahul Mitra and Rangaprabhu Parthasarathy and Raymond Li and Rebekkah Hogan and Robin Battey and Rocky Wang and Russ Howes and Ruty Rinott and Sachin Mehta and Sachin Siby and Sai Jayesh Bondu and Samyak Datta and Sara Chugh and Sara Hunt and Sargun Dhillon and Sasha Sidorov and Satadru Pan and Saurabh Mahajan and Saurabh Verma and Seiji Yamamoto and Sharadh Ramaswamy and Shaun Lindsay and Shaun Lindsay and Sheng Feng and Shenghao Lin and Shengxin Cindy Zha and Shishir Patil and Shiva Shankar and Shuqiang Zhang and Shuqiang Zhang and Sinong Wang and Sneha Agarwal and Soji Sajuyigbe and Soumith Chintala and Stephanie Max and Stephen Chen and Steve Kehoe and Steve Satterfield and Sudarshan Govindaprasad and Sumit Gupta and Summer Deng and Sungmin Cho and Sunny Virk and Suraj Subramanian and Sy Choudhury and Sydney Goldman and Tal Remez and Tamar Glaser and Tamara Best and Thilo Koehler and Thomas Robinson and Tianhe Li and Tianjun Zhang and Tim Matthews and Timothy Chou and Tzook Shaked and Varun Vontimitta and Victoria Ajayi and Victoria Montanez and Vijai Mohan and Vinay Satish Kumar and Vishal Mangla and Vlad Ionescu and Vlad Poenaru and Vlad Tiberiu Mihailescu and Vladimir Ivanov and Wei Li and Wenchen Wang and Wenwen Jiang and Wes Bouaziz and Will Constable and Xiaocheng Tang and Xiaojian Wu and Xiaolan Wang and Xilun Wu and Xinbo Gao and Yaniv Kleinman and Yanjun Chen and Ye Hu and Ye Jia and Ye Qi and Yenda Li and Yilin Zhang and Ying Zhang and Yossi Adi and Youngjin Nam and Yu and Wang and Yu Zhao and Yuchen Hao and Yundi Qian and Yunlu Li and Yuzi He and Zach Rait and Zachary DeVito and Zef Rosnbrick and Zhaoduo Wen and Zhenyu Yang and Zhiwei Zhao and Zhiyu Ma},
      year={2024},
      eprint={2407.21783},
      archivePrefix={arXiv},
      primaryClass={cs.AI},
      url={https://arxiv.org/abs/2407.21783}, 
}

@misc{bai2022training,
      title={Training a Helpful and Harmless Assistant with Reinforcement Learning from Human Feedback}, 
      author={Yuntao Bai and Andy Jones and Kamal Ndousse and Amanda Askell and Anna Chen and Nova DasSarma and Dawn Drain and Stanislav Fort and Deep Ganguli and Tom Henighan and Nicholas Joseph and Saurav Kadavath and Jackson Kernion and Tom Conerly and Sheer El-Showk and Nelson Elhage and Zac Hatfield-Dodds and Danny Hernandez and Tristan Hume and Scott Johnston and Shauna Kravec and Liane Lovitt and Neel Nanda and Catherine Olsson and Dario Amodei and Tom Brown and Jack Clark and Sam McCandlish and Chris Olah and Ben Mann and Jared Kaplan},
      year={2022},
      eprint={2204.05862},
      archivePrefix={arXiv},
      primaryClass={cs.CL},
      url={https://arxiv.org/abs/2204.05862}, 
}

@inproceedings{stiennon2020learning,
  author       = {Nisan Stiennon and
                  Long Ouyang and
                  Jeffrey Wu and
                  Daniel M. Ziegler and
                  Ryan Lowe and
                  Chelsea Voss and
                  Alec Radford and
                  Dario Amodei and
                  Paul F. Christiano},
  editor       = {Hugo Larochelle and
                  Marc'Aurelio Ranzato and
                  Raia Hadsell and
                  Maria{-}Florina Balcan and
                  Hsuan{-}Tien Lin},
  title        = {Learning to summarize with human feedback},
  booktitle    = {Advances in Neural Information Processing Systems 33: Annual Conference
                  on Neural Information Processing Systems 2020, NeurIPS 2020, December
                  6-12, 2020, virtual},
  year         = {2020},
  url          = {https://proceedings.neurips.cc/paper/2020/hash/1f89885d556929e98d3ef9b86448f951-Abstract.html},
  bibsource    = {dblp computer science bibliography, https://dblp.org}
}

@inproceedings{lambert2025rewardbench,
    title = "{R}eward{B}ench: Evaluating Reward Models for Language Modeling",
    author = "Lambert, Nathan  and
      Pyatkin, Valentina  and
      Morrison, Jacob  and
      Miranda, LJ  and
      Lin, Bill Yuchen  and
      Chandu, Khyathi  and
      Dziri, Nouha  and
      Kumar, Sachin  and
      Zick, Tom  and
      Choi, Yejin  and
      Smith, Noah A.  and
      Hajishirzi, Hannaneh",
    editor = "Chiruzzo, Luis  and
      Ritter, Alan  and
      Wang, Lu",
    booktitle = "Findings of the Association for Computational Linguistics: NAACL 2025",
    month = apr,
    year = "2025",
    address = "Albuquerque, New Mexico",
    publisher = "Association for Computational Linguistics",
    url = "https://aclanthology.org/2025.findings-naacl.96/",
    doi = "10.18653/v1/2025.findings-naacl.96",
    pages = "1755--1797",
    ISBN = "979-8-89176-195-7"
}

@inproceedings{
tan2024judgebench,
title={JudgeBench: A Benchmark for Evaluating {LLM}-Based Judges},
author={Sijun Tan and Siyuan Zhuang and Kyle Montgomery and William Yuan Tang and Alejandro Cuadron and Chenguang Wang and Raluca Popa and Ion Stoica},
booktitle={The Thirteenth International Conference on Learning Representations},
year={2025},
url={https://openreview.net/forum?id=G0dksFayVq}
}

@inproceedings{zheng2023judging,
  author       = {Lianmin Zheng and
                  Wei{-}Lin Chiang and
                  Ying Sheng and
                  Siyuan Zhuang and
                  Zhanghao Wu and
                  Yonghao Zhuang and
                  Zi Lin and
                  Zhuohan Li and
                  Dacheng Li and
                  Eric P. Xing and
                  Hao Zhang and
                  Joseph E. Gonzalez and
                  Ion Stoica},
  editor       = {Alice Oh and
                  Tristan Naumann and
                  Amir Globerson and
                  Kate Saenko and
                  Moritz Hardt and
                  Sergey Levine},
  title        = {Judging LLM-as-a-Judge with MT-Bench and Chatbot Arena},
  booktitle    = {Advances in Neural Information Processing Systems 36: Annual Conference
                  on Neural Information Processing Systems 2023, NeurIPS 2023, New Orleans,
                  LA, USA, December 10 - 16, 2023},
  year         = {2023},
  url          = {http://papers.nips.cc/paper\_files/paper/2023/hash/91f18a1287b398d378ef22505bf41832-Abstract-Datasets\_and\_Benchmarks.html},
  bibsource    = {dblp computer science bibliography, https://dblp.org}
}

@misc{shi2024judging,
      title={Judging the Judges: A Systematic Study of Position Bias in LLM-as-a-Judge}, 
      author={Lin Shi and Chiyu Ma and Wenhua Liang and Xingjian Diao and Weicheng Ma and Soroush Vosoughi},
      year={2025},
      eprint={2406.07791},
      archivePrefix={arXiv},
      primaryClass={cs.CL},
      url={https://arxiv.org/abs/2406.07791}, 
}

@misc{gu2024survey,
      title={A Survey on LLM-as-a-Judge}, 
      author={Jiawei Gu and Xuhui Jiang and Zhichao Shi and Hexiang Tan and Xuehao Zhai and Chengjin Xu and Wei Li and Yinghan Shen and Shengjie Ma and Honghao Liu and Saizhuo Wang and Kun Zhang and Yuanzhuo Wang and Wen Gao and Lionel Ni and Jian Guo},
      year={2025},
      eprint={2411.15594},
      archivePrefix={arXiv},
      primaryClass={cs.CL},
      url={https://arxiv.org/abs/2411.15594}, 
}

@inproceedings{koo2024benchmarking,
    title = "Benchmarking Cognitive Biases in Large Language Models as Evaluators",
    author = "Koo, Ryan  and
      Lee, Minhwa  and
      Raheja, Vipul  and
      Park, Jong Inn  and
      Kim, Zae Myung  and
      Kang, Dongyeop",
    editor = "Ku, Lun-Wei  and
      Martins, Andre  and
      Srikumar, Vivek",
    booktitle = "Findings of the Association for Computational Linguistics: ACL 2024",
    month = aug,
    year = "2024",
    address = "Bangkok, Thailand",
    publisher = "Association for Computational Linguistics",
    url = "https://aclanthology.org/2024.findings-acl.29/",
    doi = "10.18653/v1/2024.findings-acl.29",
    pages = "517--545"
}

@misc{chen2024humans,
      title={Humans or LLMs as the Judge? A Study on Judgement Biases}, 
      author={Guiming Hardy Chen and Shunian Chen and Ziche Liu and Feng Jiang and Benyou Wang},
      year={2024},
      eprint={2402.10669},
      archivePrefix={arXiv},
      primaryClass={cs.CL},
      url={https://arxiv.org/abs/2402.10669}, 
}

@misc{stureborg2024large,
      title={Large Language Models are Inconsistent and Biased Evaluators}, 
      author={Rickard Stureborg and Dimitris Alikaniotis and Yoshi Suhara},
      year={2024},
      eprint={2405.01724},
      archivePrefix={arXiv},
      primaryClass={cs.CL},
      url={https://arxiv.org/abs/2405.01724}, 
}

@misc{hurst2024gpt,
      title={GPT-4o System Card}, 
      author={OpenAI and : and Aaron Hurst and Adam Lerer and Adam P. Goucher and Adam Perelman and Aditya Ramesh and Aidan Clark and AJ Ostrow and Akila Welihinda and Alan Hayes and Alec Radford and Aleksander Mądry and Alex Baker-Whitcomb and Alex Beutel and Alex Borzunov and Alex Carney and Alex Chow and Alex Kirillov and Alex Nichol and Alex Paino and Alex Renzin and Alex Tachard Passos and Alexander Kirillov and Alexi Christakis and Alexis Conneau and Ali Kamali and Allan Jabri and Allison Moyer and Allison Tam and Amadou Crookes and Amin Tootoochian and Amin Tootoonchian and Ananya Kumar and Andrea Vallone and Andrej Karpathy and Andrew Braunstein and Andrew Cann and Andrew Codispoti and Andrew Galu and Andrew Kondrich and Andrew Tulloch and Andrey Mishchenko and Angela Baek and Angela Jiang and Antoine Pelisse and Antonia Woodford and Anuj Gosalia and Arka Dhar and Ashley Pantuliano and Avi Nayak and Avital Oliver and Barret Zoph and Behrooz Ghorbani and Ben Leimberger and Ben Rossen and Ben Sokolowsky and Ben Wang and Benjamin Zweig and Beth Hoover and Blake Samic and Bob McGrew and Bobby Spero and Bogo Giertler and Bowen Cheng and Brad Lightcap and Brandon Walkin and Brendan Quinn and Brian Guarraci and Brian Hsu and Bright Kellogg and Brydon Eastman and Camillo Lugaresi and Carroll Wainwright and Cary Bassin and Cary Hudson and Casey Chu and Chad Nelson and Chak Li and Chan Jun Shern and Channing Conger and Charlotte Barette and Chelsea Voss and Chen Ding and Cheng Lu and Chong Zhang and Chris Beaumont and Chris Hallacy and Chris Koch and Christian Gibson and Christina Kim and Christine Choi and Christine McLeavey and Christopher Hesse and Claudia Fischer and Clemens Winter and Coley Czarnecki and Colin Jarvis and Colin Wei and Constantin Koumouzelis and Dane Sherburn and Daniel Kappler and Daniel Levin and Daniel Levy and David Carr and David Farhi and David Mely and David Robinson and David Sasaki and Denny Jin and Dev Valladares and Dimitris Tsipras and Doug Li and Duc Phong Nguyen and Duncan Findlay and Edede Oiwoh and Edmund Wong and Ehsan Asdar and Elizabeth Proehl and Elizabeth Yang and Eric Antonow and Eric Kramer and Eric Peterson and Eric Sigler and Eric Wallace and Eugene Brevdo and Evan Mays and Farzad Khorasani and Felipe Petroski Such and Filippo Raso and Francis Zhang and Fred von Lohmann and Freddie Sulit and Gabriel Goh and Gene Oden and Geoff Salmon and Giulio Starace and Greg Brockman and Hadi Salman and Haiming Bao and Haitang Hu and Hannah Wong and Haoyu Wang and Heather Schmidt and Heather Whitney and Heewoo Jun and Hendrik Kirchner and Henrique Ponde de Oliveira Pinto and Hongyu Ren and Huiwen Chang and Hyung Won Chung and Ian Kivlichan and Ian O'Connell and Ian O'Connell and Ian Osband and Ian Silber and Ian Sohl and Ibrahim Okuyucu and Ikai Lan and Ilya Kostrikov and Ilya Sutskever and Ingmar Kanitscheider and Ishaan Gulrajani and Jacob Coxon and Jacob Menick and Jakub Pachocki and James Aung and James Betker and James Crooks and James Lennon and Jamie Kiros and Jan Leike and Jane Park and Jason Kwon and Jason Phang and Jason Teplitz and Jason Wei and Jason Wolfe and Jay Chen and Jeff Harris and Jenia Varavva and Jessica Gan Lee and Jessica Shieh and Ji Lin and Jiahui Yu and Jiayi Weng and Jie Tang and Jieqi Yu and Joanne Jang and Joaquin Quinonero Candela and Joe Beutler and Joe Landers and Joel Parish and Johannes Heidecke and John Schulman and Jonathan Lachman and Jonathan McKay and Jonathan Uesato and Jonathan Ward and Jong Wook Kim and Joost Huizinga and Jordan Sitkin and Jos Kraaijeveld and Josh Gross and Josh Kaplan and Josh Snyder and Joshua Achiam and Joy Jiao and Joyce Lee and Juntang Zhuang and Justyn Harriman and Kai Fricke and Kai Hayashi and Karan Singhal and Katy Shi and Kavin Karthik and Kayla Wood and Kendra Rimbach and Kenny Hsu and Kenny Nguyen and Keren Gu-Lemberg and Kevin Button and Kevin Liu and Kiel Howe and Krithika Muthukumar and Kyle Luther and Lama Ahmad and Larry Kai and Lauren Itow and Lauren Workman and Leher Pathak and Leo Chen and Li Jing and Lia Guy and Liam Fedus and Liang Zhou and Lien Mamitsuka and Lilian Weng and Lindsay McCallum and Lindsey Held and Long Ouyang and Louis Feuvrier and Lu Zhang and Lukas Kondraciuk and Lukasz Kaiser and Luke Hewitt and Luke Metz and Lyric Doshi and Mada Aflak and Maddie Simens and Madelaine Boyd and Madeleine Thompson and Marat Dukhan and Mark Chen and Mark Gray and Mark Hudnall and Marvin Zhang and Marwan Aljubeh and Mateusz Litwin and Matthew Zeng and Max Johnson and Maya Shetty and Mayank Gupta and Meghan Shah and Mehmet Yatbaz and Meng Jia Yang and Mengchao Zhong and Mia Glaese and Mianna Chen and Michael Janner and Michael Lampe and Michael Petrov and Michael Wu and Michele Wang and Michelle Fradin and Michelle Pokrass and Miguel Castro and Miguel Oom Temudo de Castro and Mikhail Pavlov and Miles Brundage and Miles Wang and Minal Khan and Mira Murati and Mo Bavarian and Molly Lin and Murat Yesildal and Nacho Soto and Natalia Gimelshein and Natalie Cone and Natalie Staudacher and Natalie Summers and Natan LaFontaine and Neil Chowdhury and Nick Ryder and Nick Stathas and Nick Turley and Nik Tezak and Niko Felix and Nithanth Kudige and Nitish Keskar and Noah Deutsch and Noel Bundick and Nora Puckett and Ofir Nachum and Ola Okelola and Oleg Boiko and Oleg Murk and Oliver Jaffe and Olivia Watkins and Olivier Godement and Owen Campbell-Moore and Patrick Chao and Paul McMillan and Pavel Belov and Peng Su and Peter Bak and Peter Bakkum and Peter Deng and Peter Dolan and Peter Hoeschele and Peter Welinder and Phil Tillet and Philip Pronin and Philippe Tillet and Prafulla Dhariwal and Qiming Yuan and Rachel Dias and Rachel Lim and Rahul Arora and Rajan Troll and Randall Lin and Rapha Gontijo Lopes and Raul Puri and Reah Miyara and Reimar Leike and Renaud Gaubert and Reza Zamani and Ricky Wang and Rob Donnelly and Rob Honsby and Rocky Smith and Rohan Sahai and Rohit Ramchandani and Romain Huet and Rory Carmichael and Rowan Zellers and Roy Chen and Ruby Chen and Ruslan Nigmatullin and Ryan Cheu and Saachi Jain and Sam Altman and Sam Schoenholz and Sam Toizer and Samuel Miserendino and Sandhini Agarwal and Sara Culver and Scott Ethersmith and Scott Gray and Sean Grove and Sean Metzger and Shamez Hermani and Shantanu Jain and Shengjia Zhao and Sherwin Wu and Shino Jomoto and Shirong Wu and Shuaiqi and Xia and Sonia Phene and Spencer Papay and Srinivas Narayanan and Steve Coffey and Steve Lee and Stewart Hall and Suchir Balaji and Tal Broda and Tal Stramer and Tao Xu and Tarun Gogineni and Taya Christianson and Ted Sanders and Tejal Patwardhan and Thomas Cunninghman and Thomas Degry and Thomas Dimson and Thomas Raoux and Thomas Shadwell and Tianhao Zheng and Todd Underwood and Todor Markov and Toki Sherbakov and Tom Rubin and Tom Stasi and Tomer Kaftan and Tristan Heywood and Troy Peterson and Tyce Walters and Tyna Eloundou and Valerie Qi and Veit Moeller and Vinnie Monaco and Vishal Kuo and Vlad Fomenko and Wayne Chang and Weiyi Zheng and Wenda Zhou and Wesam Manassra and Will Sheu and Wojciech Zaremba and Yash Patil and Yilei Qian and Yongjik Kim and Youlong Cheng and Yu Zhang and Yuchen He and Yuchen Zhang and Yujia Jin and Yunxing Dai and Yury Malkov},
      year={2024},
      eprint={2410.21276},
      archivePrefix={arXiv},
      primaryClass={cs.CL},
      url={https://arxiv.org/abs/2410.21276}, 
}

@misc{zhang2021mme,
      title={MME: A Comprehensive Evaluation Benchmark for Multimodal Large Language Models}, 
      author={Chaoyou Fu and Peixian Chen and Yunhang Shen and Yulei Qin and Mengdan Zhang and Xu Lin and Jinrui Yang and Xiawu Zheng and Ke Li and Xing Sun and Yunsheng Wu and Rongrong Ji},
      year={2024},
      eprint={2306.13394},
      archivePrefix={arXiv},
      primaryClass={cs.CV},
      url={https://arxiv.org/abs/2306.13394}, 
}

@INPROCEEDINGS{yue2024mmmu,
  author={Yue, Xiang and Ni, Yuansheng and Zheng, Tianyu and Zhang, Kai and Liu, Ruoqi and Zhang, Ge and Stevens, Samuel and Jiang, Dongfu and Ren, Weiming and Sun, Yuxuan and Wei, Cong and Yu, Botao and Yuan, Ruibin and Sun, Renliang and Yin, Ming and Zheng, Boyuan and Yang, Zhenzhu and Liu, Yibo and Huang, Wenhao and Sun, Huan and Su, Yu and Chen, Wenhu},
  booktitle={2024 IEEE/CVF Conference on Computer Vision and Pattern Recognition (CVPR)}, 
  title={MMMU: A Massive Multi-Discipline Multimodal Understanding and Reasoning Benchmark for Expert AGI}, 
  year={2024},
  volume={},
  number={},
  pages={9556-9567},
  doi={10.1109/CVPR52733.2024.00913}}

@INPROCEEDINGS{li2025vl,
  author={Li, Lei and Wei, Yuancheng and Xie, Zhihui and Yang, Xuqing and Song, Yifan and Wang, Peiyi and An, Chenxin and Liu, Tianyu and Li, Sujian and Lin, Bill Yuchen and Kong, Lingpeng and Liu, Qi},
  booktitle={2025 IEEE/CVF Conference on Computer Vision and Pattern Recognition (CVPR)}, 
  title={VL-RewardBench: A Challenging Benchmark for Vision-Language Generative Reward Models}, 
  year={2025},
  volume={},
  number={},
  pages={24657-24668},
  doi={10.1109/CVPR52734.2025.02296}}

@inproceedings{
chen2024mllm,
title={{MLLM}-as-a-Judge: Assessing Multimodal {LLM}-as-a-Judge with Vision-Language Benchmark},
author={Dongping Chen and Ruoxi Chen and Shilin Zhang and Yaochen Wang and Yinuo Liu and Huichi Zhou and Qihui Zhang and Yao Wan and Pan Zhou and Lichao Sun},
booktitle={Forty-first International Conference on Machine Learning},
year={2024},
url={https://openreview.net/forum?id=dbFEFHAD79}
}

@misc{yasunaga2025multimodal,
      title={Multimodal RewardBench: Holistic Evaluation of Reward Models for Vision Language Models}, 
      author={Michihiro Yasunaga and Luke Zettlemoyer and Marjan Ghazvininejad},
      year={2025},
      eprint={2502.14191},
      archivePrefix={arXiv},
      primaryClass={cs.CV},
      url={https://arxiv.org/abs/2502.14191}, 
}

@article{
    jingfgaif,
    title={{FGAIF}: Aligning Large Vision-Language Models with Fine-grained {AI} Feedback},
    author={Liqiang Jing and Xinya Du},
    journal={Transactions on Machine Learning Research},
    issn={2835-8856},
    year={2025},
    url={https://openreview.net/forum?id=Qhfw5CUVd7},
    note={}
}

@INPROCEEDINGS{yu2024rlaif,
  author={Yu, Tianyu and Zhang, Haoye and Li, Qiming and Xu, Qixin and Yao, Yuan and Chen, Da and Lu, Xiaoman and Cui, Ganqu and Dang, Yunkai and He, Taiwen and Feng, Xiaocheng and Song, Jun and Zheng, Bo and Liu, Zhiyuan and Chua, Tat-Seng and Sun, Maosong},
  booktitle={2025 IEEE/CVF Conference on Computer Vision and Pattern Recognition (CVPR)}, 
  title={RLAIF-V: Open-Source AI Feedback Leads to Super GPT-4V Trustworthiness}, 
  year={2025},
  volume={},
  number={},
  pages={19985-19995},
  doi={10.1109/CVPR52734.2025.01861}
}

@inproceedings{zhou2024calibrated,
    author       = {Yiyang Zhou and
                  Zhiyuan Fan and
                  Dongjie Cheng and
                  Sihan Yang and
                  Zhaorun Chen and
                  Chenhang Cui and
                  Xiyao Wang and
                  Yun Li and
                  Linjun Zhang and
                  Huaxiu Yao},
    editor       = {Amir Globersons and
                  Lester Mackey and
                  Danielle Belgrave and
                  Angela Fan and
                  Ulrich Paquet and
                  Jakub M. Tomczak and
                  Cheng Zhang},
    title        = {Calibrated Self-Rewarding Vision Language Models},
    booktitle    = {Advances in Neural Information Processing Systems 38: Annual Conference
                  on Neural Information Processing Systems 2024, NeurIPS 2024, Vancouver,
                  BC, Canada, December 10 - 15, 2024},
    year         = {2024},
    url          = {http://papers.nips.cc/paper\_files/paper/2024/hash/5c20c00504e0c049ec2370d0cceaf3c4-Abstract-Conference.html},
    bibsource    = {dblp computer science bibliography, https://dblp.org}
}

@inproceedings{wang2024large,
    title = "Large Language Models are not Fair Evaluators",
    author = "Wang, Peiyi  and
      Li, Lei  and
      Chen, Liang  and
      Cai, Zefan  and
      Zhu, Dawei  and
      Lin, Binghuai  and
      Cao, Yunbo  and
      Kong, Lingpeng  and
      Liu, Qi  and
      Liu, Tianyu  and
      Sui, Zhifang",
    editor = "Ku, Lun-Wei  and
      Martins, Andre  and
      Srikumar, Vivek",
    booktitle = "Proceedings of the 62nd Annual Meeting of the Association for Computational Linguistics (Volume 1: Long Papers)",
    month = aug,
    year = "2024",
    address = "Bangkok, Thailand",
    publisher = "Association for Computational Linguistics",
    url = "https://aclanthology.org/2024.acl-long.511/",
    doi = "10.18653/v1/2024.acl-long.511",
    pages = "9440--9450"
}

@inproceedings{zhangmm,
    title={{MM}-{RLHF}: The Next Step Forward in Multimodal {LLM} Alignment},
    author={YiFan Zhang and Tao Yu and Haochen Tian and Chaoyou Fu and Peiyan Li and Jianshu Zeng and Wulin Xie and Yang Shi and Huanyu Zhang and Junkang Wu and Xue Wang and Yibo Hu and Bin Wen and Tingting Gao and Zhang Zhang and Fan Yang and Di ZHANG and Liang Wang and Rong Jin},
    booktitle={Forty-second International Conference on Machine Learning},
    year={2025},
    url={https://openreview.net/forum?id=ULJ4gJJYFp}
}

@inproceedings{pu2025judge,
    author = {Pu, Shu and Wang, Yaochen and Chen, Dongping and Chen, Yuhang and Wang, Guohao and Qin, Qi and Zhang, Zhongyi and Zhang, Zhiyuan and Zhou, Zetong and Gong, Shuang and Gui, Yi and Wan, Yao and Yu, Philip S.},
    title = {Judge Anything: MLLM as a Judge Across Any Modality},
    year = {2025},
    isbn = {9798400714542},
    publisher = {Association for Computing Machinery},
    address = {New York, NY, USA},
    url = {https://doi.org/10.1145/3711896.3737409},
    doi = {10.1145/3711896.3737409},
    booktitle = {Proceedings of the 31st ACM SIGKDD Conference on Knowledge Discovery and Data Mining V.2},
    pages = {5742–5753},
    numpages = {12},
    location = {Toronto ON, Canada},
    series = {KDD '25}
}

@inproceedings{wei2025systematic,
    title={Systematic Evaluation of {LLM}-as-a-Judge in {LLM} Alignment Tasks: Explainable Metrics and Diverse Prompt Templates},
    author={Hui Wei and Shenghua He and Tian Xia and Fei Liu and Andy Wong and Jingyang Lin and Mei Han},
    booktitle={ICLR 2025 Workshop on Building Trust in Language Models and Applications},
    year={2025},
    url={https://openreview.net/forum?id=CAgBCSt8gL}
}

@inproceedings{ye2025justice,
    title={Justice or Prejudice? Quantifying Biases in {LLM}-as-a-Judge},
    author={Jiayi Ye and Yanbo Wang and Yue Huang and Dongping Chen and Qihui Zhang and Nuno Moniz and Tian Gao and Werner Geyer and Chao Huang and Pin-Yu Chen and Nitesh V Chawla and Xiangliang Zhang},
    booktitle={The Thirteenth International Conference on Learning Representations},
    year={2025},
    url={https://openreview.net/forum?id=3GTtZFiajM}
}

@inproceedings{lidevil,
    title={The Devil Is in the Details: Tackling Unimodal Spurious Correlations for Generalizable Multimodal Reward Models},
    author={Zichao Li and Xueru Wen and Jie Lou and Yuqiu Ji and Yaojie Lu and Xianpei Han and Debing Zhang and Le Sun},
    booktitle={Forty-second International Conference on Machine Learning},
    year={2025},
    url={https://openreview.net/forum?id=b0qRSUcQP7}
}

@inproceedings{li2024naturalbench,
    author       = {Baiqi Li and
                      Zhiqiu Lin and
                      Wenxuan Peng and
                      Jean de Dieu Nyandwi and
                      Daniel Jiang and
                      Zixian Ma and
                      Simran Khanuja and
                      Ranjay Krishna and
                      Graham Neubig and
                      Deva Ramanan},
      editor       = {Amir Globersons and
                      Lester Mackey and
                      Danielle Belgrave and
                      Angela Fan and
                      Ulrich Paquet and
                      Jakub M. Tomczak and
                      Cheng Zhang},
      title        = {NaturalBench: Evaluating Vision-Language Models on Natural Adversarial
                      Samples},
      booktitle    = {Advances in Neural Information Processing Systems 38: Annual Conference
                      on Neural Information Processing Systems 2024, NeurIPS 2024, Vancouver,
                      BC, Canada, December 10 - 15, 2024},
      year         = {2024},
      url          = {http://papers.nips.cc/paper\_files/paper/2024/hash/1e69ff56d0ebff0752ff29caaddc25dd-Abstract-Datasets\_and\_Benchmarks\_Track.html},
      bibsource    = {dblp computer science bibliography, https://dblp.org}
}

@inproceedings{fu2024blink,
    author="Fu, Xingyu
    and Hu, Yushi
    and Li, Bangzheng
    and Feng, Yu
    and Wang, Haoyu
    and Lin, Xudong
    and Roth, Dan
    and Smith, Noah A.
    and Ma, Wei-Chiu
    and Krishna, Ranjay",
    editor="Leonardis, Ale{\v{s}}
    and Ricci, Elisa
    and Roth, Stefan
    and Russakovsky, Olga
    and Sattler, Torsten
    and Varol, G{\"u}l",
    title="BLINK: Multimodal Large Language Models Can See but Not Perceive",
    booktitle="Computer Vision -- ECCV 2024",
    year="2025",
    publisher="Springer Nature Switzerland",
    address="Cham",
    pages="148--166",
    isbn="978-3-031-73337-6"
}

@inproceedings{leng2024mitigating,
    author={Leng, Sicong and Zhang, Hang and Chen, Guanzheng and Li, Xin and Lu, Shijian and Miao, Chunyan and Bing, Lidong},
    booktitle={2024 IEEE/CVF Conference on Computer Vision and Pattern Recognition (CVPR)}, 
    title={Mitigating Object Hallucinations in Large Vision-Language Models through Visual Contrastive Decoding}, 
    year={2024},
    volume={},
    number={},
    pages={13872-13882},
    doi={10.1109/CVPR52733.2024.01316}}

@inproceedings{liu2024paying,
    author="Liu, Shi
    and Zheng, Kecheng
    and Chen, Wei",
    editor="Leonardis, Ale{\v{s}}
    and Ricci, Elisa
    and Roth, Stefan
    and Russakovsky, Olga
    and Sattler, Torsten
    and Varol, G{\"u}l",
    title="Paying More Attention to Image: A Training-Free Method for Alleviating Hallucination in LVLMs",
    booktitle="Computer Vision -- ECCV 2024",
    year="2025",
    publisher="Springer Nature Switzerland",
    address="Cham",
    pages="125--140",
    isbn="978-3-031-73010-8"
}

@misc{wu2025balancing,
    title={Balancing Truthfulness and Informativeness with Uncertainty-Aware Instruction Fine-Tuning}, 
    author={Tianyi Wu and Jingwei Ni and Bryan Hooi and Jiaheng Zhang and Elliott Ash and See-Kiong Ng and Mrinmaya Sachan and Markus Leippold},
    year={2025},
    eprint={2502.11962},
    archivePrefix={arXiv},
    primaryClass={cs.CL},
    url={https://arxiv.org/abs/2502.11962}, 
}

@inproceedings{
    zhou2023lima,
    title={{LIMA}: Less Is More for Alignment},
    author={Chunting Zhou and Pengfei Liu and Puxin Xu and Srini Iyer and Jiao Sun and Yuning Mao and Xuezhe Ma and Avia Efrat and Ping Yu and LILI YU and Susan Zhang and Gargi Ghosh and Mike Lewis and Luke Zettlemoyer and Omer Levy},
    booktitle={Thirty-seventh Conference on Neural Information Processing Systems},
    year={2023},
    url={https://openreview.net/forum?id=KBMOKmX2he}
}

@inproceedings{
    zhang2025reviseval,
    title={RevisEval: Improving {LLM}-as-a-Judge via Response-Adapted References},
    author={Qiyuan Zhang and Yufei Wang and Tiezheng YU and Yuxin Jiang and Chuhan Wu and Liangyou Li and Yasheng Wang and Xin Jiang and Lifeng Shang and Ruiming Tang and Fuyuan Lyu and Chen Ma},
    booktitle={The Thirteenth International Conference on Learning Representations},
    year={2025},
    url={https://openreview.net/forum?id=1tBvzOYTLF}
}

@misc{zhou2025graders,
  title={Graders should cheat: privileged information enables expert-level automated evaluations}, 
  author={Jin Peng Zhou and Sébastien M. R. Arnold and Nan Ding and Kilian Q. Weinberger and Nan Hua and Fei Sha},
  year={2025},
  eprint={2502.10961},
  archivePrefix={arXiv},
  primaryClass={cs.LG},
  url={https://arxiv.org/abs/2502.10961}
}

@ARTICLE{hagos2024recent,
  author={Hagos, Desta Haileselassie and Battle, Rick and Rawat, Danda B.},
  journal={IEEE Transactions on Artificial Intelligence}, 
  title={Recent Advances in Generative AI and Large Language Models: Current Status, Challenges, and Perspectives}, 
  year={2024},
  volume={5},
  number={12},
  pages={5873-5893},
  doi={10.1109/TAI.2024.3444742}
}

@inproceedings{
    zeng2024evaluating,
    title={Evaluating Large Language Models at Evaluating Instruction Following},
    author={Zhiyuan Zeng and Jiatong Yu and Tianyu Gao and Yu Meng and Tanya Goyal and Danqi Chen},
    booktitle={The Twelfth International Conference on Learning Representations},
    year={2024},
    url={https://openreview.net/forum?id=tr0KidwPLc}
}

@inproceedings{
    hu2025explaining,
    title={Explaining Length Bias in {LLM}-Based Preference Evaluations},
    author={Zhengyu Hu and Linxin Song and Jieyu Zhang and Zheyuan Xiao and Zhengyu Chen and Hui Xiong},
    booktitle={ICLR 2025 Workshop on Navigating and Addressing Data Problems for Foundation Models},
    year={2025},
    url={https://openreview.net/forum?id=tK2pcnNlWw}
}

@INPROCEEDINGS{liu2024improved,
  author={Liu, Haotian and Li, Chunyuan and Li, Yuheng and Lee, Yong Jae},
  booktitle={2024 IEEE/CVF Conference on Computer Vision and Pattern Recognition (CVPR)}, 
  title={Improved Baselines with Visual Instruction Tuning}, 
  year={2024},
  volume={},
  number={},
  pages={26286-26296},
  doi={10.1109/CVPR52733.2024.02484}
}

@misc{abdin2024phi4technicalreport,
      title={Phi-4 Technical Report}, 
      author={Marah Abdin and Jyoti Aneja and Harkirat Behl and Sébastien Bubeck and Ronen Eldan and Suriya Gunasekar and Michael Harrison and Russell J. Hewett and Mojan Javaheripi and Piero Kauffmann and James R. Lee and Yin Tat Lee and Yuanzhi Li and Weishung Liu and Caio C. T. Mendes and Anh Nguyen and Eric Price and Gustavo de Rosa and Olli Saarikivi and Adil Salim and Shital Shah and Xin Wang and Rachel Ward and Yue Wu and Dingli Yu and Cyril Zhang and Yi Zhang},
      year={2024},
      eprint={2412.08905},
      archivePrefix={arXiv},
      primaryClass={cs.CL},
      url={https://arxiv.org/abs/2412.08905} 
}

@article{
    li2025llavaonevision,
    title={{LL}a{VA}-OneVision: Easy Visual Task Transfer},
    author={Bo Li and Yuanhan Zhang and Dong Guo and Renrui Zhang and Feng Li and Hao Zhang and Kaichen Zhang and Peiyuan Zhang and Yanwei Li and Ziwei Liu and Chunyuan Li},
    journal={Transactions on Machine Learning Research},
    issn={2835-8856},
    year={2025},
    url={https://openreview.net/forum?id=zKv8qULV6n},
    note={}
}

@misc{comanici2025gemini25pushingfrontier,
      title={Gemini 2.5: Pushing the Frontier with Advanced Reasoning, Multimodality, Long Context, and Next Generation Agentic Capabilities}, 
      author={Comanici, Gheorghe and Bieber, Eric and Schaekermann, Mike and Pasupat, Ice and Sachdeva, Noveen and Dhillon, Inderjit and Blistein, Marcel and Ram, Ori and Zhang, Dan and Rosen, Evan and others},
      year={2025},
      eprint={2507.06261},
      archivePrefix={arXiv},
      primaryClass={cs.CL},
      url={https://arxiv.org/abs/2507.06261}
}

@inproceedings{lin-etal-2022-truthfulqa,
    title = "{T}ruthful{QA}: Measuring How Models Mimic Human Falsehoods",
    author = "Lin, Stephanie  and
      Hilton, Jacob  and
      Evans, Owain",
    editor = "Muresan, Smaranda  and
      Nakov, Preslav  and
      Villavicencio, Aline",
    booktitle = "Proceedings of the 60th Annual Meeting of the Association for Computational Linguistics (Volume 1: Long Papers)",
    month = may,
    year = "2022",
    address = "Dublin, Ireland",
    publisher = "Association for Computational Linguistics",
    url = "https://aclanthology.org/2022.acl-long.229/",
    doi = "10.18653/v1/2022.acl-long.229",
    pages = "3214--3252"
}

@misc{openai2024gpt4ocard,
      title={GPT-4o System Card}, 
      author={OpenAI and : and Aaron Hurst and Adam Lerer and Adam P. Goucher and Adam Perelman and Aditya Ramesh and Aidan Clark andothers},
      year={2024},
      eprint={2410.21276},
      archivePrefix={arXiv},
      primaryClass={cs.CL},
      url={https://arxiv.org/abs/2410.21276}, 
}

@misc{geminiteam2025geminifamilyhighlycapable,
    title={Gemini: A Family of Highly Capable Multimodal Models}, 
    author={Gemini Team and Rohan Anil and Sebastian Borgeaud and Jean-Baptiste Alayrac and Jiahui Yu and Radu Soricut and others},
    year={2025},
    eprint={2312.11805},
    archivePrefix={arXiv},
    primaryClass={cs.CL},
    url={https://arxiv.org/abs/2312.11805}
}

@inproceedings{wang2024cogvlm,
    author = {Wang, Weihan and Lv, Qingsong and Yu, Wenmeng and Hong, Wenyi and Qi, Ji and Wang, Yan and Ji, Junhui and Yang, Zhuoyi and Zhao, Lei and Song, Xixuan and Xu, Jiazheng and Chen, Keqin and Xu, Bin and Li, Juanzi and Dong, Yuxiao and Ding, Ming and Tang, Jie},
    booktitle = {Advances in Neural Information Processing Systems},
    editor = {A. Globerson and L. Mackey and D. Belgrave and A. Fan and U. Paquet and J. Tomczak and C. Zhang},
    pages = {121475--121499},
    publisher = {Curran Associates, Inc.},
    title = {CogVLM: Visual Expert for Pretrained Language Models},
    url = {https://proceedings.neurips.cc/paper_files/paper/2024/file/dc06d4d2792265fb5454a6092bfd5c6a-Paper-Conference.pdf},
    volume = {37},
    year = {2024}
}

@misc{wang2024qwen2vlenhancingvisionlanguagemodels,
      title={Qwen2-VL: Enhancing Vision-Language Model's Perception of the World at Any Resolution}, 
      author={Peng Wang and Shuai Bai and Sinan Tan and Shijie Wang and Zhihao Fan and Jinze Bai and Keqin Chen and Xuejing Liu and Jialin Wang and Wenbin Ge and Yang Fan and Kai Dang and Mengfei Du and Xuancheng Ren and Rui Men and Dayiheng Liu and Chang Zhou and Jingren Zhou and Junyang Lin},
      year={2024},
      eprint={2409.12191},
      archivePrefix={arXiv},
      primaryClass={cs.CV},
      url={https://arxiv.org/abs/2409.12191}, 
}

@misc{zheng2025mllmsdeeplyaffectedmodality,
    title={MLLMs are Deeply Affected by Modality Bias}, 
    author={Xu Zheng and Chenfei Liao and Yuqian Fu and Kaiyu Lei and Yuanhuiyi Lyu and Lutao Jiang and Bin Ren and Jialei Chen and Jiawen Wang and Chengxin Li and Linfeng Zhang and Danda Pani Paudel and Xuanjing Huang and Yu-Gang Jiang and Nicu Sebe and Dacheng Tao and Luc Van Gool and Xuming Hu},
    year={2025},
    eprint={2505.18657},
    archivePrefix={arXiv},
    primaryClass={cs.AI},
    url={https://arxiv.org/abs/2505.18657}, 
}

@article{cohen1960coefficient,
  title={A coefficient of agreement for nominal scales},
  author={Cohen, Jacob},
  journal={Educational and psychological measurement},
  volume={20},
  number={1},
  pages={37--46},
  year={1960},
  publisher={Sage Publications Sage CA: Thousand Oaks, CA}
}

@inproceedings{tang-etal-2024-metrics,
    title = "Not All Metrics Are Guilty: Improving {NLG} Evaluation by Diversifying References",
    author = "Tang, Tianyi  and
      Lu, Hongyuan  and
      Jiang, Yuchen  and
      Huang, Haoyang  and
      Zhang, Dongdong  and
      Zhao, Xin  and
      Kocmi, Tom  and
      Wei, Furu",
    editor = "Duh, Kevin  and
      Gomez, Helena  and
      Bethard, Steven",
    booktitle = "Proceedings of the 2024 Conference of the North American Chapter of the Association for Computational Linguistics: Human Language Technologies (Volume 1: Long Papers)",
    month = jun,
    year = "2024",
    address = "Mexico City, Mexico",
    publisher = "Association for Computational Linguistics",
    url = "https://aclanthology.org/2024.naacl-long.367/",
    doi = "10.18653/v1/2024.naacl-long.367",
    pages = "6596--6610"
}

@InProceedings{pmlr-v235-khan24a,
  title = 	 {Debating with More Persuasive {LLM}s Leads to More Truthful Answers},
  author =       {Khan, Akbir and Hughes, John and Valentine, Dan and Ruis, Laura and Sachan, Kshitij and Radhakrishnan, Ansh and Grefenstette, Edward and Bowman, Samuel R. and Rockt\"{a}schel, Tim and Perez, Ethan},
  booktitle = 	 {Proceedings of the 41st International Conference on Machine Learning},
  pages = 	 {23662--23733},
  year = 	 {2024},
  editor = 	 {Salakhutdinov, Ruslan and Kolter, Zico and Heller, Katherine and Weller, Adrian and Oliver, Nuria and Scarlett, Jonathan and Berkenkamp, Felix},
  volume = 	 {235},
  series = 	 {Proceedings of Machine Learning Research},
  month = 	 {21--27 Jul},
  publisher =    {PMLR},
  url = 	 {https://proceedings.mlr.press/v235/khan24a.html}
}

\newpage

\appendix

\section{Prompt Templates} \label{sec:appendix-prompt}

\subsection{Prompt for Comparing Informativeness}

Table \ref{table:info-prompt} presents the prompt we used with GPT-4o to select the more informative candidate answer $y^i$.

\subsection{Prompt used in \texttt{Standard Ref}}

Table \ref{table:ref-prompt} presents the prompt we used to generate the reference answer in \texttt{Standard Ref} baseline.

\subsection{Prompts used in \texttt{\OurMethod}}

\paragraph{Prompt for generating anchor}

Table \ref{table:birch-ref-prompt} shows the prompt we used to generate the anchor in \OurMethod.

\paragraph{Prompt for judging}

Table \ref{table:judge-w-ref-prompt} shows the prompt we use in \OurMethod\ to judge with the anchor generated using the prompt in Table \ref{table:birch-ref-prompt}.

\section{Benchmark Details} \label{sec:appendix-bench}

\subsection{Introduction}

\paragraph{MLLM-as-a-Judge \cite{chen2024mllm}} is the first benchmark to evaluate multimodal LLMs as judges. It contains 4,414 image-instruction pairs from 14 datasets covering tasks such as image captioning, math reasoning, text reading, and infographic understanding. Candidate answers are generated using six MLLMs: GPT-4V \cite{openai2024gpt4ocard}, Gemini-Pro-Vision \cite{geminiteam2025geminifamilyhighlycapable}, LLaVA-1.5-13b, LLaVA-1.6-34b \cite{liu2024improved}, CogVLM \cite{wang2024cogvlm}, and Qwen-VL-Max \cite{wang2024qwen2vlenhancingvisionlanguagemodels}. The benchmark provides three settings: (1) scoring evaluation, where each answer is rated from 1 to 5, (2) pairwise comparison, where two answers are compared to select the better one, and (3) batch ranking, where answers are ordered by quality. In our work, we use only the pairwise comparison setting, which includes 7,756 samples.

\paragraph{VL-RewardBench \cite{li2025vl}} focuses on the pairwise comparison setting, with 1,247 samples from 7 datasets. It includes hallucination-oriented queries in addition to general instruction-following and reasoning tasks, and provides error-type labels. To ensure the samples are challenging enough to expose model weaknesses, it only keeps cases where multiple smaller models (such as LLaVA-1.5/1.6-7B \cite{liu2024improved} and Qwen2-VL-7B \cite{wang2024qwen2vlenhancingvisionlanguagemodels}) make errors.

\subsection{Preprocessing}

\begin{figure*}[ht]
\centering

\begin{tcolorbox}[colback=gray!10, colbacktitle=RoyalBlue, center title, title={Benchmark: VL-RewardBench}]

\begin{minipage}[t]{0.4\linewidth}
    \centering
    \includegraphics[width=\linewidth]{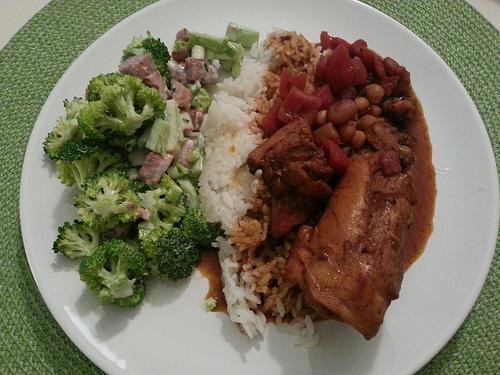}
\end{minipage}
\hfill
\begin{minipage}[t]{0.55\linewidth}
    \vspace{-11.5em}

    \textbf{Query:} Is the white food to the right of the meat on the right side?
\end{minipage}

\hr

\textbf{\textcolor{red}{Answer A:}} \textcolor{PineGreen}{No}, the white food is to the \textcolor{PineGreen}{left of the meat} on the right side. It is rice.

\vspace{1em}

\textbf{\textcolor{PineGreen}{Answer B:}} \textcolor{red}{Yes}, the white food is rice, which is located to the \textcolor{red}{right of the meat} on the right side of the plate.

\vspace{1em}

\textbf{Human preference:} \textcolor{PineGreen}{Answer B}

\end{tcolorbox}

\caption{Example showing a human labeling error in VL-RewardBench. The white food (rice) is clearly to the left of the meat on the right side, but humans incorrectly chose B.}
\label{fig:benchmark-issues-food}

\end{figure*}

\begin{figure*}[ht]
\centering

\begin{tcolorbox}[colback=gray!10, colbacktitle=RoyalBlue, center title, title={Benchmark: VL-RewardBench}]

\begin{minipage}[t]{0.4\linewidth}
    \centering
    \includegraphics[width=\linewidth]{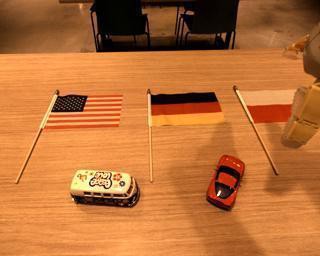}
\end{minipage}
\hfill
\begin{minipage}[t]{0.55\linewidth}
    \vspace{-12em}
    
    \textbf{Query:} In which direction should I push the red car to bring it closer to the american flag? Up, down, right or left?
\end{minipage}

\hr

\textbf{\textcolor{red}{Answer A:}} To bring the red car closer to the American flag, you should push it to the left.

\vspace{1em}

\textbf{\textcolor{PineGreen}{Answer B:}} To determine the direction in which to push the red car to bring it closer to the American flag, we can analyze the position of the red car relative to the American flag. Observing the flags' arrangement, \textcolor{red}{the American flag is positioned to the right of the red car}. Therefore, to move the car closer to the American flag, we would need to push the car towards the left. Based on this spatial analysis, the correct direction to push the red car is "left".

\vspace{1em}

\textbf{Human preference:} \textcolor{PineGreen}{Answer B}

\end{tcolorbox}

\caption{Example showing a human labeling error in VL-RewardBench. Answer B incorrectly states that the American flag is to the left of the red car, yet it was still chosen by humans.}
\label{fig:benchmark-issues-car-flag}

\end{figure*}

\begin{figure*}[ht]
\centering

\begin{tcolorbox}[colback=gray!10, colbacktitle=RoyalBlue, center title, title={Benchmark: VL-RewardBench}]

\begin{minipage}[t]{0.35\linewidth}
    \centering
    \includegraphics[width=\linewidth]{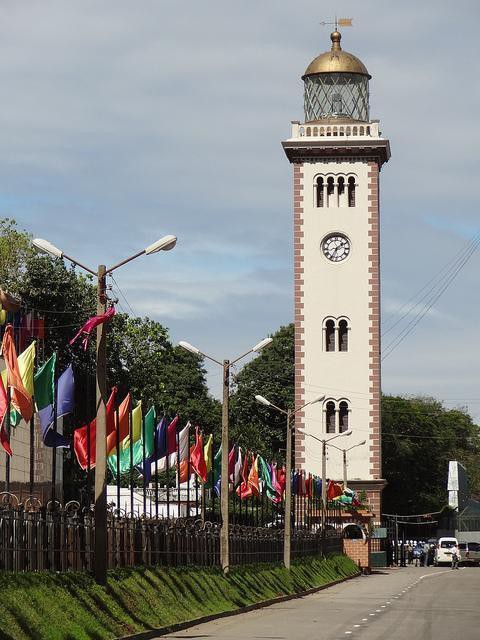}
\end{minipage}
\hfill
\begin{minipage}[t]{0.6\linewidth}
    \vspace{-17.5em}

    \textbf{Query:} How many flags can be seen?
\end{minipage}

\hr

\textbf{\textcolor{red}{Answer A:}} There are many flags present in the scene, too many to count. They appear to be hanging on a fence and arranged in a long line along the roadside.

\vspace{1em}

\textbf{\textcolor{PineGreen}{Answer B:}} There are \textcolor{red}{27 flags} visible in this image. The flags appear to be hanging from vertical poles and arranged in a line along the street leading up to the clock tower.

\vspace{1em}

\textbf{Human preference:} \textcolor{PineGreen}{Answer B}

\end{tcolorbox}

\caption{Example showing a human labeling error in VL-RewardBench. Although precisely counting the flags here is very difficult, the number is clearly much higher than the 27 claimed in Answer B. Answer A, while not giving an exact count, is better since it avoids misleading information. However, the human still chose Answer A.}
\label{fig:benchmark-issues-flag}

\end{figure*}

\begin{figure*}[ht]
\centering

\begin{tcolorbox}[colback=gray!10, colbacktitle=RoyalBlue, center title, title={Benchmark: VL-RewardBench}]

\begin{minipage}[t]{0.4\linewidth}
    \centering
    \includegraphics[width=\linewidth]{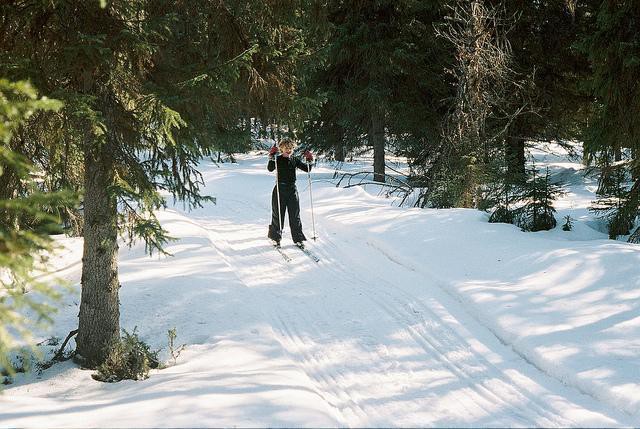}
\end{minipage}
\hfill
\begin{minipage}[t]{0.55\linewidth}
    \vspace{-10em}

    \textbf{Query:} How many tracks can be seen in the snow?
\end{minipage}

\hr

\textbf{\textcolor{red}{Answer A:}} In the snowy area, \textcolor{PineGreen}{many tracks can be seen}. This implies that many skiers may have skied in this ski resort, but the picture only captures the boy skiing alone in the snow.

\vspace{1em}

\textbf{\textcolor{PineGreen}{Answer B:}} In the snowy ground, \textcolor{red}{only a single trace can be seen}. This suggests that the skier might be skiing alone, or they are the first one to ski on this particular path.

\vspace{1em}

\textbf{Human preference:} \textcolor{PineGreen}{Answer B}

\end{tcolorbox}

\caption{Example showing a human labeling error in VL-RewardBench. The image clearly shows multiple tracks on the snow ground, matching Answer A, but humans incorrectly chose B.}
\label{fig:benchmark-issues-snow-track}

\end{figure*}

We found that human annotators made many errors when labeling the better candidate answer in the benchmarks we used. Figures \ref{fig:benchmark-issues-food}, \ref{fig:benchmark-issues-car-flag}, \ref{fig:benchmark-issues-flag} and \ref{fig:benchmark-issues-snow-track} show examples of such errors. These mistakes undermine the reliability of evaluating VLM-as-a-Judge. To address this, we manually reviewed the data samples and corrected the identified errors before evaluation.

\section{More Experiment Details} \label{sec:appendix-exp}

\subsection{Reliability of Informativeness Measuring} \label{sec:appendix-exp-reliable-info}

As described in Section \ref{sec:judge-info}, we use GPT-4o to measure informativeness, and our analysis depends on its reliability. To validate this, we randomly sample 100 examples from each benchmark, MLLM-as-a-Judge \cite{chen2024mllm} and VL-RewardBench \cite{li2025vl}, and ask human annotators to decide which candidate is more informative based on text alone, using the same rubric as GPT-4o (Figure \ref{table:info-prompt}). We then measure agreement between human and GPT-4o using match rate, which is the percentage of matching cases in [0, 1], and Cohen's kappa \cite{cohen1960coefficient} in [-1, 1], where higher values indicate better agreement.

Table \ref{table:info-kappa} shows that the match rate exceeds 0.9 and Cohen's kappa exceeds 0.8, indicating strong agreement between GPT-4o and human judgments and confirming its reliability for comparing informativeness.

\begin{table}[ht]
    \centering
        
    \resizebox{\linewidth}{!}{
    
    \begin{tabular}{lcc}
        \toprule
        
        & MLLM-as-a-Judge & VL-RewardBench \\
        
        \midrule

        \textbf{Match rate}  & 0.91	& 0.94 \\

        \textbf{Cohen's kappa} & 0.82 & 0.88 \\
        
        \bottomrule
    \end{tabular}
    }

    \caption{Agreement between human annotators and GPT-4o on informativeness measurement.}

    \label{table:info-kappa}
\end{table}

\subsection{Image Reliance Across Domains}

\begin{figure*}[ht]
    \centering

    \centering
    \begin{minipage}{0.32\linewidth}
        \centering
        \includegraphics[width=\linewidth]{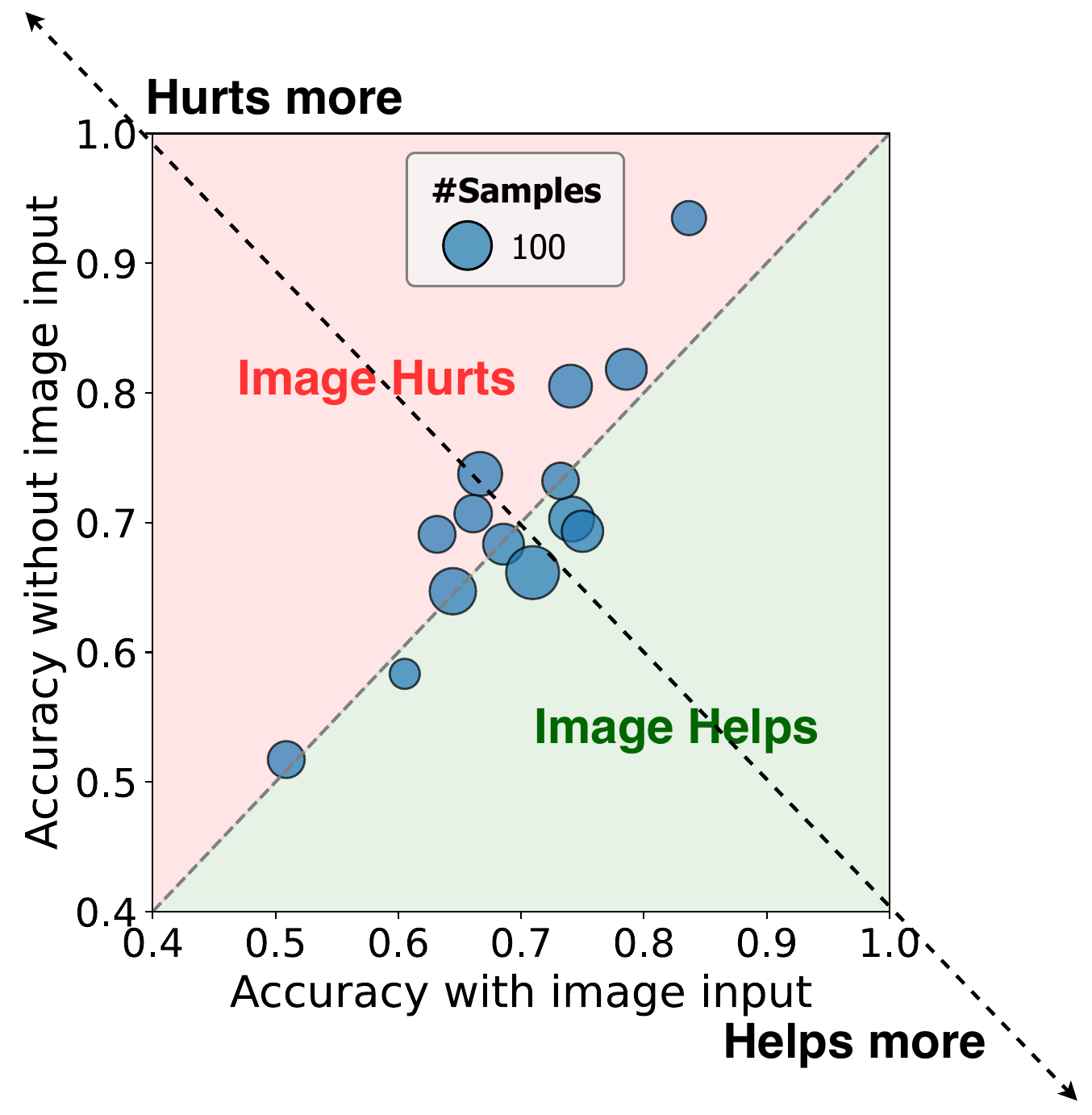}
    \end{minipage}
    \hfill
    \begin{minipage}{0.32\linewidth}
        \centering
        \includegraphics[width=\linewidth]{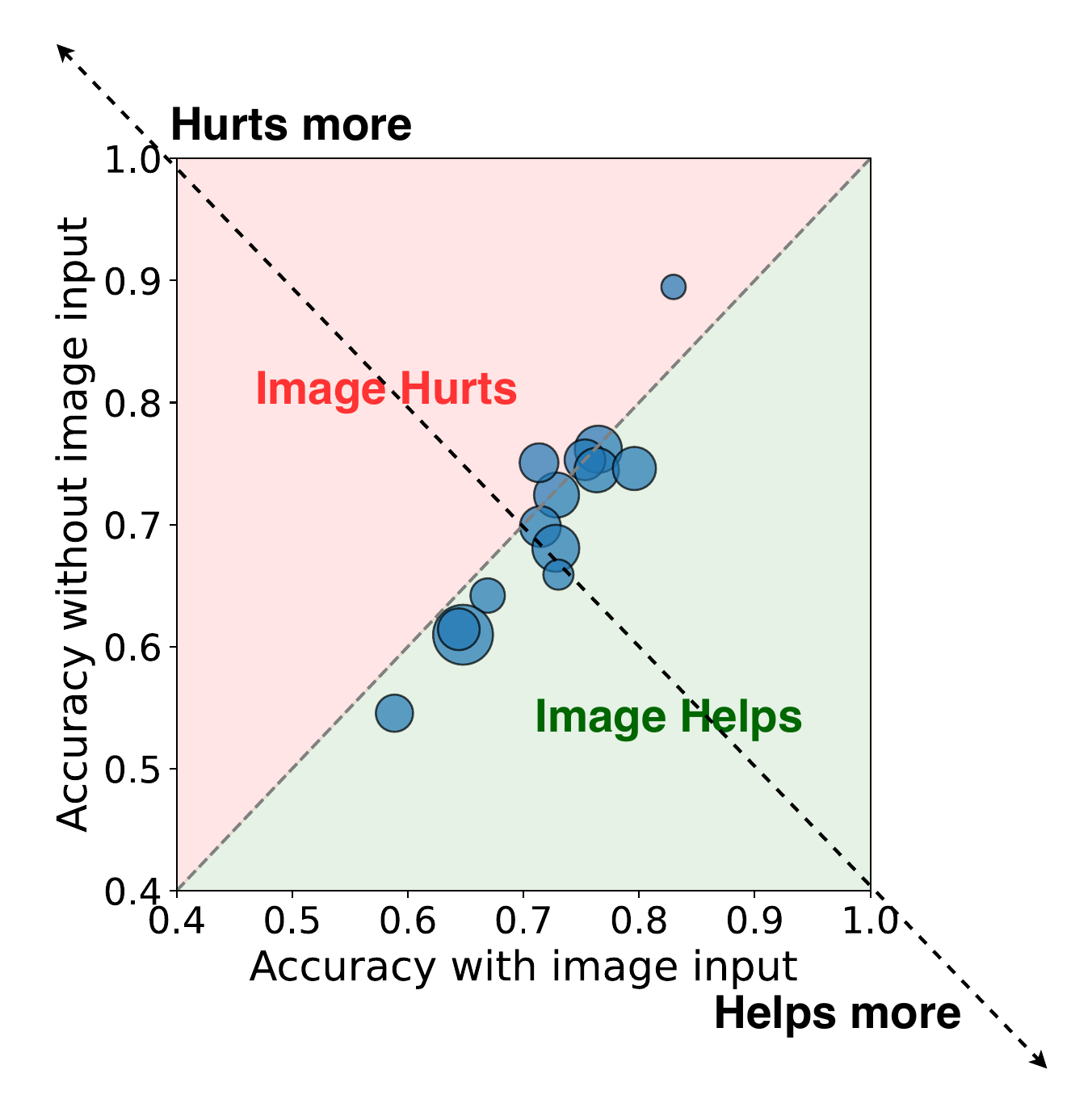}
    \end{minipage}
    \hfill
    \begin{minipage}{0.32\linewidth}
        \centering
        \includegraphics[width=\linewidth]{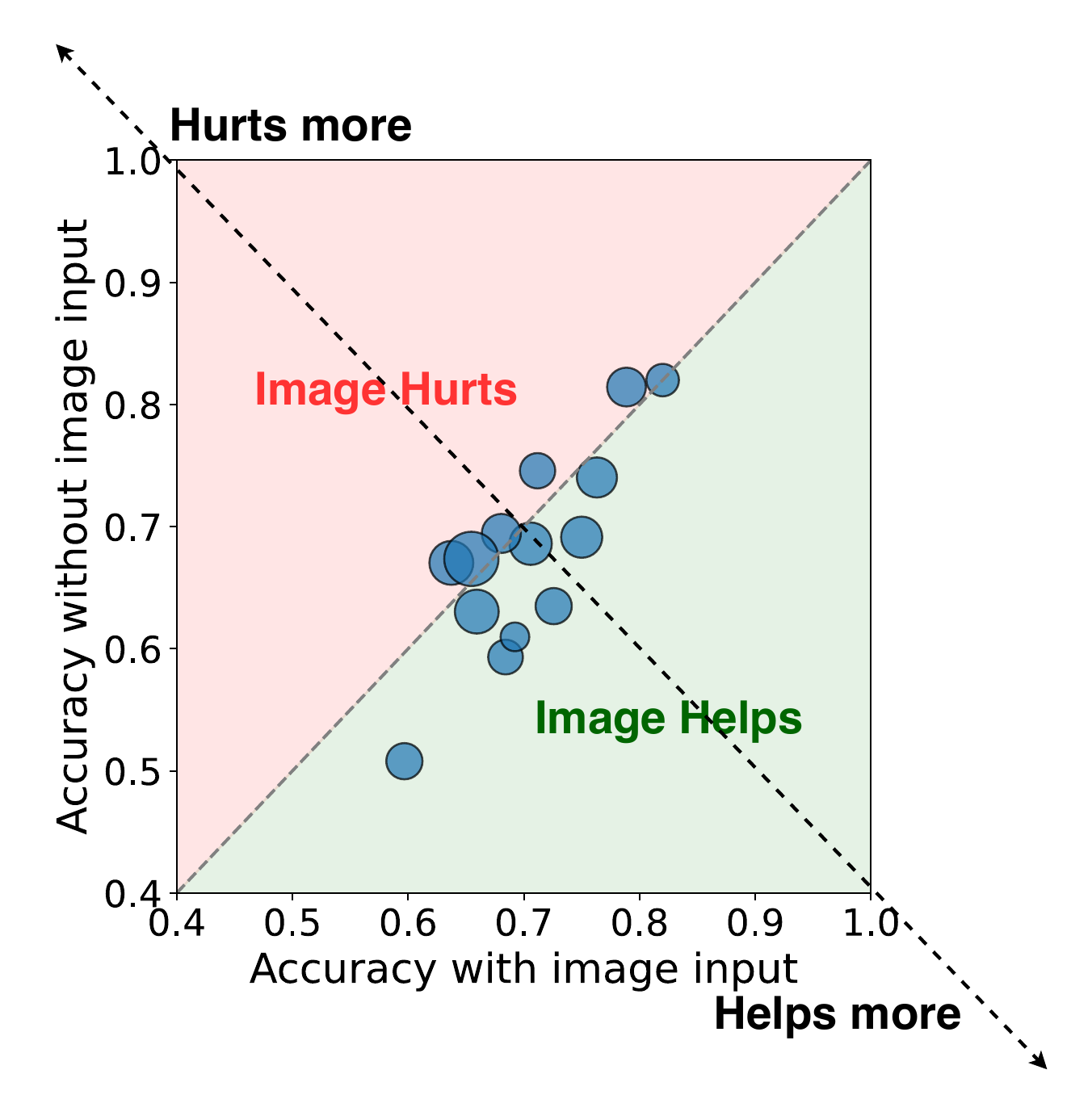}
    \end{minipage}
    
    \caption{Per-domain comparison of GPT-4o (left), Llama-3.2-Vision-90B (middle), and Gemini-2.5-Flash (right), with and without image input, on the MLLM-as-a-Judge benchmark. Each scatter point represents a domain, with its size indicating the number of samples. The x-axis shows accuracy with image input; the y-axis shows accuracy without. Points in the green area indicate that the image improves accuracy, while points in the red area indicate that the image reduces accuracy. The farther a point is from the diagonal, the greater the impact (positive or negative) of the image. For all models, most domains cluster near the diagonal, showing that images provide little additional benefit.}
    \label{fig:with-no-img-compare}
\end{figure*}

Figure \ref{fig:with-no-img-compare} shows a per-domain comparison of GPT-4o \cite{openai2024gpt4ocard}, Llama-3.2-Vision-90B \cite{dubey2024llama}, and Gemini-2.5-Flash \cite{comanici2025gemini25pushingfrontier}, with and without image input on the MLLM-as-a-Judge \cite{chen2024mllm} benchmark. Most domains lie close to the diagonal, showing that images add little benefit and sometimes even reduce performance.

\subsection{Downsampling Details for Section \ref{sec:bias-ib}} \label{sec:appendix-exp-downsampling}

We compute informativeness bias (IB) and length bias (LB) by subtracting accuracy on different subsets (Section \ref{sec:ib} and \ref{sec:lb}). Since this can be affected by randomness, variance, and differences in sample selection or subset size, we mitigate it by downsampling each subset at different ratios multiple times. We report the mean across all samples, and show variance bars in Figures \ref{fig:lb-ib-overall} and \ref{fig:ib-lb-phi4}.

Specifically, for IB the two subsets are $\mathcal{D}_\text{IDS}$ and $\mathcal{D}_\text{CDS}$, and for LB they are $\mathcal{D}_{y^c \text{ is longer}}$ and $\mathcal{D}_{y^c \text{ is shorter}}$. We downsample each at 40\%, 60\%, 80\%, and 100\% (full set), with 10 samples per ratio. This results in $4 \times 4 \times 10 = 160$ downsamplings for each IB or LB computation.

The variance from this process is reasonably low (see Figures \ref{fig:lb-ib-overall} and \ref{fig:ib-lb-phi4}), confirming the validity of our results.

\subsection{Informativeness Bias Across Domains}

\begin{figure}[ht]
  \includegraphics[width=\columnwidth]{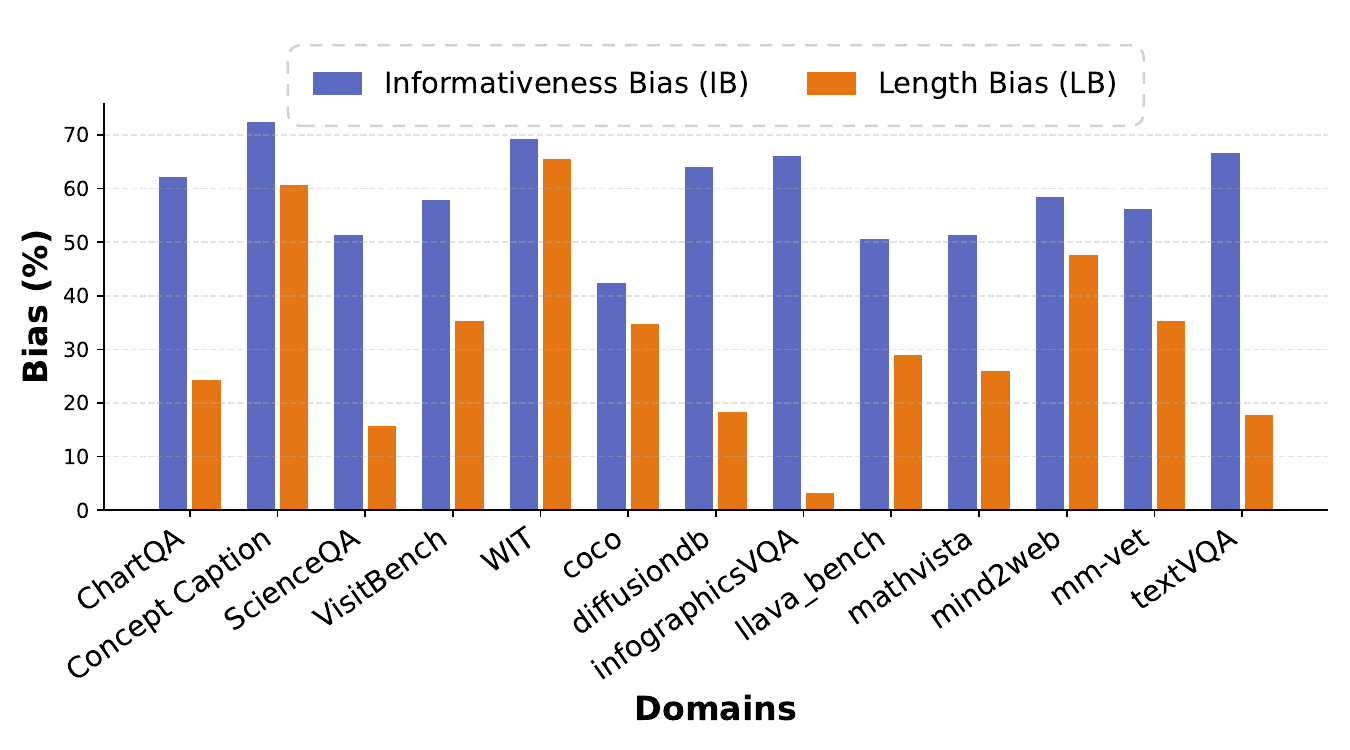}
  \caption{Informativeness bias (IB) and length bias (LB) comparison for different domains on MLLM-as-a-judge benchmark. IB is stronger than LB across all domains, confirming that informativeness is a greater source of bias than length.}
  \label{fig:mllmjudge-lb-ib-per-dataset}
\end{figure}

\begin{figure}[ht]
  \includegraphics[width=\columnwidth]{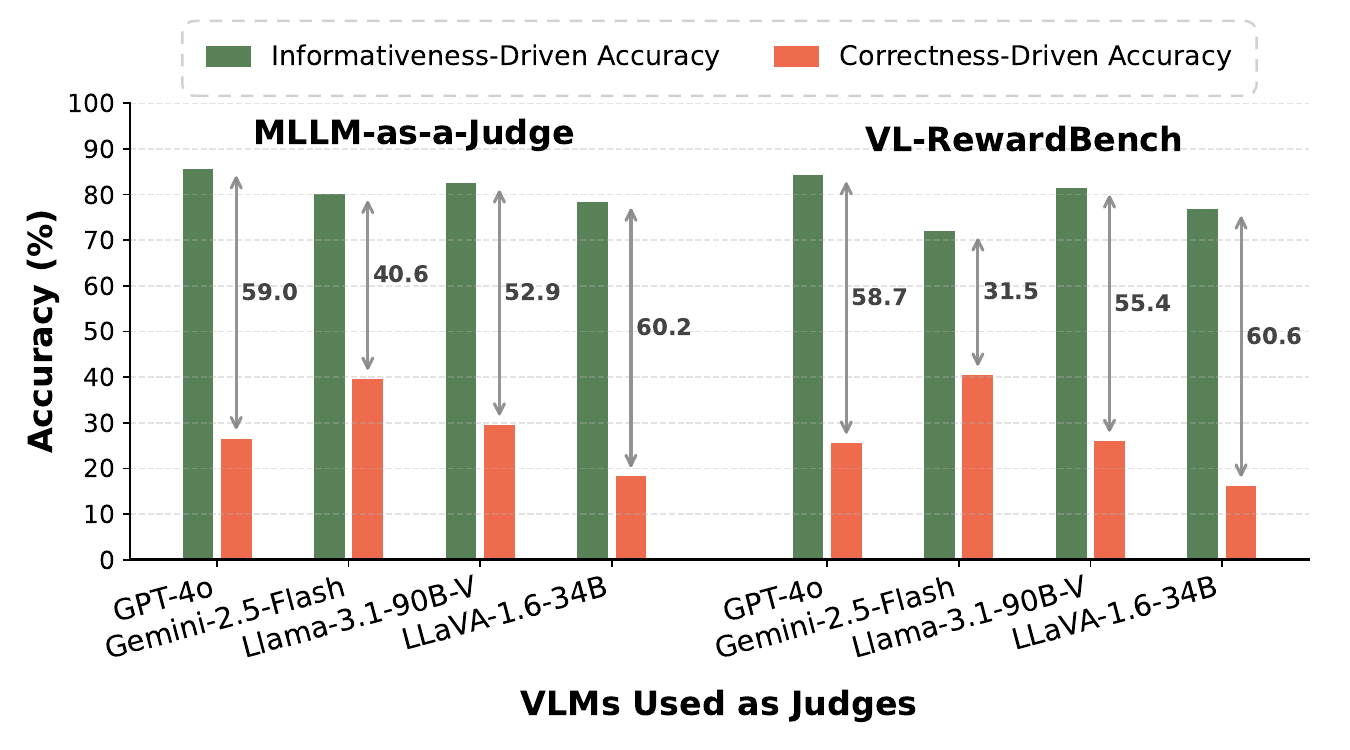}
  \caption{Accuracy on informativeness-driven ($\text{Acc}_\text{CDS}$) and correctness-driven ($\text{Acc}_\text{CDS}$) subsets of GPT-4o. $\text{Acc}_\text{CDS}$ is much lower than $\text{Acc}_\text{IDS}$, revealing weak performance in cases where correctness should drive the decision.}
  \label{fig:ida-cda}
\end{figure}

Figure \ref{fig:mllmjudge-lb-ib-per-dataset} compares informativeness bias (IB) and length bias (LB) across domains in MLLM-as-a-Judge, showing that IB is higher than LB in almost all cases. Figure \ref{fig:ida-cda} presents $\text{Acc}_\text{IDS}$ and $\text{Acc}_\text{CDS}$, the accuracies for the informativeness-driven and correctness-driven subsets. A large gap appears, with $\text{Acc}_\text{IDS}$ much higher. Together, these results indicate that IB is far more significant than LB.


\subsection{Ablation of Length Bias} \label{sec:appendix-exp-ablate-lb}

Inspired by \cite{ye2025justice, hu2025explaining}, we separate length bias from informativeness bias by equalizing the lengths (measured in words) of the two answers. Specifically, before computing informativeness bias, we extend the shorter answer by asking GPT-4o to lengthen its existing content until it matches the longer one. We strictly instruct GPT-4o not to add any new information beyond the original answer, and to increase length only by repeating existing content or adding verbosity.

After this, we recheck the dataset and keep only samples where the length difference between the two answers is within a small threshold (5 words). This process removes the effect of length bias, since both answers are made nearly equal in length.

\subsection{Lower Informativeness Bias for Phi-4-Multimodal} \label{sec:appendix-exp-phi4}

\begin{figure}[ht]
    \centering
    \includegraphics[width=\linewidth]{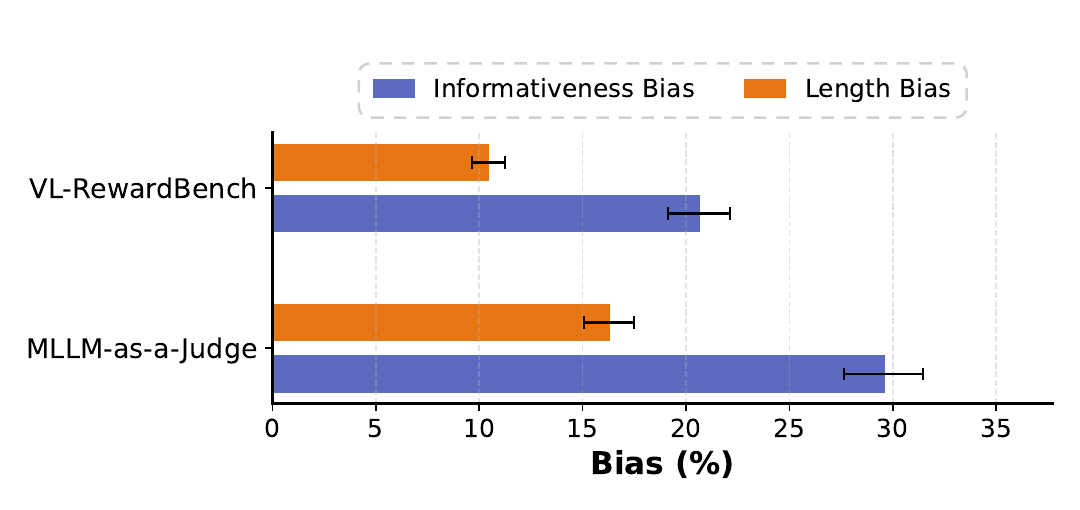}
    
    \caption{Informativeness bias (IB) and length bias (LB) for Phi-4-Multimodal \cite{abdin2024phi4technicalreport}. The mean and variance are computed by downsampling $\mathcal{D}_\text{IDS}$ and $\mathcal{D}_\text{CDS}$ at different ratios, each sampled multiple times. Both IB and LB of Phi-4-Multimodal are significantly lower than in other models (see Figure \ref{fig:lb-ib-overall}). However, a clear gap between IB and LB remains, supporting our claim that IB drives LB.}
    \label{fig:ib-lb-phi4}
\end{figure}

Figure \ref{fig:ib-lb-phi4} compares informativeness bias (IB) and length bias (LB) for Phi-4-Multimodal \cite{abdin2024phi4technicalreport}. To reduce randomness and variance, we report mean and variance computed by downsampling $\mathcal{D}_\text{IDS}$ and $\mathcal{D}_\text{CDS}$ at different ratios multiple times, as described in Appendix \ref{sec:appendix-exp-downsampling}. Phi-4-Multimodal shows IB below 30\% on both benchmarks, significantly lower than other models (see Figure \ref{fig:lb-ib-overall}, where IB is around 60\%). 

This explains why \OurMethod\ yields smaller improvements on Phi-4-Multimodal, as it is designed to reduce IB, and Phi-4-Multimodal already has low IB. As shown in Table \ref{table:acc-ida-cda-ib-lb-full}, \OurMethod\ improves $\text{Acc}_\text{CDS}$ by only 5.3\% and reduces IB by 2.0\% for Phi-4-Multimodal, much lower than the gains observed on other models.

\subsection{Why \OurMethod\ Works Well}

Table \ref{table:acc-ida-cda-ib-lb-full} extends Table \ref{table:acc-ida-cda-ib-before-after} and reports changes across multiple metrics for other models before and after applying \texttt{\OurMethod} on the MLLM-as-a-Judge \cite{chen2024mllm} benchmark. 

\OurMethod\ consistently reduces informativeness bias (IB) by substantially improving $\text{Acc}_\text{CDS}$ while keeping $\text{Acc}_\text{IDS}$ strong across all models, leading to higher overall performance. Although designed to mitigate IB, \OurMethod\ also greatly reduces length bias (LB), further supporting our finding that IB and LB are intertwined and that addressing IB also reduces LB.

\section{More Case Studies} \label{sec:appendix-case}

\subsection{Judges Misled by Informativeness, Mitigated by a Reference Answer}

\begin{figure*}[ht]
\centering

\tcbset{
    colback=gray!10, 
    colframe=white, 
    boxrule=0pt, 
    fonttitle=\bfseries,
    coltitle=white,
    colbacktitle=gray,
    width=\linewidth,
    % sharp corners,
    arc=2mm,
    center title,
    before upper={\fontsize{10}{12}\selectfont}
}

\begin{tcolorbox}[colback=RoyalBlue!10, colbacktitle=RoyalBlue, title={Benchmark: MLLM-as-a-Judge}]

\begin{minipage}[t]{0.45\linewidth}
    \centering
    \includegraphics[width=\linewidth]{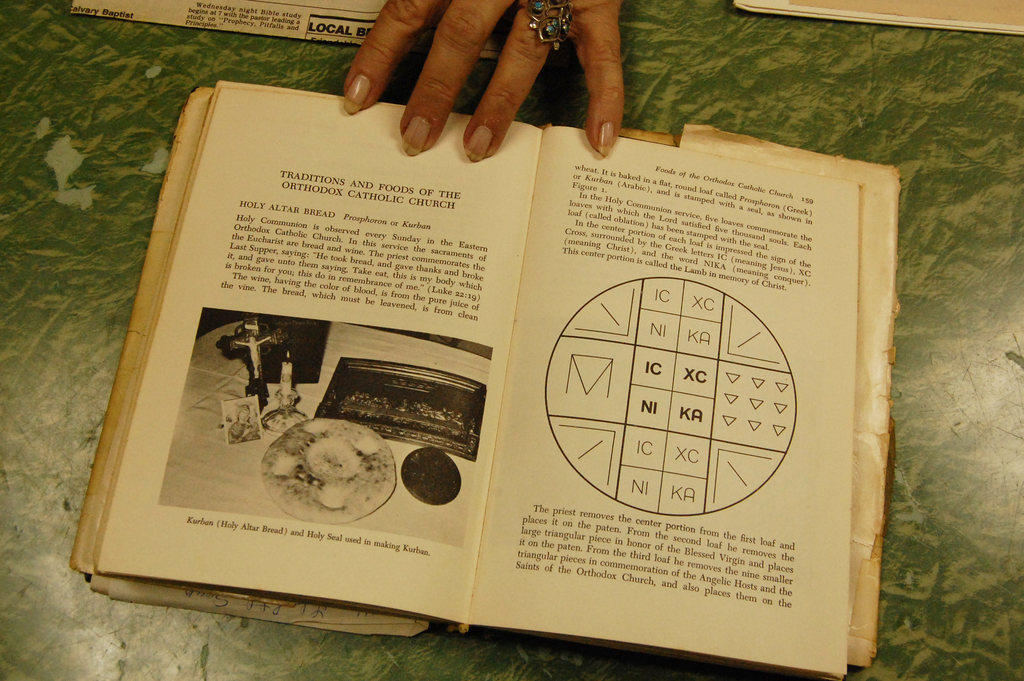}
\end{minipage}
\hfill
\begin{minipage}[t]{0.5\linewidth}
    \vspace{-12.5em}

    \textbf{Query:} This is a task of text reading on natural image. Please analyze this figure in detail and answer the following question with reason based on this figure: what is written on the top of the left page?

    \vspace{6.5em}
    \textcolor{black!60}{\footnotesize Zoom in to see the image details more clearly.}

\end{minipage}

\end{tcolorbox}

\begin{tcolorbox}[colback=Orange!10, colbacktitle=Orange, title={Candidate Answers}]

\textcolor{PineGreen}{\textbf{Answer A} (chosen):} TRADITIONS AND FOODS OF THE ORTHODOX CATHOLIC CHURCH

\vspace{1em}

\textcolor{red}{\textbf{Answer B} (rejected):} This is a picture of two open books. The left page has a black and white photo of an altar with a loaf of bread on it. The caption below the photo reads "Kurban (Holy Altar) and Holy Seal used in making Kurban." \textcolor{red}{The text on the right page is titled "Traditions and Foods of the Orthodox Catholic Church."} The first paragraph on this page begins with "Holy Communion is observed every Sunday in the Orthodox Catholic Church. In the Eastern Orthodox Church, wine, having the color of blood, and pure unleavened bread are used." Based on the information provided, we can infer that \textcolor{red}{the text on the top of the left page is "Kurban (Holy Altar) and Holy Seal used in making Kurban."}

\end{tcolorbox}

\begin{tcolorbox}[title={VLM-as-a-Judge (GPT-4o)}]

\textbf{\texttt{Without Reference}:}

\begin{itemize}
    \item \textbf{Preference:} \textcolor{red}{Answer B}
    \item \textbf{Reason:} Assistant A's response directly provides the text 'TRADITIONS AND FOODS OF THE ORTHODOX CATHOLIC CHURCH' as the answer to the user's question. \textcolor{red}{However, this is incorrect because the provided image shows that this text is on the top of the right page, not the left.} Assistant A does not analyze the image in detail or provide reasoning to support the answer. Assistant B's response provides a detailed description of the image, including the content of both pages. \textcolor{red}{The response of Assistant B demonstrates a more comprehensive analysis of the image and reasoning process compared to Assistant A.}
\end{itemize}

\textbf{\texttt{With Reference}:}

\begin{itemize} 
    \item \textbf{Preference:} \textcolor{PineGreen}{Answer A}
    \item \textbf{Reference answer generated by the judge itself:} \textcolor{PineGreen}{The text written at the top of the left page is: "TRADITIONS AND FOODS OF THE ORTHODOX CATHOLIC CHURCH.} This title introduces the content on the page, which discusses the preparation and significance of "Holy Altar Bread" in the Orthodox Catholic Church.
    \item \textbf{Reason:} \textcolor{PineGreen}{Assistant A correctly identifies the text at the top of the left page as "TRADITIONS AND FOODS OF THE ORTHODOX CATHOLIC CHURCH,"} which matches the reference answer. Assistant B, on the other hand, provides a detailed description of the image \textcolor{PineGreen}{but incorrectly infers that the text at the top of the left page is "Kurban (Holy Altar) and Holy Seal used in making Kurban," which is actually the caption below the image}. This misidentification makes Assistant B's response less accurate despite its detailed explanation.
\end{itemize}
    
\end{tcolorbox}

\caption{In this example, VLM-as-a-Judge prefers Answer B, which gives more details about the book but makes two mistakes: it treats the figure caption as text on the top of the left page, and it misplaces the actual text from the top of the left page onto the right page. The model does recognize the correct text on the top of the left page, yet still judges Answer B as correct. Using the model's own generated answer as a reference helps mitigate this issue.}
\label{fig:case-basic-ref-vs-no-ref-catholic}

\end{figure*}

\begin{figure*}[ht]
\centering

\tcbset{
    colback=gray!10, 
    colframe=white, 
    boxrule=0pt, 
    fonttitle=\bfseries,
    coltitle=white,
    colbacktitle=gray,
    width=\linewidth,
    % sharp corners,
    arc=2mm,
    center title,
    % before upper={\fontsize{10}{12}\selectfont}
}

\begin{tcolorbox}[colback=RoyalBlue!10, colbacktitle=RoyalBlue, title={Benchmark: VL-RewardBench}]

\begin{minipage}[t]{0.35\linewidth}
    \centering
    \includegraphics[width=\linewidth]{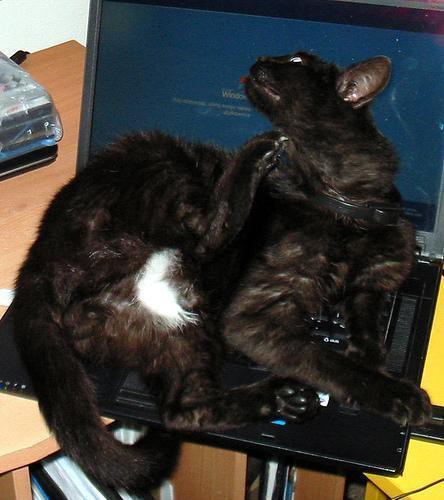}
\end{minipage}
\hfill
\begin{minipage}[t]{0.60\linewidth}
    \vspace{-15em}

    \textbf{Query:} Convey the specifics of the image with meticulous attention to detail.
\end{minipage}

\end{tcolorbox}

\begin{tcolorbox}[colback=Orange!10, colbacktitle=Orange, title={Candidate Answers}]

\textcolor{PineGreen}{\textbf{Answer A} (chosen):} A black cat with white fur around its belly is sprawled out on a laptop keyboard, which has an open screen.

\vspace{1em}

\textcolor{red}{\textbf{Answer B} (rejected):} A black cat lounges on a laptop, its fur in stark \textcolor{red}{contrast to the silver computer}. A white patch adorns the animal's chest, while its ears perk up attentively.

\end{tcolorbox}

\begin{tcolorbox}[title={VLM-as-a-Judge (Phi-4-Multimodal)}]

\textbf{\texttt{Without Reference Answer}:}

\begin{itemize}
    \item \textbf{Preference:} \textcolor{red}{Answer B}
    \item \textbf{Reason:} Assistant A misses some details, such as the cat's posture, \textcolor{red}{the silver color of the laptop}, and the cat's attentive ears. Assistant B offers a more detailed response, \textcolor{red}{highlighting the black cat's contrast with the silver laptop}.

\end{itemize}

\textbf{\texttt{With Reference Answer}:}

\begin{itemize} 
    \item \textbf{Preference:} \textcolor{PineGreen}{Answer A}
    \item \textbf{Reference answer generated by the judge itself:} A black cat with a white patch of fur on its lower belly is sprawled on a \textcolor{PineGreen}{black laptop}. The laptop screen is open, displaying a blue background.
    \item \textbf{Reason:} Assistant B, while descriptive, \textcolor{PineGreen}{incorrectly states that the laptop is silver}, deviating from the image and reference answer.
\end{itemize}
    
\end{tcolorbox}

\caption{In this example, VLM-as-a-Judge prefers Answer B, which provides more details about the cat but incorrectly claims the laptop is silver when it is actually black. The model does recognize the correct color of the laptop, yet still judges Answer B as correct. Using its own generated answer as a reference helps mitigate this issue.}
\label{fig:case-basic-ref-vs-no-ref-cat}

\end{figure*}

Figures \ref{fig:case-basic-ref-vs-no-ref-catholic}, \ref{fig:case-basic-ref-vs-no-ref-cigarette} (MLLM-as-a-Judge), and \ref{fig:case-basic-ref-vs-no-ref-cat} (VL-RewardBench) provide additional qualitative examples showing how VLM-as-a-Judge is misled by a more informative but wrong answer, even when it clearly conflicts with the judge's own response. However, when the judge uses its own answer as a reference, it can easily detect inconsistencies with the image and select the correct choice.

\subsection{Why \texttt{\OurMethod} outperforms \texttt{Standard Ref}?}

\begin{figure*}[ht]
\centering

\tcbset{
    colback=gray!10, 
    colframe=white, 
    boxrule=0pt, 
    fonttitle=\bfseries,
    coltitle=white,
    colbacktitle=gray,
    width=\linewidth,
    arc=2mm,
    center title,
    before upper={\fontsize{10}{12}\selectfont}
}

\begin{tcolorbox}[colback=RoyalBlue!10, colbacktitle=RoyalBlue, title={Benchmark: MLLM-as-a-Judge}]

\begin{minipage}[t]{0.42\linewidth}
    \centering
    \includegraphics[width=\linewidth]{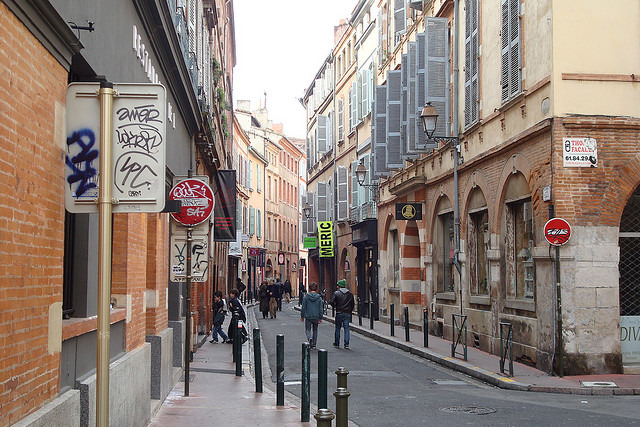}
\end{minipage}
\hfill
\begin{minipage}[t]{0.54\linewidth}
    \vspace{-11.5em}

    \textbf{Query:} Please analyse this figure in detail and answer the following
  question based on this figure: What do the red street signs mean?
\end{minipage}
    
\end{tcolorbox}

\begin{tcolorbox}[colback=Orange!10, colbacktitle=Orange, title={Candidate Answers}]

\textcolor{red}{\textbf{Answer A} (rejected):} The red street signs in the image are 'No Entry' signs, which indicate that vehicles are not allowed to enter the street.

\vspace{1em}

\textcolor{PineGreen}{\textbf{Answer B} (chosen):} The red street signs in the figure indicate that vehicles are not allowed to enter the street. \textcolor{PineGreen}{The signs are placed at the beginning of the street and are visible to drivers approaching the street.} The red street signs are a traffic control device used to regulate traffic flow and keep pedestrians safe.

\end{tcolorbox}

\begin{tcolorbox}[title={VLM-as-a-Judge (GPT-4o)}]

\textbf{\texttt{Standard Ref}:}

\begin{itemize}
    \item \textbf{Preference:} \textcolor{red}{Answer A}
    \item \textbf{Reference answer:} The red street signs in the image are circular with a white horizontal bar in the center. These are "No Entry" signs, indicating that vehicles are not allowed to enter the street in the direction the sign is facing.
    \item \textbf{Reason:} Assistant A correctly identifies the red street signs as "No Entry" signs. Assistant B also identifies the signs as indicating "No Entry" and elaborates on their function as traffic control devices, emphasizing their role in regulating traffic and ensuring safety. However, \textcolor{red}{Assistant B's response includes speculative details (e.g., placement at the beginning of the street) that are not supported by the image}. Assistant A is more aligned with the reference answer.
\end{itemize}

\textbf{\texttt{\OurMethod}:}

\begin{itemize} 
    \item \textbf{Preference:} \textcolor{PineGreen}{Answer B}
    \item \textbf{Anchor:} The red street signs in the image are 'No Entry' signs, clearly indicating that vehicles are prohibited from entering the street. \textcolor{PineGreen}{Positioned at the entrance} and visible to approaching drivers, these signs serve as a traffic control measure to regulate access and ensure pedestrian safety.
    \item \textbf{Reason:} Assistant A provides a concise and accurate explanation of the red street signs, aligning well with the reference answer but lacks depth and detail. Assistant B offers a more comprehensive response, \textcolor{PineGreen}{including additional context about the placement} and purpose of the signs, which enhances the explanation and aligns more closely with the reference answer's depth and detail.
\end{itemize}
    
\end{tcolorbox}

\caption{In this example, \texttt{Standard Ref} rejects the better Answer B, judging its placement of the sign as speculative since this detail is not in the reference. This shows how it over-penalizes informativeness. In contrast, \texttt{\OurMethod} ensures such placement details are included in the generated anchor, allowing it to correctly verify Answer B and achieve a better balance between informativeness and correctness.}
\label{fig:case-birch-vs-basic-ref-stop-sign}

\end{figure*}

\begin{figure*}[ht]
\centering

\tcbset{
    colback=gray!10, 
    colframe=white, 
    boxrule=0pt, 
    fonttitle=\bfseries,
    coltitle=white,
    colbacktitle=gray,
    width=\linewidth,
    arc=2mm,
    center title,
    before upper={\fontsize{10}{12}\selectfont}
}

\begin{tcolorbox}[colback=RoyalBlue!10, colbacktitle=RoyalBlue, title={Benchmark: MLLM-as-a-Judge}]

\begin{minipage}[t]{0.42\linewidth}
    \centering
    \includegraphics[width=\linewidth]{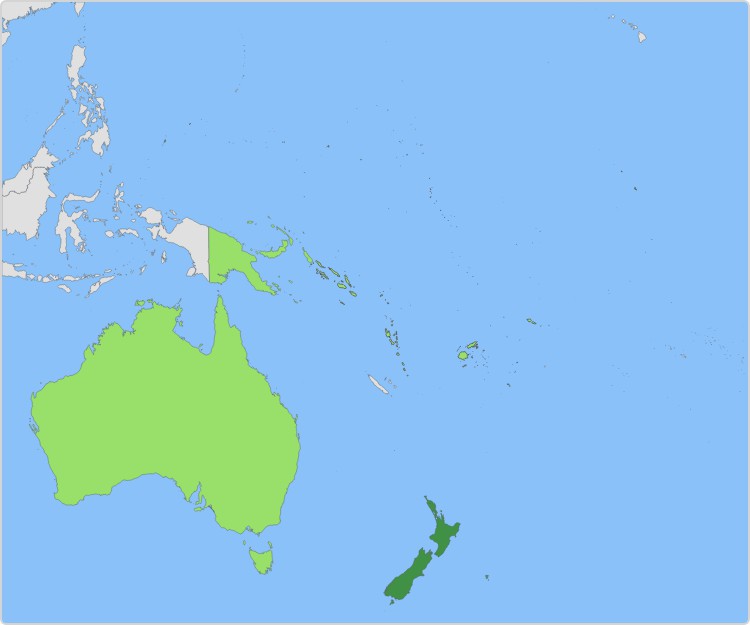}
\end{minipage}
\hfill
\begin{minipage}[t]{0.54\linewidth}
    \vspace{-14.5em}

    \textbf{Query:} Which country is highlighted?
\end{minipage}
    
\end{tcolorbox}

\begin{tcolorbox}[colback=Orange!10, colbacktitle=Orange, title={Candidate Answers}]

\textcolor{PineGreen}{\textbf{Answer A} (chosen):} The image you've provided is a map showing a large portion of the Oceania region. \textcolor{PineGreen}{The highlighted country, in green, is Australia.} Australia is the only country in the world that is also a continent by itself and is characterized by its considerable size covering the majority of the landmass shown at the center of the image. To the southeast of Australia, the two large islands highlighted in green represent New Zealand, which is another country.

\vspace{1em}

\textcolor{red}{\textbf{Answer B} (rejected):} New Zealand is highlighted.

\end{tcolorbox}

\begin{tcolorbox}[title={VLM-as-a-Judge (LLaMA-3.2-Vision-Instruct-90B)}]

\textbf{\texttt{Standard Ref}:}

\begin{itemize}
    \item \textbf{Preference:} \textcolor{red}{Answer B}
    \item \textbf{Reference answer:} The country highlighted on the map is New Zealand, which is located to the southeast of Australia and consists of two main islands, North Island and South Island, as well as numerous smaller islands.
    \item \textbf{Reason:} \textcolor{red}{Assistant A incorrectly identifies the highlighted country as Australia}, despite the image clearly showing New Zealand in dark green. Assistant B correctly identifies the highlighted country as New Zealand, aligning perfectly with the reference answer and the image. However, Assistant B's response is brief and lacks additional context, though this does not affect its correctness.
\end{itemize}

\textbf{\texttt{\OurMethod}:}

\begin{itemize} 
    \item \textbf{Preference:} \textcolor{PineGreen}{Answer A}
    \item \textbf{Anchor:} The image depicts a map of the Oceania region, with two countries highlighted in green: \textcolor{PineGreen}{Australia} and New Zealand. \textcolor{PineGreen}{Australia, the larger landmass located at the center-left of the image}, is both a country and a continent. To the southeast of Australia, New Zealand is highlighted, consisting of two smaller islands. Both countries are distinctly marked in green.
    \item \textbf{Reason:} Assistant A provides a detailed and accurate response, \textcolor{PineGreen}{identifying both Australia and New Zealand as the highlighted countries} in the image. The explanation includes relevant context about their locations and characteristics, aligning closely with the reference answer. In contrast, Assistant B's response is incomplete and incorrect, as it \textcolor{PineGreen}{only mentions New Zealand and ignores Australia, which is also highlighted} in the image.
\end{itemize}
    
\end{tcolorbox}

\caption{In this example, \texttt{Standard Ref} rejects the better Answer A, assuming Australia is not the highlighted country because the reference answer does not mention it. This shows how it over-penalizes informativeness. In contrast, \texttt{\OurMethod} includes the valid mention of Australia in the generated anchor, allowing it to correctly verify Answer A and achieve a better balance between informativeness and correctness.}
\label{fig:case-birch-vs-basic-ref-country}

\end{figure*}

\begin{figure*}[ht]
\centering

\tcbset{
    colback=gray!10, 
    colframe=white, 
    boxrule=0pt, 
    fonttitle=\bfseries,
    coltitle=white,
    colbacktitle=gray,
    width=\linewidth,
    arc=2mm,
    center title,
    % before upper={\fontsize{10}{12}\selectfont}
}

\begin{tcolorbox}[colback=RoyalBlue!10, colbacktitle=RoyalBlue, title={Benchmark: VL-RewardBench}]

\begin{minipage}[t]{0.42\linewidth}
    \centering
    \includegraphics[width=\linewidth]{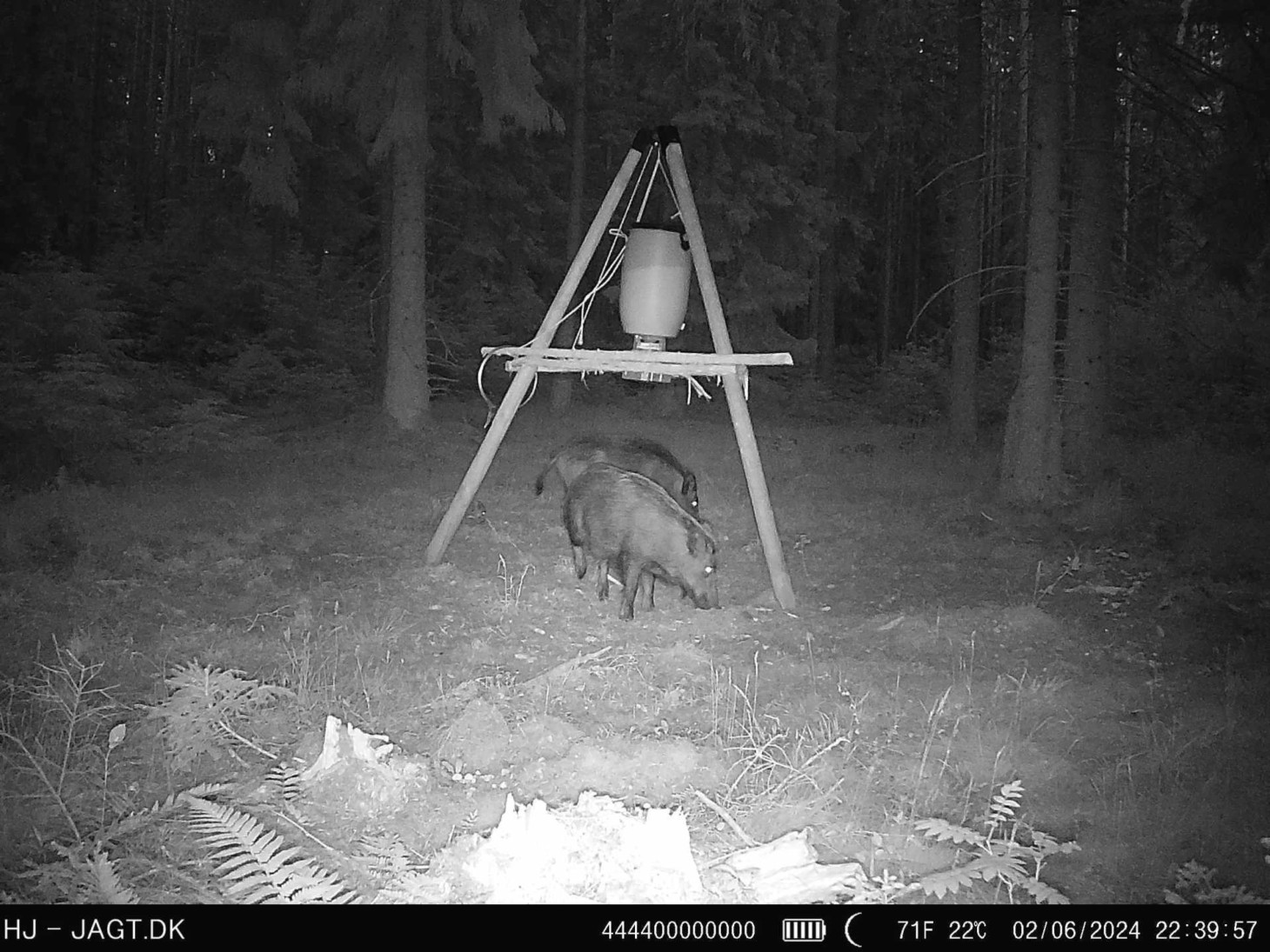}
\end{minipage}
\hfill
\begin{minipage}[t]{0.54\linewidth}
    \vspace{-12em}
    
    \textbf{Query:} Look at this image and determine which of the following animals is present: Boar, Badger Bird, Unknown
\end{minipage}
    
\end{tcolorbox}

\begin{tcolorbox}[colback=Orange!10, colbacktitle=Orange, title={Candidate Answers}]

\textcolor{red}{\textbf{Answer A} (rejected):} Boar

\vspace{1em}

\textcolor{PineGreen}{\textbf{Answer B} (chosen):} Based on the image, which shows a wild boar or pig foraging at a feeding station in a dark forest at night, the animal present is a Boar.

\end{tcolorbox}

\begin{tcolorbox}[title={VLM-as-a-Judge (LLaVA-OneVision-1.5-8B)}]

\textbf{\texttt{Standard Ref}:}

\begin{itemize}
    \item \textbf{Preference:} \textcolor{red}{Answer B}
    \item \textbf{Reference answer:} The animals present in the image are boars.
    \item \textbf{Reason:} Assistant B adds \textcolor{red}{extra context (a boar foraging at a feeding station) which is not supported} by the image or the reference answer.
\end{itemize}

\textbf{\texttt{\OurMethod}:}

\begin{itemize} 
    \item \textbf{Preference:} \textcolor{PineGreen}{Answer A}
    \item \textbf{Anchor:} Based on the image, which shows a wild boar foraging at a feeding station in a dark forest at night, the animal present is a Boar.
    \item \textbf{Reason:} Assistant B's response goes further by including details about the setting (a wild boar foraging at a \textcolor{PineGreen}{feeding station} in a dark forest at night), showing a clear understanding of the image and supports the conclusion.
\end{itemize}
    
\end{tcolorbox}

\caption{In this example, \texttt{Standard Ref} rejects the better Answer B, judging “boar foraging at a feeding station” as unsupported because the reference answer does not mention it. This shows how it over-penalizes informativeness. In contrast, \texttt{\OurMethod} includes this context in the generated anchor, allowing it to correctly verify Answer B and achieve a better balance between informativeness and correctness.}
\label{fig:case-birch-vs-basic-ref-boar}

\end{figure*}

Figures \ref{fig:case-birch-vs-basic-ref-stop-sign}, \ref{fig:case-birch-vs-basic-ref-country}, \ref{fig:case-birch-vs-basic-ref-car} (MLLM-as-a-Judge), and \ref{fig:case-birch-vs-basic-ref-boar} (VL-RewardBench) present qualitative examples showing why \texttt{Standard Ref}, which generates a reference without considering the candidate answers, performs poorly in some cases, and why our \texttt{\OurMethod} works better.

In \texttt{Standard Ref}, many details from the candidate answers may be missing in the reference. As a result, the judge often treats these extra but unverifiable details as wrong, even if they are actually correct. This leads to over-caution, an excessive focus on correctness, and weaker performance on informativeness-driven cases.

In contrast, \texttt{\OurMethod} ensures that every detail in the candidates is reflected in the reference, either corrected, removed with explanation, or preserved if valid. This way, the judge can always verify each detail against the reference, maintaining attention to correctness with image content while also preserving informativeness.

\subsection{Failure Case Analysis} \label{sec:appendix-case-failure}

We analyze failure cases where informativeness is misleading and explain why informativeness bias can persist even after applying \OurMethod.

\begin{figure*}[ht]
\centering

\tcbset{
    colback=gray!10, 
    colframe=white, 
    boxrule=0pt, 
    fonttitle=\bfseries,
    coltitle=white,
    colbacktitle=gray,
    width=\linewidth,
    % sharp corners,
    arc=2mm,
    center title,
    % before upper={\fontsize{10}{12}\selectfont}
}

\begin{tcolorbox}[colback=RoyalBlue!10, colbacktitle=RoyalBlue, title={Benchmark: MLLM-as-a-Judge}]

\begin{minipage}[t]{0.6\linewidth}
    \centering
    \includegraphics[width=\linewidth]{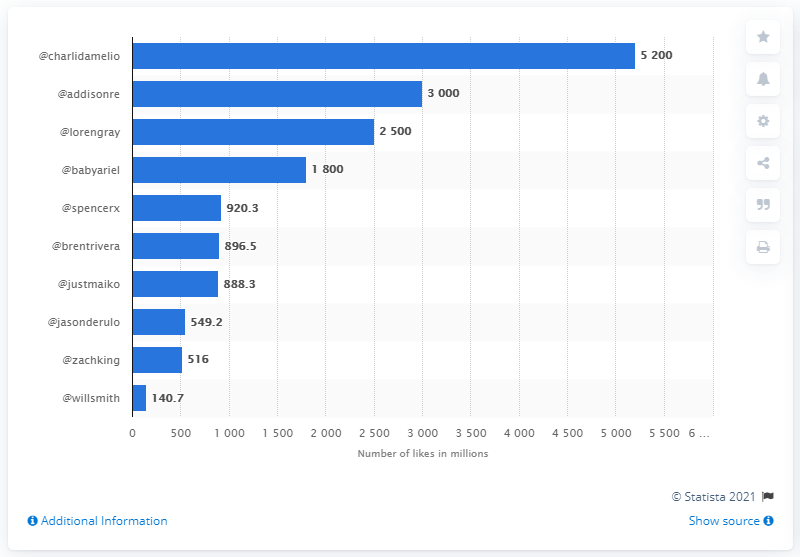}
\end{minipage}
\hfill
\begin{minipage}[t]{0.35\linewidth}
    \vspace{-15em}

    \textbf{Query:} How many likes did Charli D'Amelio have on her content?
\end{minipage}

\end{tcolorbox}

\begin{tcolorbox}[colback=Orange!10, colbacktitle=Orange, title={Candidate Answers}]

\textcolor{PineGreen}{\textbf{Answer A} (chosen):} Charli D'Amelio had \textcolor{PineGreen}{5,200 million} likes on her content.

\vspace{1em}

\textcolor{red}{\textbf{Answer B} (rejected):} Charli D'Amelio had \textcolor{red}{5.2 million} likes on her content. The figure shows the number of likes on the content of several TikTok users. Charli D'Amelio is the first user on the list, and the number of likes on her content is shown as 5.2 million.

\end{tcolorbox}

\begin{tcolorbox}[title={VLM-as-a-Judge (GPT-4o)}]

\textbf{\texttt{\OurMethod}:}

\begin{itemize} 
    \item \textbf{Preference:} \textcolor{red}{Answer B}
    
    \item \textbf{Anchor:} Charli D'Amelio had \textcolor{red}{5.2 million} likes on her content. This is evident from the bar chart, where she is listed as the first user, indicating the highest number of likes among all the individuals shown.
    
    \item \textbf{Reason:} Answer B identifies her like count as \textcolor{red}{5.2 million}, which matches the reference answer.
\end{itemize}
    
\end{tcolorbox}

\caption{Failure case study. \OurMethod\ generates an anchor with incorrect visual reasoning or is misled by the wrong Answer B, using 5.2 million as the like count instead of the correct 5,200 million. Based on this anchor, the judge incorrectly prefers Answer A.}
\label{fig:case-failure-likes}

\end{figure*}

In Figure \ref{fig:case-failure-likes}, the correct answer is 5,200 million. Answer A is correct, while Answer B incorrectly states 5.2 million. BIRCH fails to correct this inconsistency and includes it in the anchor. This may be due to limited visual or reasoning ability of the VLM, or because the model is still influenced by informativeness during anchor generation.

\begin{figure*}[ht]
\centering

\tcbset{
    colback=gray!10, 
    colframe=white, 
    boxrule=0pt, 
    fonttitle=\bfseries,
    coltitle=white,
    colbacktitle=gray,
    width=\linewidth,
    % sharp corners,
    arc=2mm,
    center title,
    % before upper={\fontsize{10}{12}\selectfont}
}

\begin{tcolorbox}[colback=RoyalBlue!10, colbacktitle=RoyalBlue, title={Benchmark: MLLM-as-a-Judge}]

\begin{minipage}[t]{0.5\linewidth}
    \centering
    \includegraphics[width=\linewidth]{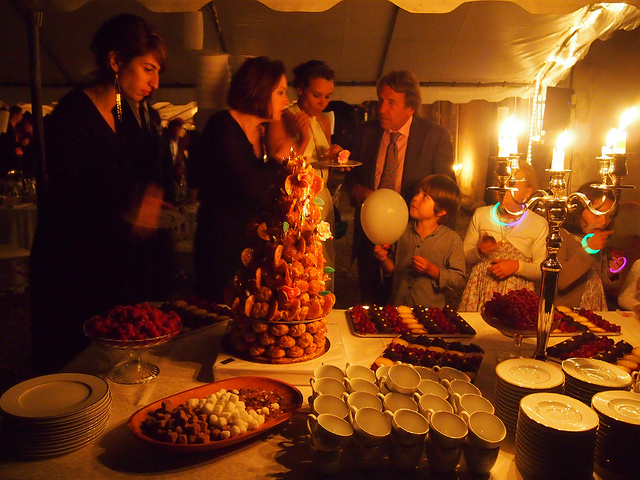}
\end{minipage}
\hfill
\begin{minipage}[t]{0.45\linewidth}
    \vspace{-12em}

    \textbf{Query:} What is filled in that oval object the boy is holding?
\end{minipage}

\end{tcolorbox}

\begin{tcolorbox}[colback=Orange!10, colbacktitle=Orange, title={Candidate Answers}]

\textcolor{PineGreen}{\textbf{Answer A} (chosen):} The oval object the boy is holding appears to be a balloon. It is \textcolor{PineGreen}{filled with a light gas}, which is commonly used in balloons to make them float.

\vspace{1em}

\textcolor{red}{\textbf{Answer B} (rejected):} That is a balloon. The object is oval-shaped, and the boy is holding its string. The object is also reflective, as light is reflecting off of its surface. This indicates that the object is not porous and is likely made of a smooth material. The object is also lightweight, as it is floating in the air. All of these properties indicate that the object is a balloon.

\end{tcolorbox}

\begin{tcolorbox}[title={VLM-as-a-Judge (GPT-4o)}]

\textbf{\texttt{\OurMethod}:}

\begin{itemize} 
    \item \textbf{Preference:} \textcolor{red}{Answer B}
    
    \item \textbf{Anchor:} The oval object the boy is holding is a white balloon, and he is holding its string. Given that it appears lightweight and to be floating in the air, it is \textcolor{PineGreen}{filled with gas, likely helium}.
    
    \item \textbf{Reason:} Answer B offers a more detailed description.
\end{itemize}
    
\end{tcolorbox}

\caption{Failure case study. \OurMethod\ generates a reliable anchor stating that the balloon held by the boy is filled with a light gas, likely helium. However, the judge still incorrectly selects the more informative Answer B, which does not specify what the balloon is filled with.}
\label{fig:case-failure-boy}

\end{figure*}

In Figure \ref{fig:case-failure-boy}, Answer 1 correctly states that the balloon is filled with a light gas. Answer 2 is more informative but does not specify what the balloon is filled with, so Answer 1 is better. In this case, the anchor is reliable, but BIRCH is still misled by the more informative Answer 2. This shows that anchors alone are not sufficient to eliminate informativeness bias, and we leave this limitation to future work.

\begin{table}[t]
    \centering
        
    \resizebox{\linewidth}{!}{

    \begin{tabular}{lccccc}
        \toprule
        
        \textbf{Methods} & \textbf{Acc} & \textbf{$\text{Acc}_\text{IDS}$} & \textbf{$\text{Acc}_\text{CDS}$} & \textbf{IB} & \textbf{LB} \\
        
        \midrule

        \textbf{GPT-4o}  & 66.5 & 85.3 & 26.3 & 59.0 & 32.9 \\

        \textbf{\; + \OurMethod\ (Ours)} & 75.8 \textcolor{PineGreen}{(+9.3)} & 90.1 \textcolor{PineGreen}{(+4.8)} & 45.3 \textcolor{PineGreen}{(+19)} & 44.8 \textcolor{PineGreen}{(-14.2)} & 20.3 \textcolor{PineGreen}{(-12.6)} \\[1em]

        \textbf{Gemini-2.5-Flash}  & 66.9 & 79.9 & 39.3 & 40.6 & 9.9 \\

        \textbf{\; + \OurMethod\ (Ours)} & 73.4 \textcolor{PineGreen}{(+6.5)} & 83.4  \textcolor{PineGreen}{(+3.5)} &  52.0 \textcolor{PineGreen}{(+12.7)} &  31.4 \textcolor{PineGreen}{(-9.2)} &  9.2 \textcolor{PineGreen}{(-0.7)} \\[1em]

        \textbf{Llama-3.2-90B}  & 65.3 & 82.2 & 29.3 & 52.9 & 26.7 \\

        \textbf{\; + \OurMethod\ (Ours)} & 75.1 \textcolor{PineGreen}{(+9.8)} & 86.6 \textcolor{PineGreen}{(+4.4)} & 50.7 \textcolor{PineGreen}{(+21.4)} & 35.9  \textcolor{PineGreen}{(-17.0)} & 20.4 \textcolor{PineGreen}{(-6.3)} \\[1em]

        \textbf{LLaVA-OV-1.5-8B}  & 65.4 & 80.4  &  33.3 &  47.1 & 25.8  \\

        \textbf{\; + \OurMethod\ (Ours)} & 71.2 \textcolor{PineGreen}{(+5.8)} & 82.9  \textcolor{PineGreen}{(+2.5)} & 46.2 \textcolor{PineGreen}{(+12.9)} & 36.7  \textcolor{PineGreen}{(-10.4)} & 20.7  \textcolor{PineGreen}{(-5.1)} \\[1em]

        \textbf{Phi-4-multimodal}  & 59.2 & 68.6 & 39.1 & 29.5 & 19.3 \\

        \textbf{\; + \OurMethod\ (Ours)} & 63.1 \textcolor{PineGreen}{(+3.9)} & 71.9  \textcolor{PineGreen}{(+3.3)} & 44.4  \textcolor{PineGreen}{(+5.3)} & 27.5   \textcolor{PineGreen}{(-2.0)} & 14.6  \textcolor{PineGreen}{(-5.3)} \\
        
        \bottomrule
    \end{tabular}
    }

    % \caption{Overall accuracy, accuracy on informativeness-driven cases $\text{Acc}_\text{IDS}$ and correctness-driven cases $\text{Acc}_\text{CDS}$, and informativeness bias (IB) before and after applying \texttt{Standard Ref} and \texttt{\OurMethod} on MLLM-as-a-Judge \cite{chen2024mllm} benchmark.}

    \caption{Overall accuracy, accuracy on informativeness-driven cases $\text{Acc}_\text{IDS}$ and correctness-driven cases $\text{Acc}_\text{CDS}$, informativeness bias (IB), and length bias (LB) before and after applying \texttt{\OurMethod} on the MLLM-as-a-Judge \cite{chen2024mllm} benchmark. \OurMethod\ reduces IB by greatly improving $\text{Acc}_\text{CDS}$ while keeping $\text{Acc}_\text{IDS}$ strong, leading to higher overall performance. Moreover, although designed to mitigate IB, \OurMethod\ also greatly reduces LB.
    }

    \label{table:acc-ida-cda-ib-lb-full}
\end{table}

\begin{figure*}[t]
  \includegraphics[width=\linewidth]{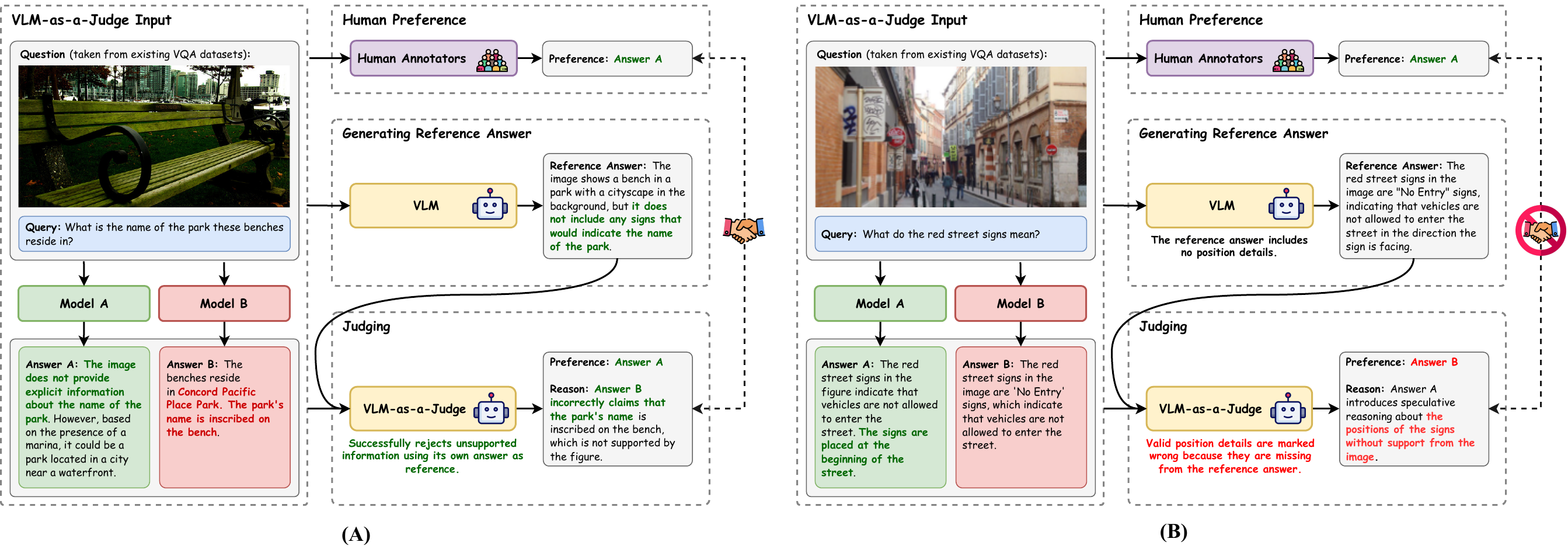}
  \caption{Successful and failed cases with the \texttt{Standard Ref} pipeline, where the judge first generates its own answer as a reference and then compares Answer A and B against it to decide which is better. \textbf{(A) Successful case:} Using the judge's answer "it does not include any signs that would indicate the name of the park" as reference, the judge correctly identifies an error in Answer B, which claims the park name is inscribed on the bench when it is not. This helps the judge overcome informativeness bias and make the right decision. \textbf{(B) Failed case:} \texttt{Standard Ref} can over-prioritize correctness and unfairly penalize valid details missing from the reference. Here, Answer A is rejected simply for mentioning the position (marked in green) of the "No Stop" sign, information that is valid but absent from the reference, leading it to be mistakenly judged as "unsupported by the image".}
  \label{fig:standard-ref-bad}
\end{figure*}

\begin{figure*}[ht]
  \includegraphics[width=\linewidth]{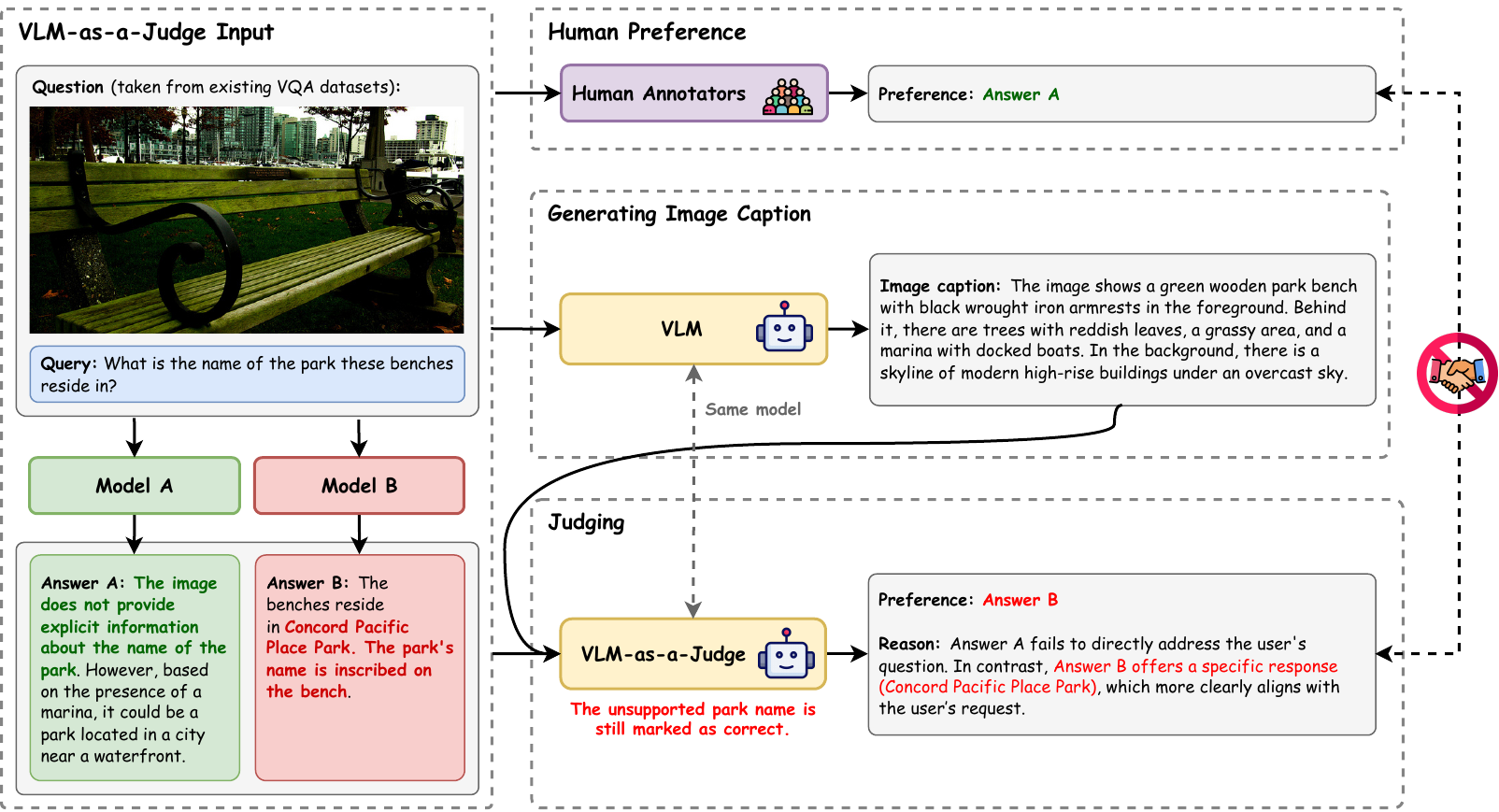}
  \caption{Illustration of the \texttt{Image Caption} baseline and why it performs poorly. First, the same VLM later used as the judge is asked to generate a caption for the image. Then, the judge takes the query, image, two candidate answers, and the caption as input, and decides which answer better addresses the query. However, the caption does not mention whether the park name is visible, so the judge still accepts the unsupported park name in Answer B as correct. This highlights the need for \OurMethod, which ensures that every detail in the candidate answers is also represented in the reference.}
  \label{fig:method-image-caption}
\end{figure*}

\begin{figure*}[ht]
\centering

\begin{tcolorbox}[colback=gray!10]

\textbf{Instruction}

Please act as an impartial judge and evaluate the informativeness of two candidate responses from different assistants for the same user instruction, which originally includes an image and a textual instruction. Your only job is to decide which response is more informative. You will not be given the image. Do NOT consider factual correctness or image relevance, just evaluate informativeness. \newline

Definition of informativeness:

\begin{enumerate}
    \item Relevance: Does the response address the user's intent clearly?
    \item Completeness: Does it provide necessary information or explanation the user would need?
    \item Conciseness: Does it avoid unnecessary details or digressions?
\end{enumerate}

Your evaluation should focus on these dimensions: helpfulness, relevance, depth, and the level of detail. There is no tie. Choose the more informative one. Output your reasoning first, then the decision. \newline

\textbf{Output format} 

Analysis: <1-2 sentence explanation>

Judgement: [[A]] or [[B]] \newline

\textbf{Inputs}

[The Start of User Instruction] \{ user query \} [The End of User Instruction]

[The Start of Assistant A's Answer] \{ Answer A \} [The End of Assistant A's Answer]

[The Start of Assistant B's Answer] \{ Answer B \} [The End of Assistant B's Answer] \\

\end{tcolorbox}

\caption{Prompt for comparing informativeness. We provide GPT-4o with the original query and two candidate answers (without the image) and ask it to focus only on helpfulness, relevance, depth, and level of detail when determining which answer is more informative, ignoring image alignment correctness.}
\label{table:info-prompt}

\end{figure*}
\begin{figure*}[ht]
\centering

\begin{tcolorbox}[colback=gray!10]

\textbf{Instruction}

You are a helpful and knowledgeable assistant. The user has provided an image and a question or instruction related to it. Carefully analyze the image and follow the instruction to generate a clear, concise answer. \newline

\textbf{Inputs}

\{ image \}

[The Start of User Instruction] \{ user query \} [The End of User Instruction]

\end{tcolorbox}

\caption{Prompt for generating reference answers in \texttt{Standard Ref} baseline.}
\label{table:ref-prompt}

\end{figure*}
\begin{figure*}[ht]
\centering

\begin{tcolorbox}[colback=gray!10]

\textbf{Instruction}

Please act as a skilled reviser and editor to refine AI assistant responses based on a user instruction referring to a figure. \newline

You will be given two assistant-generated responses. These responses are NOT guaranteed to be correct or truthful. Your tasks are:

\begin{enumerate}
    \item Revise each response individually: Correct any errors or inconsistencies with the image content. Ensure the response is accurate, truthful, and fully addresses the user's instruction. Carefully search for mistakes, since the response may not be fully correct or aligned with the image. Only make revisions when necessary, and DO NOT alter the original style or tone. Remove irrelevant or speculative parts that cannot reasonably be inferred from the image, and leave comments explaining what was removed and why. If the original response is good enough, leave it unchanged.
    \item Merge the two revised responses into one unified response: Eliminate redundancy and overlap between the two responses. When both responses mention the same idea, keep the version that is more accurate, helpful, relevant, comprehensive, creative, and detailed. Maintain as much of the original style and tone as possible while merging.
\end{enumerate}

Be precise and objective in your edits. \newline

\textbf{Output format} 

Revised Response A: <revised response A> 

Revised Response B: <revised response B> 

Merged Response: <merged response> \newline

\textbf{Inputs}

\{ image \}

[The Start of User Instruction] \{ user query \} [The End of User Instruction]

[The Start of Assistant A's Answer] \{ Answer A \} [The End of Assistant A's Answer]

[The Start of Assistant B's Answer] \{ Answer B \} [The End of Assistant B's Answer] \\

\end{tcolorbox}

\caption{Prompt for generating an anchor in \OurMethod.}
\label{table:birch-ref-prompt}

\end{figure*}
\begin{figure*}[ht]
\centering

\begin{tcolorbox}[colback=gray!10]

\textbf{Instruction}

Please act as an impartial judge and evaluate the quality of two responses generated by different AI assistants for the same user question and figure. \newline

You will be provided with:

\begin{enumerate}
    \item A user instruction
    \item A figure (relevant to the instruction)
    \item A reliable and accurate reference answer
    \item Two candidate responses (Assistant A and Assistant B)
\end{enumerate}

You should carefully examine the figure and compare the two responses with the reference answer. \newline

You should then choose the assistant that better follows the user's instructions and more effectively answers the user's question. \newline

Your evaluation should consider factors such as the helpfulness, relevance, accuracy, depth, creativity, and level of detail of their responses. \newline

Do not let the position or order of responses influence your decision. Do not allow the length of the responses to influence your evaluation. Do not favor certain names of the assistants. Be as objective as possible. \newline
    
\textbf{Output format} 

Analysis: <1-2 sentence explanation comparing Assistant A and Assistant B.>

Judgement: [[A]] or [[B]] (Choose [[A]] if Assistant A is better, or [[B]] if Assistant B is better) \newline

\textbf{Inputs}

\{ image \}

[The Start of User Instruction] \{ user query \} [The End of User Instruction]

[The Start of Assistant A's Answer] \{ Answer A \} [The End of Assistant A's Answer]

[The Start of Assistant B's Answer] \{ Answer B \} [The End of Assistant B's Answer]

[The Start of Reference Answer] \{ Reference Answer \} [The End of Reference Answer]

\end{tcolorbox}

\caption{Prompt for judging with reference answer in \OurMethod.}
\label{table:judge-w-ref-prompt}

\end{figure*}

\begin{figure*}[ht]
\centering

\tcbset{
    colback=gray!10, 
    colframe=white, 
    boxrule=0pt, 
    fonttitle=\bfseries,
    coltitle=white,
    colbacktitle=gray,
    width=\linewidth,
    arc=2mm,
    center title,
    before upper={\fontsize{10}{12}\selectfont}
}

\begin{tcolorbox}[colback=RoyalBlue!10, colbacktitle=RoyalBlue, title={Benchmark: MLLM-as-a-Judge}]

\begin{minipage}[t]{0.35\linewidth}
    \centering
    \includegraphics[width=\linewidth]{others/case-studies/basic-ref-vs-no-ref/cigarette.jpg}
\end{minipage}
\hfill
\begin{minipage}[t]{0.6\linewidth}
    \vspace{-13em}

    \textbf{Query:} Please analyse this figure in detail and follow the following instruction based on this figure: Generate a question based on the given image such that it receives the following answer: cigarette

    \vspace{7em}
    \textcolor{black!60}{\footnotesize Zoom in to see the image details more clearly.}
\end{minipage}

\end{tcolorbox}

\begin{tcolorbox}[colback=Orange!10, colbacktitle=Orange, title={Candidate Answers}]

\textcolor{red}{\textbf{Answer A} (rejected):} The image depicts a person riding a motorcycle with a helmet on. The person has tattoos on their left arm and is wearing a sleeveless shirt, which allows for the visibility of the tattoos. \textcolor{red}{On the person's right hand}, there is an object that appears to be a cigarette between the fingers, suggested by its slim, cylindrical shape and the way it's being held, typical of how one would hold a cigarette. The setting appears to be outdoors during daytime, as indicated by the sunlight and shadows. Based on these observations, a suitable question to elicit the response 'cigarette' could relate to \textcolor{red}{what the person is holding in their right hand} while operating the motorcycle.

\vspace{1em}

\textcolor{PineGreen}{\textbf{Answer B} (chosen):} What is the man on the motorcycle \textcolor{PineGreen}{holding in his mouth?} The answer is "cigarette". The man's mouth is slightly open and there is a small, cylindrical object between his lips. The object is likely a cigarette because it is thin and has a filter on the end. The man is also holding the object in a way that suggests that he is smoking it.

\end{tcolorbox}

\begin{tcolorbox}[title={VLM-as-a-Judge (LLaMA-3.2-Vision-Instruct-90B)}]

\textbf{\texttt{Without Reference Answer}:}

\begin{itemize}
    \item \textbf{Preference:} \textcolor{red}{Answer B}
    \item \textbf{Reason:} Assistant A provides a detailed description of the image, focusing on the person riding the motorcycle and the object in their right hand, which is identified as a cigarette. The assistant suggests a question that aligns with the user's instruction by asking about what the person is holding in their right hand. The response is comprehensive and adheres to the instruction effectively. Assistant B proposes a question about what the man is holding in his mouth. While the response is relevant, \textcolor{red}{it assumes the cigarette is in the mouth without clear evidence from the image, which could lead to inaccuracies}. Additionally, \textcolor{red}{the explanation is less detailed compared to Assistant A's response}.
\end{itemize}

\textbf{\texttt{With Reference Answer}:}

\begin{itemize} 
    \item \textbf{Preference:} \textcolor{PineGreen}{Answer B}
    \item \textbf{Reference answer generated by the judge itself:} What object is the person riding the motorcycle \textcolor{PineGreen}{holding in their mouth}?
    \item \textbf{Reason:} Assistant A provides a detailed description of the image but \textcolor{PineGreen}{incorrectly identifies the cigarette as being held in the person's right hand}, which is inconsistent with the reference answer and the image. \textcolor{PineGreen}{Assistant B correctly identifies the cigarette as being held in the person's mouth}, aligning with the reference answer and the image. Assistant B's response is more accurate and helpful.
\end{itemize}
    
\end{tcolorbox}

\caption{In this example, VLM-as-a-Judge prefers Answer A, which provides more details about the environment but incorrectly states that the cigarette is in the person’s right hand when it is actually in their mouth. The model does recognize the correct location of the cigarette, yet still judges Answer A as correct. Using its own generated answer as a reference helps mitigate this issue.}
\label{fig:case-basic-ref-vs-no-ref-cigarette}

\end{figure*}

\begin{figure*}[ht]
\centering

\tcbset{
    colback=gray!10, 
    colframe=white, 
    boxrule=0pt, 
    fonttitle=\bfseries,
    coltitle=white,
    colbacktitle=gray,
    width=\linewidth,
    arc=2mm,
    center title,
    before upper={\fontsize{10}{12}\selectfont}
}

\begin{tcolorbox}[colback=RoyalBlue!10, colbacktitle=RoyalBlue, title={Benchmark: MLLM-as-a-Judge}]

\begin{minipage}[t]{0.42\linewidth}
    \centering
    \includegraphics[width=\linewidth]{others/case-studies/birch-vs-basic-ref/car.jpg}
\end{minipage}
\hfill
\begin{minipage}[t]{0.54\linewidth}
    \vspace{-9em}

    \textbf{Query:} What is the make of the car on the left?
\end{minipage}

\end{tcolorbox}

\begin{tcolorbox}[colback=Orange!10, colbacktitle=Orange, title={Candidate Answers}]

\textcolor{red}{\textbf{Answer A} (rejected):} The make of the car on the left appears to be Volkswagen, as indicated by the logo visible on its front.

\vspace{1em}

\textcolor{PineGreen}{\textbf{Answer B} (chosen):} The car on the left appears to be a Volkswagen (VW), as indicated by the distinctive front grille design and the logo on the hood. Volkswagen is a German multinational automotive manufacturing company known for its wide range of vehicles, including sedans, hatchbacks, and SUVs. The specific model of this car is not clearly identifiable from this angle and without more context, but \textcolor{PineGreen}{it's part of the brand's lineup of electric or hybrid vehicles, as evidenced by the presence of a charging cable and plug connected to the car}.

\end{tcolorbox}

\begin{tcolorbox}[title={VLM-as-a-Judge (Gemini-2.5-Flash)}]

\textbf{\texttt{Standard Ref}:}

\begin{itemize}
    \item \textbf{Preference:} \textcolor{red}{Answer A}
    \item \textbf{Reference answer:} The car on the left is a Volkswagen, as indicated by the logo on the front grille.
    \item \textbf{Reason:} Assistant A provides a concise and accurate response, correctly identifying the make of the car on the left as Volkswagen based on the visible logo. Assistant B also identifies the car as a Volkswagen \textcolor{red}{but speculates about the car's model and type (electric or hybrid), which is not supported by the image}. This extra information does not enhance the response and could potentially mislead.
\end{itemize}

\textbf{\texttt{\OurMethod}:}

\begin{itemize} 
    \item \textbf{Preference:} \textcolor{PineGreen}{Answer B}
    \item \textbf{Anchor:} The car on the left appears to be a Volkswagen (VW), as identified by the logo on its front grille. \textcolor{PineGreen}{It is connected to a charging station via a cable, indicating it is likely part of Volkswagen's lineup of electric or hybrid vehicles.} The specific model, however, cannot be determined from this angle.
    \item \textbf{Reason:} Assistant A provides a concise and accurate answer, identifying the make of the car as Volkswagen based on the visible logo. However, it lacks additional context or detail. Assistant B also correctly identifies the make as Volkswagen and provides more depth by mentioning the grille design, the charging cable, and \textcolor{PineGreen}{the likelihood of the car being part of Volkswagen's electric or hybrid lineup}. This additional context aligns well with the reference answer, making Assistant B's response more comprehensive and informative.
\end{itemize}
    
\end{tcolorbox}

\caption{In this example, \texttt{Standard Ref} rejects the better Answer B, reasoning that electric or hybrid vehicles cannot be inferred from the image (even though the charging cable and plug suggest this) because the reference answer does not mention it. This shows how it over-penalizes informativeness. In contrast, \texttt{\OurMethod} includes the likelihood of the car being electric or hybrid in the generated anchor, allowing it to correctly verify Answer B and achieve a better balance between informativeness and correctness.}
\label{fig:case-birch-vs-basic-ref-car}

\end{figure*}

\end{document}